\documentclass{tlp}

\usepackage{times}

\usepackage[linesnumbered,ruled,vlined]{algorithm2e}

\usepackage{graphicx}
\usepackage{color}
\usepackage{float}
\usepackage{xspace,epsfig,url}
\usepackage{epstopdf}

\usepackage{tikz}
\usetikzlibrary{positioning,decorations.pathmorphing}

\usepackage{subfig}
\usepackage{amsmath,amssymb,latexsym} 

\usepackage{enumerate}

\usepackage[T1]{fontenc}

\usepackage{pdfpages}

\def\pharmgkb{{\sc PharmGKB}}
\def\drugbank{{\sc DrugBank}}
\def\sider{{\sc SIDER}}
\def\biogrid{\hbox{{\sc BioGRID}}}
\def\ctd{{\sc CTD}}
\def\donto{{\sc Disease Ontology}}
\def\orphadata{{\sc Orphadata}}

\def\bqasp{{\sc BioQuery-ASP}}
\def\expgenasp{{\sc ExpGen-ASP}}
\def\bqcnl{{\sc BioQuery-CNL}}
\def\bqcnlnew{{\sc BioQuery-CNL*}}

\def\clasp{{\sc clasp}}
\def\gringo{{\sc gringo}}

\def\dlvhex{{\sc dlvhex}}

\def\ii#1{\hbox{\it #1\/}}

\def\no{\ii{not}\ }
\def\lar{\leftarrow}
\def\ba{\begin{array}}
\def\ea{\end{array}}
\def\beq{\begin{equation}}
\def\eeq#1{\label{#1}\end{equation}}
\def\bi{\begin{itemize}}
\def\ei{\end{itemize}}

\def\sep{\,|\,}
\def\seq#1{\left\langle #1 \right\rangle}

\newtheorem{definition}{Definition}
\newtheorem{example}{Example}

\newtheorem{proposition}{Proposition}

\newtheorem{corollary}{Corollary}

\begin{document}
\title[Generating Explanations for Biomedical Queries]
{Generating Explanations for Biomedical Queries}

\author[Esra Erdem, Umut Oztok]{
ESRA ERDEM, UMUT OZTOK\\
Sabanc{\i} University, Orhanl{\i}, Tuzla, \.Istanbul 34956, Turkey \\
\email{\{esraerdem,uoztok\}@sabanciuniv.edu}}

\pagerange{\pageref{firstpage}--\pageref{lastpage}}
\volume{\textbf{10} (3):} \jdate{January 2012} \setcounter{page}{1}
\pubyear{2012}

\submitted{25 October 2012}
\revised{17 July 2013}
\accepted{19 August 2013}

\maketitle \label{firstpage}

\begin{abstract}
We introduce novel mathematical models and algorithms to generate
(shortest or $k$ different) explanations for biomedical queries,
using answer set programming. We implement these algorithms and
integrate them in \bqasp. We illustrate the usefulness of these
methods with some complex biomedical queries
related to drug discovery, over the biomedical
knowledge resources \pharmgkb, \drugbank, \biogrid, \ctd, \sider,
\donto\ and \orphadata.
\end{abstract}



\section{Introduction}
\label{sec:intro}

Recent advances in health and life sciences have led to generation
of a large amount of biomedical data, represented in various
biomedical databases or ontologies. That these databases and
ontologies are represented in different formats and
constructed/maintained independently from each other at different
locations, have brought about many challenges for answering complex
biomedical queries that require integration of knowledge represented
in these ontologies and databases.
One of the challenges for the users is to be able to represent such
a biomedical query in a natural language, and get its answers in an
understandable form. Another challenge is to extract relevant
knowledge from different knowledge resources, and integrate them
appropriately using also definitions, such as, chains of gene-gene
interactions, cliques of genes based on gene-gene relations, or
similarity/diversity of genes/drugs. Furthermore, once an answer is
found for a complex query, the experts may need further explanations
about the answer.

Table~\ref{tab:queries} displays a list of complex biomedical
queries that are important from the point of view of drug discovery.
In the queries, drug-drug interactions present negative interactions
among drugs, and gene-gene interactions present both negative and
positive interactions among genes. Consider, for instance the
query~Q6. New molecule synthesis by changing substitutes of parent
compound may lead to different biochemical and physiological
effects; and each trial may lead to different indications. Such
studies are important for fast inventions of new molecules.  For
example, while developing the drug Lovastatin (a member of the drug
class of Hmg-coa reductase inhibitors, used for lowering
cholesterol) from {\it Aspergillus terreus} (a sort of fungus) in
1979, scientists at Merck derived a new molecule named Simvastatin
that also belongs to the same drug category (a hypolipidemic drug
used to control elevated cholesterol) targeting the same gene.
Therefore, identifying genes targeted by a group of drugs
automatically by means of queries like Q6 may be useful for experts.

Once an answer to a query is found, the experts may ask for an
explanation to have a better understanding. For instance, an answer
for the query~Q3 in Table~\ref{tab:queries} is ``ADRB1''.  A
shortest explanation for this answer is as follows:
\begin{itemize}
\item[]
The drug Epinephrine targets the gene ADRB1 according to \ctd. \\
The gene~DLG4 interacts with the gene ADRB1 according to \biogrid.
\end{itemize}
An answer for the query~Q8 is ``CASK''. A shortest explanation for
this answer is as follows:

\medskip
The distance of the gene CASK from the start gene is 2.

\hspace{2em}  The gene CASK interacts with the gene DLG4 according to \biogrid.

\hspace{2em}  The distance of the gene DLG4 from the start gene is 1.

\hspace{4em}     The gene DLG4 interacts with the gene ADRB1 according to \biogrid.

\hspace{4em}     ADRB1 is the start gene.
\medskip

\noindent (Statements with more indentations provide explanations
for statements with less indentations.)

\begin{table}[t!]
\caption{A list of complex biomedical queries.} \label{tab:queries}
\fbox{
\begin{minipage}{\textwidth}
{\small
\begin{itemize}
\item [Q1] What are the drugs that treat the disease Asthma and that target the gene ADRB1?

\item [Q2] What are the side effects of the drugs that treat the disease Asthma and that target the gene~ADRB1?

\item [Q3] What are the genes that are targeted by the drug Epinephrine and that interact with the gene~DLG4?

\item [Q4] What are the genes that interact with at least $3$ genes and that are targeted by the drug~Epinephrine?

\item [Q5] What are the drugs that treat the disease Asthma or that react with the drug Epinephrine?

\item [Q6]
What are the genes that are targeted by all the drugs that belong to
the category Hmg-coa reductase inhibitors?

\item [Q7] What are the cliques of $5$ genes, that contain the gene DLG4?

\item [Q8] What are the genes that are related to the gene ADRB1 via a gene-gene interaction chain of length at most $3$?

\item [Q9] What are the $3$ most similar genes that are targeted by the drug Epinephrine?

\item [Q10] What are the genes that are related to the gene DLG4 via a gene-gene interaction chain of length at most $3$ and
that are targeted by the drugs that belong to the category Hmg-coa
reductase inhibitors?
\item [Q11] What are the drugs that treat the disease Depression and that do not target the gene~ACYP1?

\item [Q12] What are the symptoms of diseases that are treated by the drug Triadimefon?

\item [Q13] What are the $3$ most similar drugs that target the gene DLG4?

\item [Q14] What are the $3$ closest drugs to the drug Epinephrine?
\end{itemize}}
\end{minipage}}
\end{table}

\begin{figure}[t!]
\begin{center}
\includegraphics[scale=0.4]{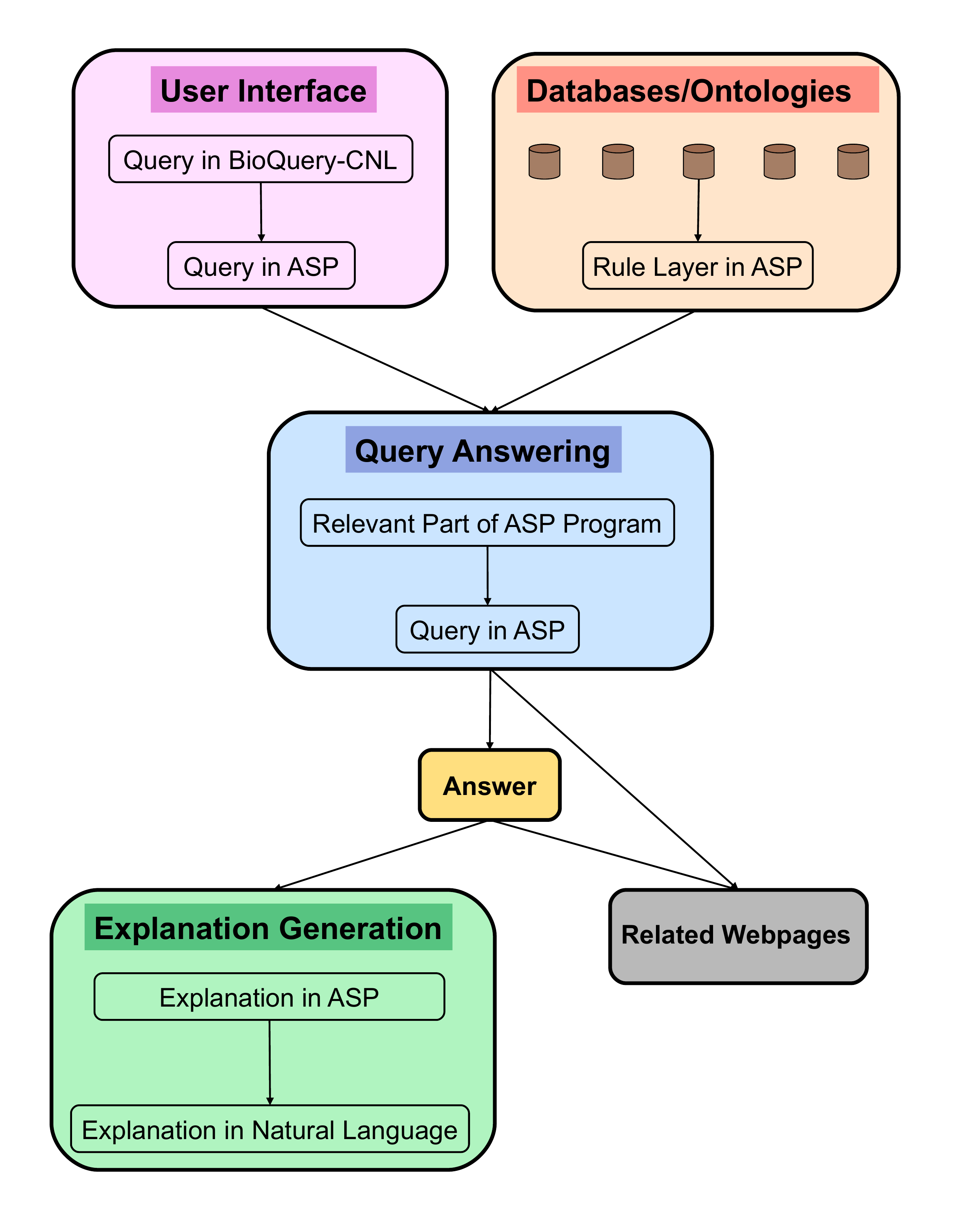}
\vspace{-\baselineskip}
\caption{System overview of \bqasp.}
\label{fig:bqasp}
\end{center}
\end{figure}

To address the first two challenges described above (i.e.,
representing complex queries in natural language and finding answers
to queries efficiently), novel methods and a software system, called
\bqasp~\cite{ErdemEO11} (Figure~\ref{fig:bqasp}), have been
developed using Answer Set Programming
(ASP)~\cite{mar99,nie99,Lifschitz-Answer-2002,Baral03,Lifschitz08,BrewkaET11}:

\begin{itemize}
\item Erdem and Yeniterzi~\cite{ErdemY09} developed a controlled
natural language, \hbox{\bqcnl}, for expressing biomedical queries
related to drug discovery. For instance, queries Q1--Q10 in
Table~\ref{tab:queries} are in this language. Recently, this
language has been extended (called \bqcnlnew) to cover queries
Q11--Q13~\cite{Oztok12}. Some algorithms have been introduced to
translate a given query in \bqcnl\ (resp. \bqcnlnew) to a program in
ASP as well.
\item Bodenreider et al.~\cite{BodenreiderCDEK08} introduced methods to extract
biomedical information from various knowledge resources and
integrate them by a rule layer. This rule layer not only integrates
those knowledge resources but also provides definitions of auxiliary
concepts.
\item Erdem et al.~\cite{ErdemEEO11} have introduced an algorithm for
query answering by identifying the relevant parts of the rule layer
and the knowledge resources with respect to a given query.
\end{itemize}
The details of representing biomedical queries in natural language
and answering them using ASP are explained in a companion article.
The focus of this article is the last challenge: generating
explanations for biomedical queries.

Most of the existing biomedical querying systems (e.g., web services
built over the available knowledge resources) support keyword search
but not complex queries like the queries in Table~\ref{tab:queries}.
None of the existing systems can provide informative explanations
about the answers, but point to related web pages of the knowledge
resources available online.

The contributions of this article can be summarized as follows.

\begin{itemize}
\item
We have formally defined ``explanations'' in ASP, utilizing properties of programs
and graphs. We have also defined variations of explanations, such as
``shortest explanations'' and ``$k$ different explanations''.

\item
We have introduced novel generic algorithms to generate explanations
for biomedical queries. These algorithms can compute shortest or $k$
different explanations. We have analyzed the termination, soundness,
and complexity of those algorithms.

\item
We have developed a computational tool, called \expgenasp, that
implements these explanation generation algorithms.

\item
We have showed the applicability of our methods to generate
explanations for answers of complex biomedical queries related to
drug discovery.

\item
We have embedded \expgenasp\ into \bqasp\ so that the experts can
obtain explanations regarding the answers of biomedical queries, in
a natural language.
\end{itemize}

The rest of the article is organized as follows. In
Section~\ref{sec:asp}, we provide a summary of Answer Set
Programming. Next, in Section~\ref{sec:query}, we give an overview
of \bqasp, in particular, the earlier work done on answering biomedical queries in ASP.
Then, in Sections~\ref{sec:exp}--\ref{sec:kDiff}, we provide some definitions and algorithms
related to generating shortest or $k$ different explanations for an answer, also in ASP.
Next, Section~\ref{sec:expEX} illustrates the usefulness of these algorithms on
some complex queries over the biomedical knowledge resources
\pharmgkb~\cite{pharmgkb},\footnote{\url{http://www.pharmgkb.org/}}
\drugbank~\cite{drugbank},\footnote{\url{http://www.drugbank.ca/}}
\biogrid~\cite{biogrid},\footnote{\url{http://thebiogrid.org/}}
\ctd~\cite{ctd},\footnote{\url{http://ctd.mdibl.org/}}
\sider~\cite{sider},\footnote{\url{http://sideeffects.embl.de/}}
\donto~\cite{diseaseontology}\footnote{\url{http://disease-ontology.org}} and
\orphadata.\footnote{\url{http://www.orphadata.org}}
In Sections~\ref{sec:expNL} and~\ref{sec:expIMP}, we discuss
how to present explanations to the user in a natural language,
and embedding of these algorithms in \bqasp.
In Section~\ref{sec:related}, we provide a detailed analysis of the related work on
``justifications''~\cite{PontelliSE09} in comparison to explanations; and
in Section~\ref{sec:relatedWork}, we briefly discuss other related work.
We conclude in Section~\ref{sec:conclusion} by summarizing our
contributions and pointing out some possible future work.
Proofs are provided in the online appendix of the paper.


\section{Answer Set Programming}
\label{sec:asp}

Answer Set Programming (ASP)~\cite{mar99,nie99,Lifschitz-Answer-2002,Baral03,Lifschitz08,BrewkaET11} is a form of declarative programming paradigm
oriented towards solving combinatorial search problems as well as
knowledge-intensive problems. The idea is to represent a problem as
a ``program'' whose models (called ``answer
sets''~\cite{GelfondL88,Gelfond-Classical-1991}) correspond to the solutions. The answer
sets for the given program can then be computed by special
implemented systems called answer set solvers. ASP has a high-level
representation language that allows recursive definitions,
aggregates, weight constraints, optimization statements, and default
negation.

ASP also provides efficient solvers, such as
\clasp~\cite{GebserKNS07}. Due to the continuous improvement of the
ASP solvers and highly expressive representation language of ASP
which is supported by a strong theoretical background that results
from a years of intensive research, ASP has been applied fruitfully
to a wide range of areas. Here are, for instance, three applications
of ASP used in industry:

\begin{itemize}
\item
\textit{Decision Support Systems}: An ASP-based system was developed
to help flight controllers of space shuttle solve some planning and
diagnostic tasks~\cite{Nogueira-A-Prolog-2001} (used by United Space
Alliance).

\item
\textit{Automated Product Configuration}: A web-based commercial
system uses an ASP-based product configuration
technology~\cite{Tiihonen-Practical-2003} (used by Variantum
Oy).

\item
\textit{Workforce Management}: An ASP-based system is developed to
build teams of employees to handle incoming ships by taking into
account a variety of requirements, e.g., skills, fairness,
regulations~\cite{RiccaGAMLIL2O12} (used by Gioia Tauro seaport).
\end{itemize}

Let us briefly explain the syntax and semantics of ASP programs and
describe how a computational problem can be solved in ASP.

\subsection{Programs} \label{sec:prog}

\paragraph*{Syntax} The input language of ASP programs are composed of three sets
namely \textit{constant symbols}, \textit{predicate symbols}, and
\textit{variable symbols} where intersection of constant symbols and
variable symbols is empty. The basic elements of the ASP programs
are \textit{atoms}. An atom $p(\vec{t})$ is composed of a predicate
symbol $p$ and \textit{terms} $\vec{t} = t_1, \dots , t_k$ where
each $t_i$ ($1 \leq i \leq k$) is either a constant or a variable. A
\textit{literal} is either an atom~$p(\vec{t})$ or its negated form
$not \ p(\vec{t})$.

An ASP program is  a finite set of \textit{rules} of the form:
\begin{equation}\label{eq_rule}
A \leftarrow A_1, \ldots, A_k, not \ A_{k+1}, \ldots, not \ A_m
\end{equation}
where  $m \geq k \geq 0$ and each $A_i$ is an atom; whereas, $A$ is
an atom or $\bot$.

For a rule $r$ of the form (\ref{eq_rule}), $A$ is called the
\textit{head} of the rule and denoted by $H(r)$. The conjunction of
the literals $A_1, \ldots, A_k, not\ A_{k+1}, \ldots, not\ A_{m}$ is
called the \textit{body} of $r$. The set $\{A_1,...,A_k\}$ of atoms
(called the positive part of the body) is denoted by~$B^+(r)$, and
the set $\{A_{k+1},...,A_m\}$ of atoms (called the negative part of
the body) is denoted by~$B^-(r)$, and all the atoms in the body are
denoted by $B(r) = B^+(r) \cup B^-(r)$.

We say that a rule $r$ is a \textit{fact} if $B(r) = \emptyset$, and
we usually omit the $\leftarrow$ sign. Furthermore, we say that a
rule $r$ is a \textit{constraint} if the head of $r$ is $\bot$, and
we usually omit the $\bot$ sign.

\paragraph*{Semantics (Answer Sets)} Answer sets of a program are defined over \textit{ground programs}.
We call an atom, rule, or program \textit{ground}, if it does not
contain any variables. Given a program~$\Pi$, the set ${\cal
U}_{\Pi}$ represents all the constants in $\Pi$, and the set ${\cal
B}_{\Pi}$ represents all the ground atoms that can be constructed
from atoms in $\Pi$ with constants in ${\cal U}_{\Pi}$. Also,
$Ground(\Pi)$ denotes the set of all the ground rules which are
obtained by substituting all variables in rules with the set of all
possible constants in ${\cal U}_{\Pi}$.

A subset $I$ of ${\cal B}_{\Pi}$ is called an
\textit{interpretation} for $\Pi$. A ground atom $p$ is true with
respect to an interpretation $I$ if $p \in I$; otherwise, it is
false. Similarly, a set $S$ of atoms is true (resp., false) with
respect to $I$ if each atom $p \in S$ is true (resp., false) with
respect to~$I$. An interpretation $I$ \textit{satisfies} a ground
rule $r$, if $H(r)$ is true with respect to $I$  whenever $B^+(r)$
is true and $B^-(r)$ is false with respect to $I$. An interpretation
$I$ is called a \textit{model} of a program $\Pi$ if it satisfies
all the rules in $\Pi$.

The \textit{reduct} $\Pi^I$ of a program $\Pi$ with respect to an
interpretation $I$ is defined as follows:
$$
\Pi^I = \{H(r) \leftarrow B^+(r) \ | \ r \in Ground(\Pi) \ s.t. \ I
\cap B^-(r) = \emptyset\}
$$

An interpretation $I$ is an \textit{answer set} for a program $\Pi$,
if it is a subset-minimal model for~$\Pi^I$, and $AS(\Pi)$ denotes
the set of all the answer sets of a program $\Pi$.

For example, consider the following program $\Pi_1$:
\begin{equation}
\label{eq:prog1} p \leftarrow not \ q
\end{equation}
\noindent and take an interpretation $I = \{p\}$. The reduct
$\Pi_1^I$ is as follows:
\begin{equation}
\label{eq:reduct} p
\end{equation}
\noindent The interpretation~$I$ is a model of the reduct~(\ref{eq:reduct}). Let us
take a strict subset $I'$ of $I$ which is $\emptyset$. Then,
the reduct~$\Pi_1^{I'}$ is again equal to~(\ref{eq:reduct}); however, $I'$ does
not satisfy~(\ref{eq:reduct}). Therefore, $I = \{p\}$ is a
subset-minimal model; hence an answer set of~$\Pi_1$. Note also that
$\{p\}$ is the only answer set of~$\Pi$.

\subsection{Generate-And-Test Representation Methodology with Special
ASP Constructs} \label{sec:generateTestASP}

The idea of ASP~\cite{Lifschitz08} is to represent a computational
problem as a program whose answer sets correspond to the solutions
of the problem, and to find the answer sets for that program using
an answer set solver.

When we represent a problem in ASP, two kinds of rules play an
important role: those that ``generate'' many answer sets
corresponding to ``possible solutions'', and those that can be used
to ``eliminate'' the answer sets that do not correspond to
solutions. The rules
\begin{equation} \label{eq:pi2}
\begin{array}{l}
p \leftarrow not \ q\\
q \leftarrow not \ p
 \end{array}
\end{equation}
are of the former kind: they generate the answer sets $\{p\}$
and~$\{q\}$. Constraints are of the latter kind. For instance,
adding the constraint
$$
\leftarrow p
$$
to program (\ref{eq:pi2}) eliminates the answer sets for the program
that contain~$p$.

In ASP, we use special constructs of the form \beq
 \{A_1,\dots,A_n\}^c
\eeq{choice} (called {\em choice expressions}), and of the form \beq
l \le \{A_1,\dots,A_m\} \le u \eeq{cardinality} (called {\em
cardinality expressions}) where each $A_i$ is an atom and $l$ and
$u$ are nonnegative integers denoting the ``lower bound'' and the
``upper bound''~\cite{Simons-Extending-2002}. Programs using these
constructs can be viewed as abbreviations for normal nested programs
defined in~\cite{Ferraris-Weight-2005}. Expression~(\ref{choice})
describes subsets of $\{A_1,\dots,A_n\}$. Such expressions can be
used in heads of rules to generate many answer sets. For instance,
the answer sets for the program
\beq \{p,q,r\}^c\ \lar\ \eeq{ex:p-q-r}
\begin{sloppypar}
\noindent are arbitrary subsets of~$\{p,q,r\}$.
Expression~(\ref{cardinality}) describes the subsets of the set
$\{A_1,\dots,A_m\}$ whose cardinalities are at least $l$ and at
most~$u$. Such expressions can be used in constraints to eliminate
some answer sets. For instance, adding the constraint
\end{sloppypar}
\vspace{-2ex}
$$
\lar\ 2 \le \{p,q,r\}
$$
to program (\ref{ex:p-q-r}) eliminates the answer sets for
(\ref{ex:p-q-r}) whose cardinalities are at least 2. We abbreviate
the rules
$$
\ba l
 \{A_1,\dots,A_m\}^c \lar \ii{Body} \\
\lar not\ (l \le \{A_1,\dots,A_m\}) \\
\lar not\ (\{A_1,\dots,A_m\} \le u) \ea
$$
by the rule
$$
l \le \{A_1,\dots,A_m\}^c \le u \lar \ii{Body} .
$$

In ASP, there are also special constructs that are useful for
optimization problems. For instance, to compute answer sets that
contain the maximum number of elements from the set $\{A_1, \ldots,
A_m\}$, we can use the following optimization statement.
$$
{\sc maximize} \langle\{A_1,\ldots,A_m\}\rangle
$$

\subsection{Presenting Programs to Answer Set Solvers}

Once we represent a computational problem as a program whose answer
sets correspond to the solutions of the problem,  we can use an
answer set solver to compute the solutions of the problem. To
present a program to an answer set solver, like \clasp, we need to
make some syntactic modifications.

Recall that answer sets for a program are defined over ground
programs. Thus, the input of ASP solvers should be ground
instantiations of the programs. For that, programs go through a
``grounding'' phase in which variables in the program (if exists)
are substituted by constants. For \clasp, we use the ``grounder''
\gringo~\cite{GebserKKS11}.

Although the syntax of the input language of \gringo\ is somewhat
more restricted than the class of programs defined above, it
provides a number of useful special constructs. For instance, the
head of a rule can be an expression of one of the forms
$$
\ba l \{A_1,\dots,A_n\}^c\\
    l\leq \{A_1,\dots,A_n\}^c\\
    \{A_1,\dots,A_n\}^c\leq u\\
    l\leq \{A_1,\dots,A_n\}^c\leq u
\ea
$$
but the superscript $^c$ and the sign $\leq$ are dropped. The body
can also contain cardinality expressions but the sign $\leq$ is
dropped. In the input language of \gringo, {\tt :-} stands for
$\lar$, and each rule is followed by a period. For facts $\lar$ is
dropped. For instance, the rule
$$
1 \le \{p,q,r\}^c \le 1 \lar
$$
can be presented to \gringo\ as follow:
{
\begin{verbatim}
1{p,q,r}1.
\end{verbatim}
}
Variables in a program are represented by strings whose initial
letters are capitalized. The constants and predicate symbols, on the
other hand, start with a lowercase letter. For instance, the program
$\Pi_n$
$$
p_i \lar \no\ p_{i+1}   \qquad (1 \leq i \leq n)
$$
can be presented to \gringo\ as follows:
{
\begin{verbatim}
index(1..n).
p(I) :- not p(I+1), index(I).
\end{verbatim}
}
\noindent Here, the auxiliary predicate {\tt index} is a ``domain
predicate'' used to describe the ranges of variables. Variables can
be also used ``locally'' to describe the list of formulas. For
instance, the rule
$$1\leq \{p_1,\dots,p_n\}\leq 1$$
can be expressed in \gringo\ as follows
{
\begin{verbatim}
index(1..n).
1{p(I) : index(I)}1.
\end{verbatim}
}


\section{Answering Biomedical Queries}
\label{sec:query}

We have earlier developed the software system
\bqasp~\cite{ErdemEO11} (see Figure~\ref{fig:bqasp}) to answer
complex queries that require appropriate integration of relevant
knowledge from different knowledge resources and auxiliary
definitions such as chains of drug-drug interactions, cliques of
genes based on gene-gene relations, or similar/diverse genes. As
depicted in Figure~\ref{fig:bqasp}, \bqasp\ takes a query in a
controlled natural language and transforms it into ASP. Meanwhile,
it extracts knowledge from biomedical databases and ontologies, and
integrates them in ASP. Afterwards, it computes an answer to the
given query using an ASP solver.

Let us give an example to illustrate these stages; the details of
representing biomedical queries in natural language and answering
them using ASP are explained in a companion article though.

First of all, let us mention that knowledge related to drug
discovery is extracted from the biomedical databases/ontologies and
represented in ASP. If the biomedical ontology is in RDF(S)/OWL then
we can extract such knowledge using the ASP solver
\dlvhex~\cite{dlvhex} by making use of external predicates.
For instance, consider as an external theory a Drug Ontology
described in RDF. All triples from this theory can be exported using
the external predicate {\tt \&rdf}:

{\begin{verbatim}
triple_drug(X,Y,Z) :- &rdf["URI for Drug Ontology"](X,Y,Z).
\end{verbatim}}

\noindent Then the names of drugs can be extracted by \dlvhex\ using
the rule:

{\begin{verbatim}
drug_name(A) :- triple_drug(_,"drugproperties:name",A).
\end{verbatim}}

Some knowledge resources are provided as relational databases, or
more often as a set of triples (probably extracted from ontologies
in RDF). In such cases, we use short scripts to transform
the relations into ASP.

To relate the knowledge extracted from the biomedical databases or
ontologies and also provide auxiliary definitions, a rule layer is
constructed in ASP. For instance, drugs targeting genes are described
by the relation {\tt drug\_gene} defined in the rule layer as follows:
{\begin{verbatim}
drug_gene(D,G) :- drug_gene_pharmgkb(D,G).
drug_gene(D,G) :- drug_gene_ctd(D,G).
\end{verbatim}}

\noindent where {\tt drug\_gene\_pharmgkb} and {\tt drug\_gene\_ctd}
are relations for extracting knowledge from relevant knowledge resources.
The auxiliary concept of reachability of a gene from another gene by means
of a chain of gene-gene interactions is defined in the rule layer as well:
{\begin{verbatim}
gene_reachable_from(X,1) :- gene_gene(X,Y), start_gene(Y).
gene_reachable_from(X,N+1) :- gene_gene(X,Z),
   gene_reachable_from(Z,N), 0 < N, N < L,
   max_chain_length(L).
\end{verbatim}}

Now, consider, for instance, the query~Q11 from Table~\ref{tab:queries}.
\bi
\item[Q11] What are the drugs that treat the disease Depression and that do not target the gene~ACYP1?
\ei
This type of queries might be important in terms of drug
repurposing~\cite{ChongS07} which has achieved a number of successes
in drug development, including the famous example of Pfizer's
Viagra~\cite{Gower09}.

This query is then translated into the following program in the
language of \gringo:
{
\begin{verbatim}
what_be_drugs(DRG) :-  cond1(DRG), cond2(DRG).
cond1(DRG) :- drug_disease(DRG,"Depression").
cond2(DRG) :- drug_name(DRG), not drug_gene(DRG,"ACYP1").
answer_exists :- what_be_drugs(DRG).
:- not answer_exists.
\end{verbatim}
}

\noindent where {\tt cond1} and {\tt cond2} are invented relations,
{\tt drug\_name}, {\tt drug\_disease} and {\tt drug\_gene} are
defined in the rule layer.

Once the query  and the rule layer are in ASP,
the parts of the rule layer that are relevant to the given query
are identified by an algorithm~\cite{ErdemEEO11}. For some queries,
the relevant part of the program is almost 100 times smaller than
the whole program (considering the number of ground rules).

Then, given the query as an ASP program and the relevant knowledge
as an ASP program, we can find answers to the query by computing an
answer set for the union of these two programs using \clasp. For the
query above an answer computed in this way is ``Fluoxetine''.


\section{Explaining an Answer for a Query}
\label{sec:exp}

Once an answer is found for a complex biomedical query, the experts
may need informative explanations about the answer, as discussed in
the introduction. With this motivation, we study generating explanations for
complex biomedical queries. Since the queries, knowledge extracted
from databases and ontologies, and the rule layer are in ASP, our
studies focus on explanation generation within the context of ASP.

Before we introduce our methods to generate explanations for a given query,
let us introduce some definitions regarding explanations in ASP.

Let $\Pi$ be the relevant part of a ground ASP program with respect
to a given biomedical query $Q$ (also a ground ASP program), that
contains rules describing the knowledge extracted from biomedical
ontologies and databases, the knowledge integrating them, and the
background knowledge. Rules in $\Pi \cup Q$ generally do not contain
cardinality/choice expressions in the head; therefore, we assume
that in $\Pi \cup Q$ only bodies of rules contain cardinality
expressions. Let $X$ be an answer set for $\Pi \cup Q$. Let $p$ be
an atom that characterizes an answer to the query $Q$. The goal is
to find an ``explanation'' as to why $p$ is computed as an answer to
the query $Q$, i.e., why is $p$ in $X$? Before we introduce a
definition of an explanation, we need the following notations and
definitions.

We say that a set~$X$ of atoms {\em satisfies} a cardinality
expression $C$ of the form
$$
l \leq \{A_1, \ldots, A_m\} \leq u
$$
if the cardinality of $X \cap \{A_1, \ldots, A_m\}$ is within the
lower bound $l$ and upper bound $u$. Also $X$ {\em satisfies} a set
$SC$ of cardinality expressions (denoted by $X\models SC$), if $X$
satisfies every element of $SC$.

Let $\Pi$ be a ground ASP program, $r$ be a rule in $\Pi$, $p$ be an
atom in $\Pi$, and $Y$ and $Z$ be two sets of atoms. Let
$B_{card}(r)$ denote the set of cardinality expressions that appear
in the body of $r$.
We say that $r$ {\em supports} an atom $p$ using atoms in $Y$ but
not in $Z$ (or with respect to $Y$ but $Z$), if the following hold:
\begin{equation}
\begin{array}{l}
\label{eq:support}
H(r) = p, \\
B^+(r) \subseteq Y \backslash Z, \\
B^-(r) \, \cap \, Y = \emptyset, \\
Y \models B_{card}(r)
\end{array}
\end{equation}
We denote the set of rules in $\Pi$ that support $p$ with respect to
$Y$ but $Z$, by $\Pi_{Y,Z}(p)$.

We now introduce definitions about explanations in ASP. We first
define a generic tree whose vertices are labeled by either atoms or
rules.

\begin{definition}[Vertex-labeled tree]
A \emph{vertex-labeled tree} $\seq{V,E,l,\Pi,X}$ for a program $\Pi$
and a set $X$ of atoms is a tree $\seq{V,E}$ whose vertices are
labeled by a function $l$ that maps $V$ to $\Pi \cup X$. In this
tree, the vertices labeled by an atom (resp., a rule) are called
\emph{atom vertices} (resp., \emph{rule vertices}).
\end{definition}

For a vertex-labeled tree $T = \seq{V,E,l,\Pi,X}$ and a vertex $v$
in $V$, we introduce the following notations:
\begin{itemize}
\item $\ii{anc}_T(v)$ denotes the set of atoms which are labels of ancestors of $v$.
\item $\ii{des}_T(v)$ denotes the set of rule vertices which are descendants of $v$.
\item $\ii{child}_E(v)$ denotes the set of children of $v$.
\item $\ii{sibling}_E(v)$ denotes the set of siblings of $v$.
\item $\ii{out}_E(v)$ denotes the set of out-going edges of $v$.
\item $\ii{deg}_E(v)$  denotes the degree of $v$ and equals to $|\ii{out}_E(v)|$.
\item If $\ii{deg}_E(v) = 0$, then $v$ is a \emph{leaf} vertex.
\item $\ii{leaf}(T)$ denotes the set of leaves of $T$.
\item The \emph{root} of $T$ is the root of $\seq{V,E}$.
\item $T$ is \emph{empty} if $\seq{V,E} = \seq{\emptyset,\emptyset}$.
\end{itemize}

We now define a specific class of vertex-labeled trees which
contains all possible ``explanations'' for an atom.
\begin{definition}[And-or explanation tree]
\label{def:andor} Let $\Pi$ be a ground ASP program, $X$ be an
answer set for $\Pi$, $p$ be an atom in $X$. The \emph{and-or
explanation tree} for $p$ with respect to $\Pi$ and $X$ is a
vertex-labeled tree $T = \seq{V, E, l, \Pi, X}$ that satisfies the
following:
\begin{itemize}
\label{expTree}
\item[{\it (i)}] for the root $v \in V$ of the tree, $l(v) = p$;
\item[{\it (ii)}] for every atom vertex $v \in V$,
$$
\ii{out}_E(v) = \{(v,v^{\prime}) \sep (v,v^{\prime}) \in E, \,
l(v^{\prime}) \in \Pi_{X,{\scriptsize
\ii{anc}_T(v^{\prime}})}(l(v))\};$$
\item[{\it (iii)}] for every rule vertex $v \in V$,
$$
\ii{out}_E(v) = \{(v, v^{\prime}) \sep (v,v^{\prime}) \in E, \,
l(v^{\prime}) \in B^+(l(v))\};$$
\item[{\it (iv)}] each leaf vertex is a rule vertex.
\end{itemize}
\end{definition}
Let us explain conditions $(i)-(iv)$ in Definition \ref{def:andor}
in detail.
\begin{enumerate}[{\it (i)}]
\item The root of the and-or explanation tree $T$ is labeled by the atom $p$. Intuitively,
$T$ contains all possible explanations for $p$.
\item For every atom vertex $v \in V$, there is an out-going edge $(v,v^{\prime})$ to a rule vertex
$v^{\prime} \in V$ under the following conditions: the rule that
labels $v^{\prime}$ supports the atom that labels $v$, using atoms
in $X$ but not any atom that labels an ancestor of $v^{\prime}$. We
want to exclude the atoms labeling ancestors of $v^{\prime}$ to
ensure that the height of the and-or explanation tree is finite
(e.g., otherwise, due to cyclic dependencies the tree may be
infinite).
\item For every rule vertex $v \in V$, there is an out-going edge $(v,v^{\prime})$ to an atom vertex if
the atom that labels $v^{\prime}$ is in the positive body of the
rule that labels $v$. In this way, we make sure that every atom in
the positive body of the rule that labels $v$ takes part in
explaining the head of the rule that labels $v$.
\item Together with Conditions $(ii)$ and $(iii)$ above, this condition guarantees that the leaves of the and-or explanation tree
are rule vertices that are labeled by facts in the reduct of the
given ASP program $\Pi$ with respect to the given answer set $X$.
Intuitively, this condition expresses that the leaves are
self-explanatory.
\end{enumerate}

\begin{figure}[htbp]
\begin{center}
\begin{tikzpicture}[level distance = 10mm, sibling distance = 25mm]
\node{$a$}[->]
  child
  {
    node{$a \lar b,c$}
    child
    {
      node{$b$}
      child
      {
        node{$b \lar c$}
    child
    {
      node{$c$}
      child
      {
        node{$c \lar$}
      }
        }
      }
    }
    child
    {
      node{$c$}
      child
      {
        node{$c \lar$}
      }
    }
  }
  child
  {
    node{$a \lar d$}
    child
    {
      node{$d$}
      child
      {
        node{$d \lar$}
      }
    }
  };

\end{tikzpicture}
\end{center}
\caption{The and-or explanation tree for Example \ref{ex:andor}.}
\label{fig:andor}

\end{figure}
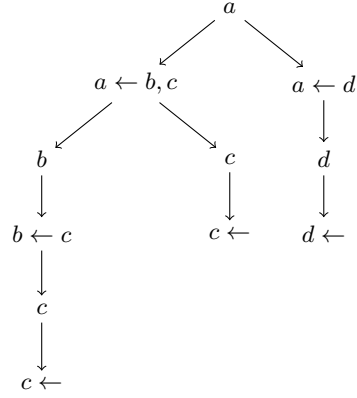

\begin{example}
\label{ex:andor} Let $\Pi$ be the program
$$
\begin{array}{l}
a \lar b,c \\
a \lar d \\
d \lar \\
b \lar c \\
c \lar
\end{array}
$$
and $X =~\{a,b,c,d\}$. The and-or explanation tree for $a$ with
respect to $\Pi$ and $X$ is shown in Figure~\ref{fig:andor}.
Intuitively, the and-or explanation tree includes all possible
``explanations'' for an atom. For instance, according to
Figure~\ref{fig:andor}, the atom $a$ has two explanations:
\begin{itemize}
\item One explanation is characterized by the rules that label the vertices in the left
subtree of the root: $a$ is in $X$ because the rule
$$
a \lar b,c
$$
support $a$. Moreover, this rule can be ``applied to generate $a$''
because $b$ and $c$, the atoms in its positive body, are in $X$.
Further, $b$ is in $X$ because the rule
$$
b \lar c
$$
supports $b$. Further, $c$ is in $X$ because $c$ is supported by the
rule
$$
c \lar
$$
which is self-explanatory.
\item The other explanation is characterized by the rules that label the vertices in the
right subtree of the root: $a$ is in $X$ because the rule
$$
a \lar d
$$
supports $a$. Further, this rule can be ``applied to generate $a$''
because $d$ is in $X$. In addition, $d$ is in $X$ because $d$ is
supported by the rule
$$
d \lar
$$
which is self-explanatory.
\end{itemize}
\end{example}

\begin{proposition}
\label{prop:andorNonEmpty} Let $\Pi$ be a ground ASP program and $X$
be an answer set for $\Pi$. For every $p$ in $X$, the and-or
explanation tree for $p$ with respect to $\Pi$ and $X$ is not empty.
\end{proposition}

\begin{figure}[htbp]
\begin{minipage}[b]{0.5\linewidth}
\centering
\begin{tikzpicture}[level distance = 10mm, sibling distance = 20mm]
\node{$a$}[->]
  child
  {
    node{$a \lar b,c$}
    child
    {
      node{$b$}
      child
      {
        node{$b \lar c$}
    child
    {
      node{$c$}
      child
      {
        node{$c \lar$}
      }
        }
      }
    }
    child
    {
      node{$c$}
      child
      {
        node{$c \lar$}
      }
    }
  };
\end{tikzpicture}
\end{minipage}%
\begin{minipage}[b]{0.5\linewidth}
\centering
\begin{tikzpicture}[level distance = 10mm, sibling distance = 20mm]
\node{$a$}[->]
 child
  {
    node{$a \lar d$}
    child
    {
      node{$d$}
      child
      {
        node{$d \lar$}
      }
    }
  };
\end{tikzpicture}
\end{minipage}
\caption{Explanation trees for Example \ref{ex:extree}.}
\label{fig:exptree}
\end{figure}
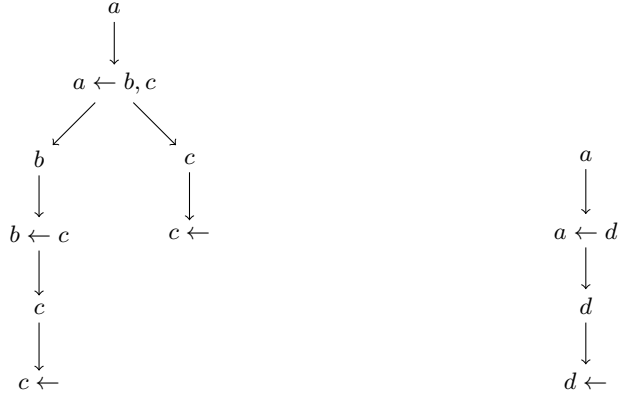

Note that in the and-or explanation tree, atom vertices are the
``or'' vertices, and rule vertices are the ``and'' vertices. Then,
we can obtain a subtree of the and-or explanation tree that contains
an explanation, by visiting only one child of every atom vertex and
every child of every rule vertex, starting from the root of the
and-or explanation tree. Here is precise definition of such a
subtree, called an explanation tree.
\begin{definition}[Explanation tree]
\label{def:expTree} Let $\Pi$ be a ground ASP program, $X$ be an
answer set for $\Pi$, $p$ be an atom in $X$, and $T =
\seq{V,E,l,\Pi,X}$ be the and-or explanation tree for $p$ with
respect to $\Pi$ and $X$. An \emph{explanation tree} in $T$ is a
vertex-labeled tree \hbox{$T^{\prime} =
\seq{V^{\prime},E^{\prime},l,\Pi,X}$} such that
\begin{itemize}
\item[{\it (i)}] $\seq{V^{\prime},E^{\prime}}$ is a subtree of $\seq{V,E}$;
\item[{\it (ii)}] the root of $\seq{V^{\prime},E^{\prime}}$ is the root of $\seq{V,E}$;
\item[{\it (iii)}] for every atom vertex $v^{\prime} \in V^{\prime}$, $\ii{deg}_{E^{\prime}}(v^{\prime}) = 1$;
\item[{\it (iv)}] for every rule vertex $v^{\prime} \in V^{\prime}$,  $\ii{out}_{E}(v^{\prime}) \subseteq E^{\prime}$.
\end{itemize}
\end{definition}
\begin{example}
\label{ex:extree} Let $T$ be the and-or explanation tree in Figure
\ref{fig:andor}. Then, Figure \ref{fig:exptree} illustrates the
explanation trees in $T$. These explanation trees characterize the
two explanations for $a$ explained in Example~\ref{ex:andor}.
\end{example}

After having defined the and-or explanation tree and an explanation
tree for an atom, let us now define an explanation for an atom.
\begin{definition}[Explanation]
\label{def:explanation} Let $\Pi$ be a ground ASP program, $X$ be an
answer set for~$\Pi$, and $p$ be an atom in~$X$. A vertex-labeled
tree $\seq{V^{\prime}, E^{\prime},l,\Pi,X}$ is an \emph{explanation}
for $p$ with respect to $\Pi$ and $X$ if there exists an explanation
tree $\seq{V,E,l,\Pi,X}$ in the and-or explanation tree for $p$ with
respect to $\Pi$ and $X$ such that
\begin{itemize}
\item[{\it (i)}] $V^{\prime} = \{v \sep v \, \mbox{is a rule vertex in} \, V\}$;
\item[{\it (ii)}] $E^{\prime} = \{(v_1, v_2) \sep (v_1, v), (v, v_2) \in E, \, \mbox{for some atom vertex} \, v \in V\}$.
\end{itemize}
\end{definition}

Intuitively, an explanation can be obtained from an explanation tree
by ``ignoring'' its atom vertices.

\begin{figure}[htbp]
\subfloat[]{
\label{fig:exp:a}
\begin{minipage}[b]{0.5\linewidth}
\centering
\begin{tikzpicture}[level distance = 10mm, sibling distance = 20mm]
\node{$a \lar b,c$}[->]
  child
  {
    node{$b \lar c$}
    child
    {
      node{$c \lar$}
    }
  }
  child
  {
    node{$c \lar$}
  };
\end{tikzpicture}
\end{minipage}
}%
\subfloat[]{
\label{fig:exp:b}
\begin{minipage}[b]{0.5\linewidth}
\centering
\begin{tikzpicture}[level distance = 10mm, sibling distance = 20mm]
\node{$a \lar d$}[->]
  child
  {
    node{$d \lar$}
  };
\end{tikzpicture}
\end{minipage}
}
\caption{Explanations for Example \ref{ex:ex}.}
\label{fig:exp}
\end{figure}
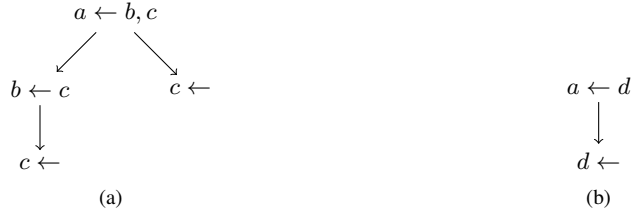

\begin{example}
\label{ex:ex} Let $\Pi$ and $X$ be defined as in Example
\ref{ex:andor}. Then, Figure \ref{fig:exp} depicts two explanations
for $a$ with respect to $\Pi$ and $X$, described in
Example~\ref{ex:andor}.
\end{example}


\begin{figure}[htbp]
\subfloat[]{
\begin{minipage}[t]{0.5\linewidth}
\centering
\begin{tikzpicture}[level distance = 10mm, level 1/.style={sibling distance = 40mm},
                                       level 2/.style={sibling distance = 25mm}]
\node{$a$}[->]
  child
  {
    node{$a \lar b,c, not\ e$}
    child
    {
      node{$b$}
      child
      {
        node{$b \lar c$}
    child
    {
      node{$c$}
      child
      {
        node{$c \lar$}
      }
        }
      }
    }
    child
    {
      node{$c$}
      child
      {
        node{$c \lar$}
      }
    }
  }
  child
  {
    node{$a \lar d, 1 \leq\{b,c\} \leq 2$}
    child
    {
      node{$d$}
      child
      {
        node{$d \lar$}
      }
    }
  };
\end{tikzpicture}
\end{minipage}
}
\subfloat[]{
\begin{minipage}[t]{0.5\linewidth}
\centering
\begin{tikzpicture}[level distance = 10mm, sibling distance = 20mm]
\node{$a \lar d, 1\leq \{b,c\} \leq 2$}[->]
  child
  {
    node{$d \lar$}
  };
\end{tikzpicture}
\end{minipage}
}
\caption{(a) The and-or explanation tree for $a$ and (b) an explanation for $a$.}
\label{fig:explanationNegAggr}
\end{figure}
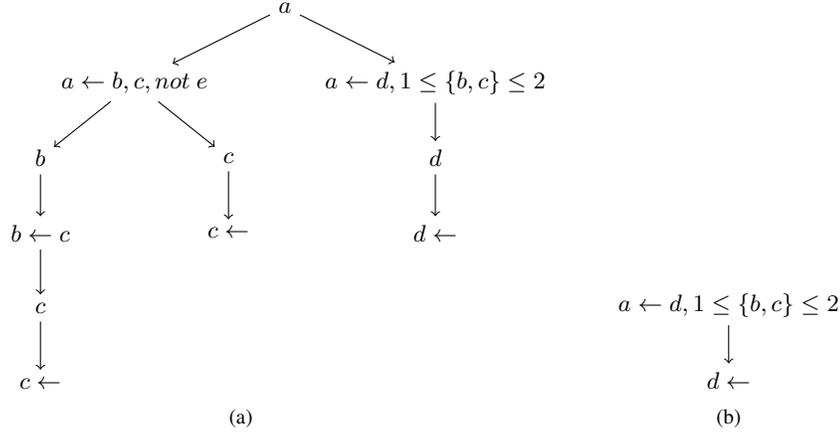

So far, we have considered only positive programs in the examples.
Our definitions can also be used in programs that contain negation
and aggregates in the bodies of rules.
\begin{example}
\label{ex:explanationNegAggr} Let $\Pi$ be the program
$$
\begin{array}{l}
a \lar b,c, not\ e \\
a \lar d, not\ b \\
a \lar d, 1 \leq \{b,c\} \leq 2 \\
d \lar \\
b \lar c \\
c \lar
\end{array}
$$
and $X =~\{a,b,c,d\}$. The and-or explanation tree for $a$ with
respect to $\Pi$ and $X$ is shown in
Figure~\ref{fig:explanationNegAggr}(a). Here, the rule $a \lar d,
not\ b$ is not included in the tree as $b$ is in $X$, whereas the
rule $a \lar b,c, not\ e$ is in the tree as $e$ is not in $X$ and,
$b$ and $c$ are in $X$. Also, the rule $a \lar d, 1 \leq \{b,c\}
\leq 2$ is in the tree as $d$ is in $X$ and the cardinality
expression $1 \leq \{b,c\} \leq 2$ is satisfied by $X$. An
explanation for $a$ with respect to $\Pi$ and $X$ is shown in
Figure~\ref{fig:explanationNegAggr}(b).
\end{example}


\noindent {
Note that our definition of an and-or explanation tree
considers positive body parts of the rules only to provide explanations.
Therefore, explanation trees do not provide further explanations
for negated literals (e.g., why an atom is not included in the answer set),
or aggregates (e.g., why a cardinality constraint is satisfied)
as seen in the example above.
}


\section{Generating Shortest Explanations}
\label{sec:shortExp}

\begin{figure}[!t]
\begin{center}
{\footnotesize \hspace*{-4.1cm}
\begin{tikzpicture}[level 1/.style={level distance = 17mm},
                    level 2/.style={level distance = 17mm, sibling distance = 55mm},
                    level 3/.style={level distance = 17mm, sibling distance = 47mm}]
\node{$\ii{what\_be\_genes("CASK")} \lar
\ii{gene\_reachable\_from("CASK",3)}$} [->]
  child
  {
    node {
          \begin{tabular}{l}
          $\ii{gene\_reachable\_from("CASK",3)} \lar$ \\
          \hspace{5pt} $\ii{gene\_gene("CASK","DLG1"),}$\\
          \hspace{5pt} $\ii{gene\_reachable\_from("DLG1",2)},\ldots$
          \end{tabular}
          }
    child
    {
      node
      {
        \begin{tabular}{l}
          $\ii{gene\_gene("CASK","DLG1")} \lar$ \\
          \hspace{5pt} $\ii{gene\_gene\_biogrid("CASK","DLG1")}$
        \end{tabular}
      }
      child
      {
        node
        { $\vdots$
        }
      }
    }
   child
   {
     node
     {
        \begin{tabular}{l}
          $\ii{gene\_reachable\_from("DLG1",2)} \lar$ \\
          \hspace{5pt} $\ii{gene\_gene("DLG1","DLG4")}, $ \\
          \hspace{5pt} $\ii{gene\_reachable\_from("DLG4",1)},
          \ldots$\\
        \end{tabular}
     }
     child
     {
        node
        {
          \begin{tabular}{l}
          $\ii{gene\_gene("DLG1","DLG4")} \lar \ldots$ \\
          \end{tabular}
         }
         child
         {
           node {$\vdots$}
         }
      }
      child
      {
         node
         {
            \begin{tabular}{l}
             $\ii{gene\_reachable\_from("DLG4",1)} \lar \ldots$ \\
             \end{tabular}
         }
         child
         {  node{$\vdots$}
         }
      }
   }
};
\end{tikzpicture}
}
\end{center}
\caption{Another explanation for Q8.}
\label{fig:expQ8_3}
\end{figure}
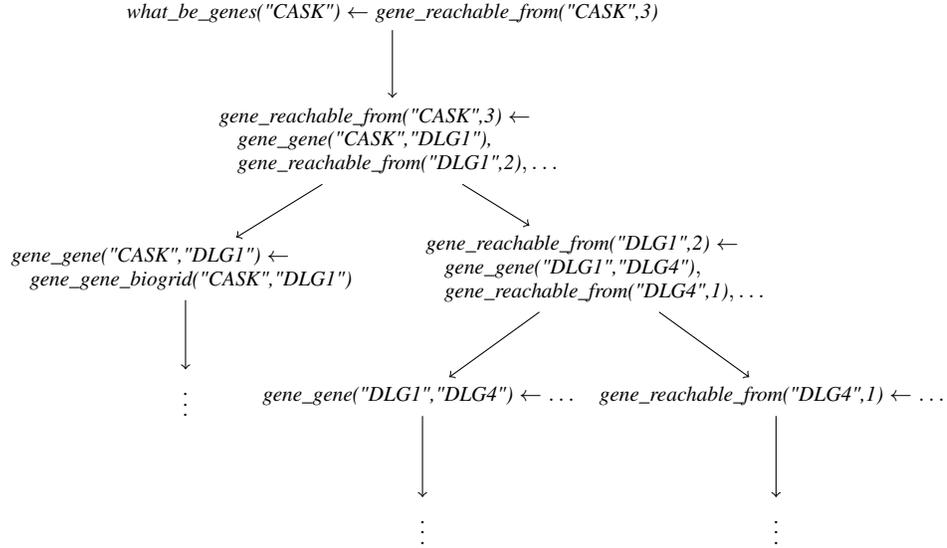

As can be seen in Figure~\ref{fig:exp}, there might be more than one
explanation for a given atom. Hence, it is not surprising that one
may prefer some explanations to others. Consider biomedical queries
about chains of gene-gene interactions like the query Q8 in
Table~\ref{tab:queries}. Answers of such queries may contain chains
of gene-gene interactions with different lengths. For instance, an
answer for this query is ``CASK''. Figure~\ref{fig:expQ8} shows
an explanation for this answer. Here, ``CASK'' is related to
``ADRB1'' via a gene-gene chain interaction of length $2$ (the chain
``CASK''--``DLG4''--``ADRB1''). Another explanation is partly shown in
Figure~\ref{fig:expQ8_3}. Now, ``CASK'' is related to ``ADRB1''
via a gene-gene chain interaction of length $3$ (the chain
``CASK''--`` DLG1''--``DLG4''--``ADRB1'').
Since gene-gene
interactions are important for drug discovery, it may be more
desirable for the experts to reason about chains with shortest
lengths.

With this motivation, we consider generating shortest
explanations. Intuitively, an explanation~$S$ is shorter than
another explanation~$S'$ if the number of rule vertices involved
in~$S$ is less than the number of rule vertices involved in~$S'$.
Then we can define shortest explanations as follows.
\begin{figure}[!t]
\begin{center}
{\footnotesize
\hspace*{-4.1cm}
\begin{tikzpicture}[level 1/.style={level distance = 17mm},
                    level 2/.style={level distance = 17mm, sibling distance = 70mm},
                    level 3/.style={level distance = 17mm, sibling distance = 40mm}]
\node{$\ii{what\_be\_genes("CASK")} \lar \ii{gene\_reachable\_from("CASK",2)}$} [->]
  child
  {
    node {
        \begin{tabular}{l}
         $\ii{gene\_reachable\_from("CASK",2)} \lar$\\
         \hspace{5pt} $\ii{gene\_gene("CASK","DLG4"),}$\\
         \hspace{5pt} $\ii{gene\_reachable\_from("DLG4",1)},\ldots$
         \end{tabular}
         }
    child
    {
      node
      {
        \begin{tabular}{l}
          $\ii{gene\_gene("CASK","DLG4")} \lar$ \\
        \hspace{5pt} $\ii{gene\_gene\_biogrid("CASK","DLG4")}$
        \end{tabular}
      }
      child
      {
        node
    {
          \begin{tabular}{l}
        $\ii{gene\_gene\_biogrid("CASK","DLG4")}$\\
        \textcolor{white}{dummy}
      \end{tabular}
    }
      }
    }
    child
    {
      node
      {
        \begin{tabular}{l}
          $\ii{gene\_reachable\_from("DLG4",1)} \lar$ \\
          \hspace{5pt} $\ii{gene\_gene("DLG4","ADRB1"),}$\\
          \hspace{5pt} $\ii{start\_gene("ADRB1")}$
        \end{tabular}
      }
      child
      {
        node
        {
          \begin{tabular}{l}
            $\ii{gene\_gene("DLG4","ADRB1")} \lar$ \\
        \hspace{5pt} $\ii{gene\_gene\_biogrid("DLG4","ADRB1")}$
          \end{tabular}
        }
        child
        {
          node {$\ii{gene\_gene\_biogrid("DLG4","ADRB1")}$}
        }
      }
      child
      {
        node
    {
          \begin{tabular}{l}
            $\ii{start\_gene("ADRB1")}$\\
            \textcolor{white}{dummy}
          \end{tabular}
        }
      }
    }
  };
\end{tikzpicture}
}
\end{center}
\caption{A shortest explanation for Q8.}
\label{fig:expQ8}
\end{figure}
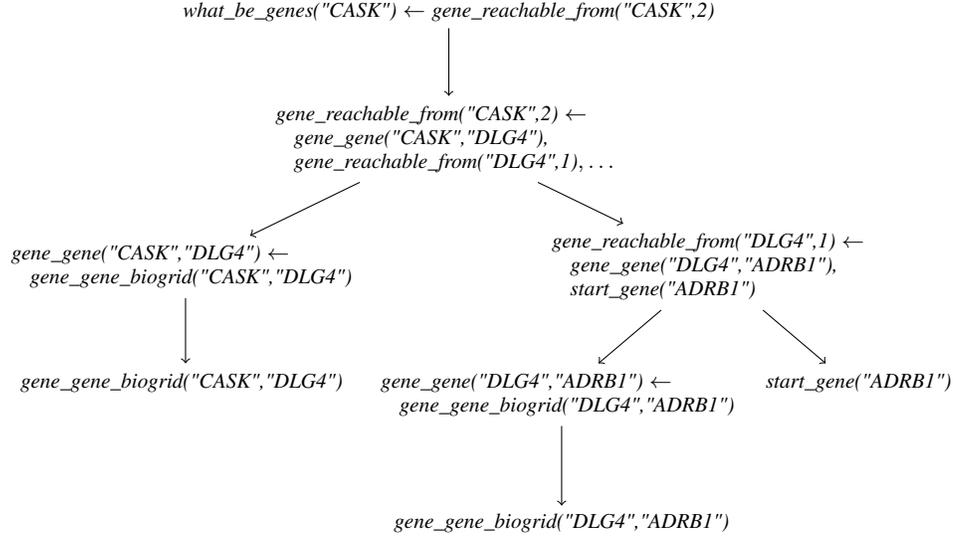

\begin{definition}[Shortest explanation]
Let $\Pi$ be a ground ASP program, $X$ be an answer set for $\Pi$,
$p$ be an atom in $X$, and $S$ be an explanation (with vertices $V$)
for $p$ with respect to $\Pi$ and $X$. Then, $S$ is a \emph{shortest
explanation} for $p$ with respect to $\Pi$ and $X$ if there exists
no explanation $S^{\prime}$ (with vertices $V^{\prime}$) for $p$
with respect to $\Pi$ and $X$ such that $|V^{\prime}| < |V|$.
\end{definition}

\begin{example}
Let $\Pi$ and $X$ be defined as in Example \ref{ex:andor}. Then,
Figure \ref{fig:exp}(b) is the shortest explanation for $a$ with
respect to $\Pi$ and $X$.
\end{example}

To compute shortest explanations, we define a weight function that
assigns weights to the vertices of the and-or explanation tree.
Basically, the weight of an atom vertex (``or'' vertex) is equal to
the minimum weight among weights of its children and the weight of a
rule vertex (``and'' vertex) is equal to sum of weights of its
children plus $1$. Then the idea is to extract a shortest
explanation by propagating the weights of the leaves up and then
traversing the vertices that contribute to the weight of the root.
Let us define the weight of vertices in the and-or explanation tree.

\begin{definition}[Weight function]
\label{def:weight} Let $\Pi$ be a ground ASP program, $X$ be an
answer set for $\Pi$, $p$ be an atom in $X$, and $T =
\seq{V,E,l,\Pi,X}$ be the and-or explanation tree for $p$ with
respect to $\Pi$ and $X$. The weight function $W_T$ for $T$ maps
vertices in $V$ to a positive integer and it is defined as follows.
\[ W_T(v) =
\left\{
\begin{array} {ll}
\mbox{min}\{W_T(c) \sep c \in \ii{child}_E(v)\} & \mbox{if v is an atom vertex in $V$}; \\
1 + \sum_{\mbox{{\scriptsize $c \in \ii{child}_E(v)$}}} W_T(c) &
\mbox{otherwise}.
\end{array}
\right. \]
\end{definition}

\begin{algorithm}[h]
\caption{Generating Shortest
Explanations} \label{alg:shortExp} \KwIn{$\Pi:$ ground ASP program,
$X:$ answer set for $\Pi$, $p:$ atom in $X$.} \KwOut{a shortest
explanation for $p$ w.r.t $\Pi$ and $X$, or an empty vertex-labeled
tree.} $\seq{V,E,l,\Pi,X} := \mbox{createTree}(\Pi, X, p, \{\})$\;
\If{$\seq{V,E}$ is not empty} {
  $v \lar $ root of $\seq{V,E}$\;
  $\mbox{calculateWeight}(\Pi, X, V, l, v, E,  W_T)$\;
  $\seq{V^{\prime},E^{\prime},l,\Pi,X} := \mbox{extractExp}(\Pi, X, V, l, v, E, W_T, \emptyset, \ii{min})$\;
  \KwRet{$\seq{V^{\prime},E^{\prime},l,\Pi,X}$}\;
} \Else { \KwRet{$\seq{\emptyset,\emptyset,l,\Pi,X}$}\; }
\end{algorithm}

Using this weight function, we develop Algorithm~\ref{alg:shortExp} to generate shortest
explanations. Let us describe this algorithm.
Algorithm~\ref{alg:shortExp} starts by creating the and-or
explanation tree $T$ for $p$ with respect to $\Pi$ and $X$
(Line~$1$); for that it uses Algorithm~\ref{alg:createTree}. If $T$
is not empty, then Algorithm~\ref{alg:shortExp} assigns weights to
the vertices of $T$ (Line~$4$), using
Algorithm~\ref{alg:calculateWeight}. As the final step,
Algorithm~\ref{alg:shortExp} extracts a shortest explanation from
$T$ (Line~$5$), using Algorithm~\ref{alg:extractExp}. The idea is to
traverse an explanation tree of $T$, by the help of the weight
function, and construct an explanation, which would be a shortest
one, by contemplating only the rule vertices in the traversed
explanation tree. If $T$ is empty, Algorithm~\ref{alg:shortExp}
returns an empty vertex-labeled tree.

{
Algorithm~\ref{alg:createTree} (with the call $\mbox{createTree}(\Pi, X, p, \{\})$)
creates the and-or explanation tree for $p$ with respect to $\Pi$ and $X$ recursively.
With a call $\mbox{createTree}(\Pi, X, d, L)$,
where $L$ intuitively denotes the atoms labeling the atom vertices created so far,
the algorithm considers two cases: $d$ being an atom or a rule. In the former case,
1) the algorithm creates an atom vertex $v$ for $d$, 2) it identifies the rules
that support $d$, 3) for each such rule, it creates a vertex labeled tree
(i.e., a subtree of the resulting and-or explanation tree), and
4) it connects these trees to the atom vertex $v$.
In the latter case, if $d$ is a rule in $\Pi$, 1) the algorithm creates
a rule vertex $v$ for $d$,
2) it identifies the atoms in the positive part of the rule, 3) it creates
the and-or explanation tree for each such atom, and 4) it connects these
 trees to the rule vertex $v$.

Once the and-or explanation tree is created, Algorithm~\ref{alg:calculateWeight}
assigns weights to all vertices in the tree by propagating the weights of
the leaves (i.e., 1) up to the root in a bottom-up fashion using the weight
function definition (i.e., Definition~\ref{def:weight}).

After that, Algorithm~\ref{alg:shortExp}
(with the call $\mbox{extractExp}(\Pi, X, V, l, v, E, W_T, \emptyset, \ii{min})$)
extracts a shortest explanation
in a top-down fashion starting from the root by examining the weights of
the vertices. In particular, if a visited vertex $v$ is an atom vertex
then the algorithm proceeds with the child of $v$ with the minimum weight;
otherwise, it considers all the children of $v$.
}

\begin{algorithm}[t!]
\caption{createTree}
\label{alg:createTree} \KwIn{$\Pi:$ ground ASP program, $X:$ answer
set for $\Pi$, $d:$ an atom in $X$ or a rule in $\Pi$, $L:$ set of
atoms in $X$.} \KwOut{A vertex-labeled tree.} $V := \emptyset, \ E
:= \emptyset$ \; \If{$d \in X \backslash L$} {
  $v \lar $ Create an atom vertex s.t. $l(v) = d$ \;
  $L := L \cup \{d\}$, \, $V := V \cup \{v\}$ \;
  \ForEach{$r \in \Pi_{X,L}(d)$}
  {
    $\seq{V^{\prime}, E^{\prime}, l, \Pi, X} := \mbox{createTree}(\Pi, X, r, L)$\;
    \If{$\seq{V^{\prime}, E^{\prime}} \neq \seq{\emptyset, \emptyset}$}
    {
      $v^{\prime} \lar $ root of $\seq{V^{\prime},E^{\prime}}$ s.t. $l(v^{\prime}) = r$\;
      $V := V \cup V^{\prime}$, \, $E := E \cup \{(v,v^{\prime})\} \cup E^{\prime}$ \;
    }
  }
  \lIf{$E = \emptyset$}
  {
    \KwRet{$\seq{\emptyset, \emptyset, l, \Pi, X}$}\;
  }
} \ElseIf{$d \in \Pi$} {
  $v \lar $ Create a rule vertex s.t. $l(v) = d$ \;
  \ForEach{$a \in B^+(d)$}
  {
    $\seq{V^{\prime}, E^{\prime},l,\Pi,X} := \mbox{createTree}(\Pi, X, a, L)$\;
    \lIf{$\seq{V^{\prime},E^{\prime}} =  \seq{\emptyset,\emptyset}$}
    {
      \KwRet{$\seq{\emptyset,\emptyset, l, \Pi, X}$}\;
    }
    $v^{\prime} \lar$ root of $\seq{V^{\prime},E^{\prime}}$ s.t. $l(v^{\prime}) = a$\;
    $V := V \cup V^{\prime}$,\, $E := E \cup \{(v,v^{\prime})\} \cup E^{\prime}$\;
  }
} \KwRet{$\seq{V,E,l,\Pi,X}$}\;
\end{algorithm}

\begin{algorithm}[t!]
\caption{calculateWeight}
\label{alg:calculateWeight} \KwIn{$\Pi:$ ground ASP program, $X:$
answer set for $\Pi$, $V:$ set of vertices, $l:V \rightarrow \Pi
\cup X$, $v:$ vertex in $V$, $E:$ set of edges, $W_T:$ candidate
weight function.} \KwOut{Weight of $v$.} \If{$l(v) \in X$} {
  \lForEach{$c \in \ii{child}_E(v)$}
  {
    $W_T(c) := \mbox{calculateWeight}(\Pi, X, V, l, c, E, W_T)$\;
  }
  $W_T(v) := min\{W_T(c) \sep c \in \ii{child}_E(v)\}$\;
} \ElseIf{$l(v) \in \Pi$} {
  $W_T(v) := 1$\;
  \lForEach{$c \in \ii{child}_E(v)$}
  {
    $W_T(v) := W_T(v) + \mbox{calculateWeight}(\Pi, X, V, l, c, E, W_T)$\;

  }
} \KwRet{$W_T(v)$}\;
\end{algorithm}

\begin{algorithm}[t]
\caption{extractExp}
\label{alg:extractExp} \KwIn{$\Pi:$ ground ASP program, $X:$ answer
set for $\Pi$, $V_t:$ set of vertices, $l : V_t \rightarrow \Pi \cup
X$, $v:$ vertex in $V_t$, $E_t:$ set of edges, $W_T:$ weight
function of $T$, $r:$ rule vertex in $V_t$ or $\emptyset$,
$\ii{op}$: string $\ii{min}$ or $\ii{max}$.}

\KwOut{A vertex-labeled tree $\seq{V,E,l,\Pi,X}$.} $V := \emptyset,
\ E := \emptyset$\; \If{$l(v) \in X$} {
  $c \leftarrow$ Pick $\ii{op}$ weighted child of $v$ \;
  \lIf{$r \neq \emptyset$}
  {
    $E := E \cup \{(r, c)$\}\;
  }
  $\seq{V^{\prime},E^{\prime}, l, \Pi, X} :=  \mbox{extractExp}(\Pi, X, V_t, l, c, E_t, W_T, r, \ii{op})$\;
  $V := V \cup V^{\prime}$,\, $E := E \cup E^{\prime}$\;
} \ElseIf{$l(v) \in \Pi$} {
  $V := V \cup \{v\}$\;
  \ForEach{$c \in \ii{child}_{E_{t}}(v)$}
  {
  $\seq{V^{\prime},E^{\prime}, l, \Pi, X} := \mbox{extractExp}(\Pi, X, V_t, l, c, E_t, W_T, v, \ii{op})$\;
    $V := V \cup V^{\prime}$,\, $E := E \cup E^{\prime}$\;
  }
} \KwRet{$\seq{V,E,l,\Pi,X}$}\;
\end{algorithm}

\begin{figure}[!t]
\centering
\includegraphics[scale=.9]{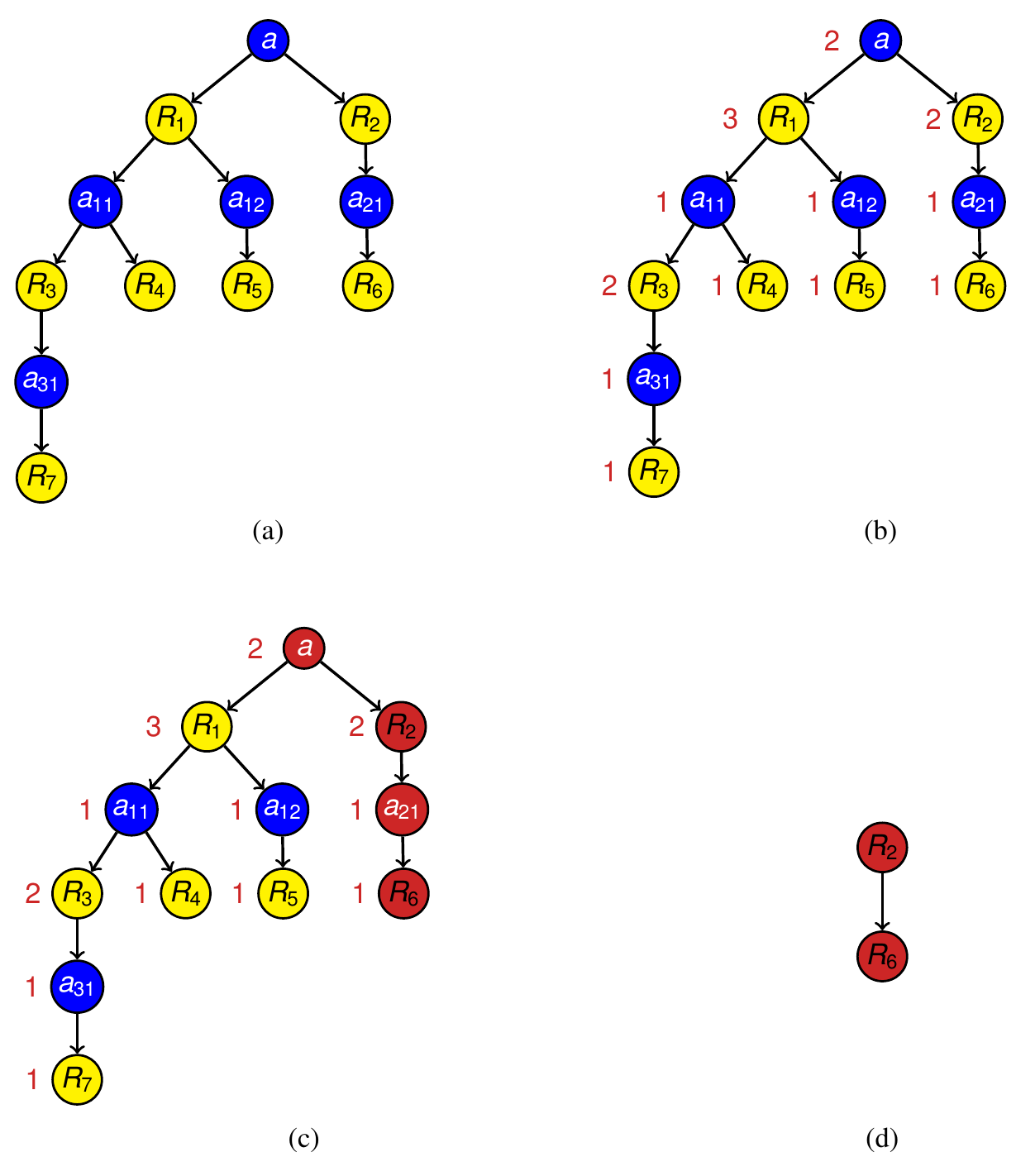}
\caption{A generic execution of Algorithm~\ref{alg:shortExp}.}
\label{fig:shortestExpFig}
\end{figure}

The execution of Algorithm~\ref{alg:shortExp} is also illustrated in
Figure~\ref{fig:shortestExpFig}. First, the and-or explanation tree
is generated, which has a generic structure as in
Figure~\ref{fig:shortestExpFig}(a). Here, yellow vertices denote
atom vertices and blue vertices denote rule vertices. Then, this
tree is weighted as in Figure~\ref{fig:shortestExpFig}(b). Then,
starting from the root, a subtree of the and-or explanation tree is
traversed by visiting minimum weighted child of every atom vertex
and every child of every rule vertex. This process is shown in
Figure~\ref{fig:shortestExpFig}(c), where red vertices form the
traversed subtree. From this subtree, an explanation is extracted by
ignoring atom vertices and keeping the parent-child relationship of
the tree as it is. The resulting explanation is depicted in
Figure~\ref{fig:shortestExpFig}(d).

\begin{proposition}
\label{prop:shortExpTerminate} Given a ground ASP program $\Pi$, an
answer set $X$ for $\Pi$, and an atom $p$ in $X$,
Algorithm~\ref{alg:shortExp} terminates.
\end{proposition}

\begin{proposition}
\label{prop:shortExpSound} Given a ground ASP program $\Pi$, an
answer set $X$ for $\Pi$, and an atom $p$ in $X$,
Algorithm~\ref{alg:shortExp} either finds a shortest explanation for
$p$ with respect to $\Pi$ and $X$ or returns an empty vertex-labeled
tree.
\end{proposition}

\begin{proposition}
\label{prop:shortExpComplexity} Given a ground ASP program $\Pi$, an
answer set $X$ for $\Pi$, and an atom $p$ in $X$, the time
complexity of Algorithm~\ref{alg:shortExp} is $O(|\Pi|^{|X|} \times
|\mathcal{B}_{\Pi}|)$.
\end{proposition}

We generate the complete and-or explanation tree while finding a
shortest explanation. In fact, we can find a shortest explanation by
creating a partial and-or explanation tree using a branch-and-bound
idea. In particular, the idea is to compute the weights of vertices
during the creation of the and-or explanation tree and, in case
there exists a branch of the and-or explanation tree that exceeds
the weight of a vertex computed so far, to stop branching on
unnecessary parts of the and-or explanation tree. Then, a shortest
explanation can be extracted by the same method used previously,
i.e., by traversing a subtree of the and-or explanation tree and
ignoring the atom vertices in this subtree. For instance, consider
Figure~\ref{fig:shortestExpFig}(b). Assume that we first create the
right branch of the root. Since the weight of an atom vertex is
equal to the minimum weight among its children weights, we know that
the weight of the root is at most 2. Now, we check whether it is
necessary to branch on the left child of the root. Note that the
weight of a rule vertex is equal to 1 plus the sum of its children
weights. As $R_1$ has two children, its weight is at least 3.
Therefore, it is redundant to branch on the left child of the root.
This improvement is not implemented and is a future work.


\section{Generating $k$ Different Explanations}
\label{sec:kDiff}

When there is more than one explanation for an answer of a query, it
might be useful to provide the experts with several more
explanations that are different from each other. For instance,
consider the query Q5 in Table~\ref{tab:queries}.

\begin{itemize}
\item [Q5] What are the drugs that treat the disease Asthma or that react with the drug Epinephrine?
\end{itemize}

\noindent An answer for this query is ``Doxepin''.
According to one explanation, ``Doxepin'' reacts with ``Epinephrine'' with respect to \drugbank.
At this point, the expert may not be convinced and ask for a different explanation.
Another explanation for this answer is that ``Doxepin'' treats ``Asthma'' according to \ctd.
Motivated by this example, we study generating different explanations.
\begin{algorithm}[t]
\KwIn{$\Pi$: ground ASP program,
$X$: answer set for $\Pi$, $p$: atom in $X$, $k:$ a positive
integer. Assume there are $n$ different explanations for $p$ w.r.t
$\Pi$ and $X$.} \KwOut{$\ii{min}\{n,k\}$ different explanations for
$p$ with respect to $\Pi$ and $X$.} $K:=\emptyset$, $R_0 :=
\emptyset$\; $\seq{V,E,l,\Pi,X} := \mbox{createTree}(\Pi, X, p,
\{\})$\;
  $v \lar $ root of $\seq{V,E}$\;
  \For{$i=1,2,\ldots k$}
  {
    $\mbox{calculateDifference}(\Pi, X, V, l, v, E, R_{i-1}, W_{T,R_{i-1}})$\;
    \lIf{$W_{T,R_{i-1}}(v) = 0$}
    {
      \KwRet{$K$}\;
    }
    $\seq{V',E',l,\Pi,X} := \mbox{extractExp}(\Pi,X,V,l,v,E,W_{T,R_{i-1}},\emptyset, \ii{max})$\;
    $K_i \lar \seq{V',E',l,\Pi,X}$\;
    $K := K \cup \{K_i\}$\;
    $R_i := R_{i-1} \cup \{v \sep \mbox{rule vertex}\ v \in V^{\prime}\}$\;
  }
\KwRet{$K$}\; \caption{Generating $k$ Different Explanations}
\label{alg:kDiff}
\end{algorithm}

We introduce an algorithm (Algorithm~\ref{alg:kDiff}) to compute $k$
different explanations for an atom $p$ in $X$ with respect to $\Pi$
and $X$. For that, we define a distance measure $\Delta_D$ between a
set $Z$ of (previously computed) explanations, and an (to be
computed) explanation $S$. We consider the rule vertices $R_Z$ and
$R_S$ contained in $Z$ and $S$, respectively. Then, we define the
function $\Delta_D$ that measures the distance between $Z$ and $S$
as follows:
$$\Delta_D(Z,S) = |R_S \backslash R_Z|.$$
In the following, we sometimes use $R_Z$ and $R_S$ instead of $Z$
and $S$ in $\Delta_D$. Also, we denote by $\ii{RVertices}(S)$ the
set of rule vertices of a vertex-labeled tree $S$.

Let us now explain Algorithm~\ref{alg:kDiff}. It computes a set $K$
of $k$ different explanations iteratively. Initially, $K =
\emptyset$. First, we compute the and-or explanation tree~$T$
(Line~$2$). Then, we enter into a loop that iterates at most $k$
times (Line~$4$). At each iteration $i$, an explanation $K_i$ that
is most distant from the previously computed $i-1$ explanations is
extracted from $T$. Let us denote the rule vertices included in the
previously computed $i-1$ explanations by $R_{i-1}$. Then,
essentially, at each iteration we pick an explanation $K_i$ such
that $\Delta_D(R_{i-1}, \ii{RVertices}(K_i))$ is maximum. To be able
to find such a $K_i$, we need to define the ``contribution'' of each
vertex $v$ in $T$ to the distance measure $\Delta_D(R_{i-1},
\ii{RVertices}(K_i))$ if $v$ is included in explanation $K_i$:
\[
W_{T,R_{i-1}}(v) = \left\{
\begin{array}{ll}
\ii{max}\{W_{T,R_{i-1}}(v') \sep v' \in \ii{child}_E(v)\} & \mbox{if} \ v \ \mbox{is an atom vertex}; \\
\sum_{v' \in \ii{child}_E(v)} W_{T,R_{i-1}}(v') &  \mbox{if} \ v \ \mbox{is a rule vertex and}\ \\
                                                & v \in R_{i-1}; \\
1 + \sum_{v' \in \ii{child}_E(v)} W_{T,R_{i-1}}(v')  &
\mbox{otherwise}.
\end{array}
\right.
\]
Note that this function is different from $W_T$. Intuitively, $v$
contributes to the distance measure if it is not included in
$R_{i-1}$. The contributions of vertices in $T$ are computed by
Algorithm~\ref{alg:calcDiff} (Line~$5$) by propagating the
contributions up in the spirit of Algorithm~\ref{alg:calculateWeight}.
Then, $K_i$ is extracted from weighted-$T$ by using
Algorithm~\ref{alg:extractExp} (Line~$7$).

\begin{algorithm}[t]
\KwIn{$\Pi:$ ground ASP program,
$X:$ answer set for $\Pi$, $V:$ set of vertices, $l:V \rightarrow
\Pi \cup X$, $v:$ vertex in $V$, $E:$ set of edges, $R:$ set of rule
vertices in $V$, $D_R:$ candidate distance function.}
\KwOut{distance of $v$.} \If{$l(v) \in X$} {
  \ForEach{$c \in \ii{child}_E(v)$}
  {
    $D_R(c) := \mbox{calculateDifference}(\Pi, X, V, l, c, E, R, D_R)$\;
  }
  $D_R(v) := \ii{max}\{D_R(c) \sep c \in \ii{child}_E(v)\}$\;
} \ElseIf{$l(v) \in \Pi$} {
  \lIf{$v \notin R$}
  {
    $D_R(v) := 1$\;
  }
  \lElse
  {
    $D_R(v) := 0$\;
  }

  \ForEach{$c \in \ii{child}_E(v)$}
  {
    $D_R(v) := D_R(v) + \mbox{calculateDifference}(\Pi, X, V, l, c, E, R, D_R)$\;
  }
} \KwRet{$D_R(v)$}\; \caption{calculateDifference}
\label{alg:calcDiff}
\end{algorithm}

\begin{figure}[!t]
\centering
\includegraphics[scale=0.8]{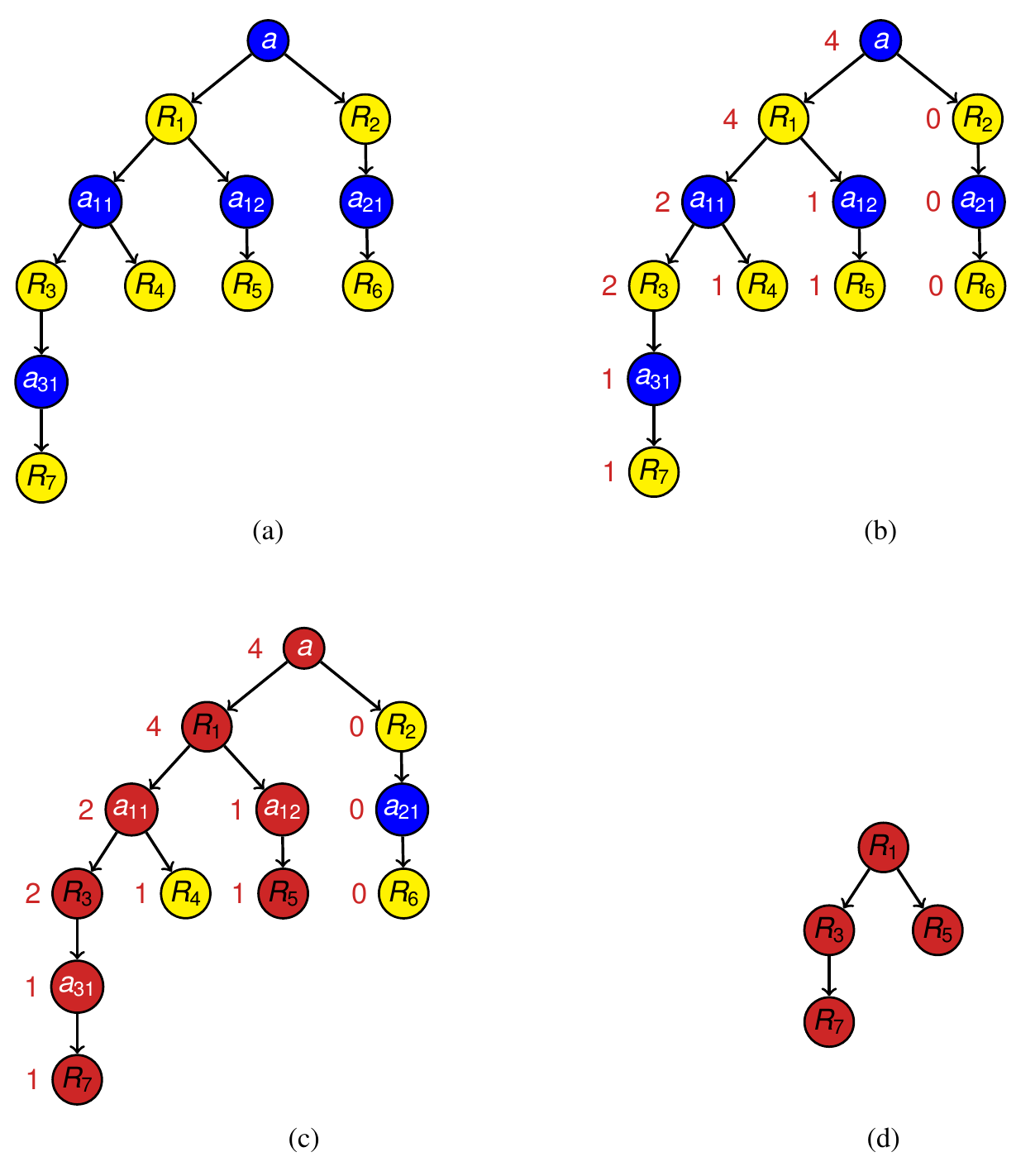}
\caption{A generic execution of Algorithm~\ref{alg:kDiff}.}
\label{fig:kDiffFig}
\end{figure}

The execution of Algorithm~\ref{alg:kDiff} is also illustrated in
Figure~\ref{fig:kDiffFig}. Similar to Algorithm~\ref{alg:shortExp},
which generates shortest explanations, first the and-or explanation
tree is created, which has a generic structure as shown in
Figure~\ref{fig:kDiffFig}(a). Recall that yellow vertices denote
atom vertices and blue vertices denote rule vertices. For the sake
of example, assume that $R = \{R_2, R_6\}$. Then, the goal is to
generate an explanation that contains different rule vertices from
the rule vertices in $R$ as much as possible. For that, the weights
of vertices are assigned according to the weight function $W_{T,R}$
as depicted in Figure~\ref{fig:kDiffFig}(b). Here, the weight of the
root implies that there exists an explanation which contains $4$
different rule vertices from the rule vertices in $R$ and this
explanation is the most different one. Then, starting from the root,
a subtree of the and-or explanation tree is traversed by visiting
maximum weighted child of every atom vertex, and every child of
every rule vertex. This subtree is shown in
Figure~\ref{fig:kDiffFig}(c) by red vertices. Finally, an
explanation is extracted by ignoring the atom vertices and keeping
the parent-child relationship as it is, from this subtree. This
explanation is illustrated in Figure~\ref{fig:kDiffFig}(d).

\begin{proposition}
\label{prop:kDiffTerminate} Given a ground ASP program $\Pi$, an
answer set $X$ for $\Pi$, an atom $p$ in $X$, and a positive integer
$k$, Algorithm~\ref{alg:kDiff} terminates.
\end{proposition}

\begin{proposition}
\label{prop:kDiffSound} Let $\Pi$ be a ground ASP program, $X$ be an
answer set for $\Pi$, $p$ be an atom in $X$, and $k$ be a positive
integer. Let $n$ be the number of different explanations for $p$
with respect to $\Pi$ and~$X$. Then, Algorithm~\ref{alg:kDiff}
returns $\ii{min}\{n,k\}$ different explanations for $p$ with
respect to $\Pi$ and~$X$.
\end{proposition}

\noindent Furthermore, at each iteration $i$ of the loop in
Algorithm~\ref{alg:kDiff} the distance $\Delta_D(R_{i-1},K_i)$ is
maximized.
\begin{proposition}
\label{prop:iteration} Let $\Pi$ be a ground ASP program, $X$ be an
answer set for $\Pi$, $p$ be an atom in $X$, and $k$ be a positive
integer. Let $n$ be the number of explanations for $p$ with respect
to $\Pi$ and~$X$. Then, at the end of each iteration $i$ ($1\leq i
\leq \ii{min}\{n,k\}$) of the loop in Algorithm~\ref{alg:kDiff},
$\Delta_D(R_{i-1},\ii{RVertices}(K_i))$ is maximized, i.e., there is
no other explanation $K'$ such that
$\Delta_D(R_{i-1},\ii{RVertices}(K_i)) <
\Delta_D(R_{i-1},\ii{RVertices}(K'))$.
\end{proposition}

This result leads us to some useful consequences. First,
Algorithm~\ref{alg:kDiff} computes ``longest'' explanations if
$k=1$. The following corollary shows how to compute longest
explanations.
\begin{corollary}
\label{cor:longestExplanation} Let $\Pi$ be a ground ASP program,
$X$ be an answer set for $\Pi$, $p$ be an atom in $X$, and $k=1$.
Then, Algorithm~\ref{alg:kDiff} computes a longest explanation for
$p$ with respect to $\Pi$ and $X$.
\end{corollary}
\noindent Next, we show that Algorithm~\ref{alg:kDiff} computes $k$
different explanations such that for every~$i$ ($1\leq i \leq k$)
the $i^{th}$ explanation is the most distant explanation from the
previously computed $i-1$ explanations.
\begin{corollary}
\label{cor:kDiff} Let $\Pi$ be a ground ASP program, $X$ be an
answer set for $\Pi$, $p$ be an atom in $X$, and $k$ be a positive
integer. Let $n$ be the number of explanations for $p$ with respect
to $\Pi$ and $X$. Then, Algorithm~\ref{alg:kDiff} computes
$\ii{min}\{n,k\}$ different explanations $K_1, \ldots,
K_{\ii{min}{\{n,k\}}}$ for $p$ with respect to $\Pi$ and $X$ such
that for every $j$ ($2 \leq j \leq \ii{min}\{n,k\}$)
$\Delta_D(\bigcup_{z=1}^{j-1} \ii{RVertices}(K_z), K_j)$ is
maximized.
\end{corollary}

The following proposition shows that the time complexity of
Algorithm~\ref{alg:kDiff} is exponential in the size of the given
answer set.
\begin{proposition}
\label{prop:kDiffComplexity} Given a ground ASP program $\Pi$, an
answer set $X$ for $\Pi$, an atom $p$ in $X$ and a positive integer
$k$, the time complexity of Algorithm~\ref{alg:kDiff} is $O(k \times
|\Pi|^{|X|+1} \times |\mathcal{B}_{\Pi}|)$.
\end{proposition}


\section{Experiments with Biomedical Queries}
\label{sec:expEX}

Our algorithms for generating explanations are applicable to the
queries Q1, Q2, Q3, Q4, Q5, Q8, Q10, Q11 and Q12 in
Table~\ref{tab:queries}. The ASP programs for the other queries
involve choice expressions. For instance, the query Q7 asks for
cliques of 5 genes. We use the following rule to generate a possible
set of 5 genes that might form a clique.
{
\begin{verbatim}
5{clique(GEN):gene_name(GEN)}5.
\end{verbatim}
}
\noindent Our algorithms apply to ASP programs that contain a single
atom in the heads of the rules, and negation and cardinality
expressions in the bodies of the rules. Therefore, our methods are
not applicable to the queries which are presented by ASP programs
that include choice expressions.

\begin{table}[t]
\caption{Experimental results for generating shortest explanations
for some biomedical queries, using Algorithm~\ref{alg:shortExp}.}
\label{tab:expShortestResults}
\begin{center}
{\footnotesize
\begin{tabular}{|c|c|c|c|c|c|}
\textbf{Query} & \textbf{CPU Time} & \textbf{Explanation} & \textbf{Answer Set} & \textbf{And-Or Tree} & \textbf{\gringo\ calls}\\
 & & \textbf{Size} & \textbf{Size} & \textbf{Size} & \\
\hline\hline Q1 & 52.78s
 & 5
 & 1.964.429
 & 16
 & 0
 \\
\hline
Q2 & 67.54s
 & 7
 & 2.087.219
 & 233
 & 1
 \\
\hline
Q3 & 31.15s
 & 6
 & 1.567.652
 & 15
 & 0
 \\
\hline
Q4 & 1245.83s
 & 6
 & 19.476.119
 & 6690
 & 4
 \\
\hline
Q5 & 41.75s
 & 3
 & 1.465.817
 & 16
 & 0
 \\
\hline
Q8 & 40.96s
 & 14
 & 1.060.288
 & 28
 & 4
 \\
\hline
Q10 & 1601.37s
 & 14
 & 1.612.128
 & 3419
 & 193
 \\
\hline
Q11 & 113.40s
 & 6
 & 2.158.684
 & 5528
 & 5
 \\
\hline
Q12 & 327.22s
 & 5
 & 10.338.474
 & 10
 & 1
 \\
\end{tabular}
}
\end{center}
\end{table}

In Table~\ref{tab:expShortestResults}, we present the results for
generating shortest explanations for the queries Q1, Q2, Q3, Q4, Q5,
Q8, Q10, Q11 and Q12. In this table, the second column denotes the
CPU timings to generate shortest explanations in seconds. The third
column consists of the sizes of explanations, i.e., the number of
rule vertices in an explanation. In the fourth column, the sizes of
answer sets, i.e., the number of atoms in an answer set, are given.
The fifth column presents the sizes of the and-or explanation trees,
i.e., the number of vertices in the tree.

Before telling what the
last column presents, let us clarify an issue regarding the
computation of explanations. Since answer sets contain millions of
atoms, the relevant ground programs are also huge. Thus,
first grounding the programs and then generating explanations over
those grounded programs is an overkill in terms of computational
efficiency. To this end, we apply another method and do grounding
when it is necessary. To better explain the idea, let us present our
method by an example. At the beginning, we have a ground atom
for which we are looking for shortest explanations. Assume that this atom is
$\ii{what\_be\_genes("ADRB1")}$. Then, we find the rules whose heads
are of the form $\ii{what\_be\_genes(GN)}$, and instantiate $GN$
with ``ADRB1''.
For instance, assume that the following rule exists in the program:
$$
\ii{what\_be\_genes(GN)} \lar \ii{drug\_gene(DRG,GN)}
$$
Then, by such an instantiation, we obtain the following instance
of this rule:
$$
\ii{what\_be\_genes("ADRB1")} \lar \ii{drug\_gene(DRG,"ADRB1")}
$$
Next, if the rules that we obtain by instantiating their heads are
not ground, we ground them using the grounder \gringo\ considering
the answer set. We apply the same method for the atoms that are now
ground, to find the relevant rules and ground them if necessary.
This allows us to deal with a relevant subset of the rules while
generating explanations. The last column of
Table~\ref{tab:expShortestResults} presents the number of times
\gringo\ is called for such incremental grounding.
For instance, for the queries Q1, Q3 and Q5, \gringo\ is never called.
However, \gringo\ is called 193 times during the computation of
a shortest explanation for the query~Q10.

As seen from the results presented in
Table~\ref{tab:expShortestResults}, the computation time is not very
much related to the size of the explanation. As also suggested by
the complexity results of Algorithm~\ref{alg:shortExp} (i.e.,
$O(|\Pi|^{|X|} \times |\mathcal{B}_{\Pi})|$), the computation time
for generating shortest explanations greatly depends on the sizes of
the answer set and the and-or explanation tree. For instance, for
the query Q4, the answer set contains approximately 19 million
atoms, the size of the and-or explanation tree is 6690, and it takes
1245 CPU seconds to compute a shortest explanation, whereas for the
query Q8, the answer set approximately contains 1 million atoms, the
and-or explanation tree has 28 vertices, and it takes 40 CPU seconds
to compute a shortest explanation. Also, the number of times
\gringo\ is called during the computation affects the computation
time. For instance, for the query Q10 the answer set approximately
contains 1.6 million atoms, the and-or explanation tree has 3419
vertices, and it takes 1600 CPU seconds to compute a shortest
explanation.

Table~\ref{tab:expkDiffResults} shows the computation times for
generating different explanations for the answers of the same
queries, if exists. As seen from these results, the time for
computing 2 and 4 different explanations is slightly different than
the time for computing shortest explanations.

\begin{table}[t]
\caption{Experimental results for generating different explanations
for some biomedical queries, using Algorithm~\ref{alg:kDiff}.}
\label{tab:expkDiffResults}
\begin{center}
{\small
\begin{tabular}{|c|c|c|c|}
\textbf{Query} & \multicolumn{3}{c|}{\textbf{CPU Time}} \\
\cline{2-4}
& $2$ different & $4$ different & Shortest \\
\hline
Q1 & 53.73s & - & 52.78s
 \\
\hline
Q2 & 66.88s & 67.15s & 67.54s
 \\
\hline
Q3 & 31.22s & - & 31.15s
 \\
\hline
Q4 & 1248.15s & 1251.13s & 1245.83s
 \\
\hline
Q5 & - & - & 41.75s
 \\
\hline
Q8 & - & - & 40.96s
 \\
\hline
Q10 & 1600.49s & 1602.16s & 1601.37s
 \\
\hline
Q11 & 113.25s & 112.83s & 113.40s
 \\
\hline
Q12 & - & - & 327.22s
 \\
\end{tabular}
}
\end{center}
\end{table}


\section{Presenting Explanations in a Natural Language}
\label{sec:expNL}

An explanation for an answer of a biomedical query may not be easy
to understand, since the user may not know the syntax of ASP rules
neither the meanings of predicates.
To this end, it is better to present explanations to the
experts in a natural language.

Observe that leaves of an explanation denote facts extracted from
the biomedical resources. Also some internal vertices contain
informative explanations such as the position of a drug in a chain
of drug-drug interactions. Therefore, there is a corresponding
natural language explanation for some vertices in the tree.
Such a correspondence can be stored in a predicate look-up table,
like Table~\ref{tab:lookup}. Given such a look-up table,
a pre-order depth-first traversal of an explanation and generating natural
language expressions corresponding to vertices of the explanation
lead to an explanation in natural language~\cite{Oztok12}.

For instance, the explanation in Figure~\ref{fig:expQ8} is expressed
in natural language as illustrated in the introduction.

\begin{table}[b!]
\caption{Predicate look-up table used while expressing explanations
in natural language.} \label{tab:lookup}
\begin{center}
{\footnotesize
\begin{tabular}{|c|l|}
\hline
\textbf{Predicate} & \textbf{Expression in Natural Language} \\
\hline
$\ii{gene\_gene\_biogrid(x,y)}$ & The gene x interacts with the gene y according to \biogrid. \\
$\ii{drug\_disease\_ctd(x,y)}$ & The disease y is treated by the drug x according to \ctd. \\
$\ii{drug\_gene\_ctd(x,y)}$ & The drug x targets the gene y according to \ctd.\\
$\ii{gene\_disease\_ctd(x,y)}$ & The disease y is related to the gene x according to \ctd. \\
$\ii{disease\_symptom\_do(x,y)}$ & The disease x has the symptom y according to \donto.\\
$\ii{drug\_category\_drugbank(x,y)}$ & The drug x belongs to the category y according to \drugbank.\\
$\ii{drug\_drug\_drugbank(x,y)}$ &  The drug x reacts with the drug y according to \drugbank.\\
$\ii{drug\_sideeffect\_sider(x,y)}$ & The drug x has the side effect y according to \sider.\\
$\ii{disease\_gene\_orphadata(x,y)}$ & The disease x is related to the gene y according to \orphadata. \\
$\ii{drug\_disease\_pharmgkb(x,y)}$ & The disease y is treated by the drug x according to \pharmgkb.\\
$\ii{drug\_gene\_pharmgkb(x,y)}$ &  The drug x targets the gene y according to \pharmgkb.\\
$\ii{disease\_gene\_pharmgkb(x,y)}$ & The disease x is related to the gene y according to \pharmgkb. \\
$\ii{start\_drug(x)}$ & The drug x is the start drug. \\
$\ii{start\_gene(x)}$ & The gene x is the start gene. \\
$\ii{drug\_reachable\_from}(x,l)$ & The distance of the drug x from the start drug is l. \\
$\ii{gene\_reachable\_from}(x,l)$ & The distance of the gene x from the start gene is l. \\
\hline
\end{tabular}
}
\end{center}
\end{table}


\section{Implementation of Explanation Generation Algorithms}
\label{sec:expIMP}

Based on the algorithms introduced above, we have developed a
computational tool called \expgenasp~\cite{Oztok12}, using the programming language
C++. Given an ASP program and its answer set, \expgenasp\ generates
shortest explanations as well as $k$ different explanations.

The input of \expgenasp\ are
\begin{itemize}
\item an ASP program $\Pi$,
\item an answer set $X$ for $\Pi$,
\item an atom $p$ in $X$,
\item an option that is used to generate either a shortest explanation or $k$ different explanations,
\item a predicate look-up table,
\end{itemize}
and the output are
\begin{itemize}
\item a shortest explanation for $p$ with respect to $\Pi$ and $X$ in a natural language
(if shortest explanation option is chosen),
\item $k$ different explanations for $p$ with respect to $\Pi$ and $X$ in a natural language
(if $k$ different explanations option is chosen).
\end{itemize}
For generating shortest explanations (resp., $k$ different
explanations), \expgenasp\ utilizes Algorithm~\ref{alg:shortExp}
(resp., Algorithm~\ref{alg:kDiff}).

\begin{figure}[h]
\begin{center}
\includegraphics[scale=.73]{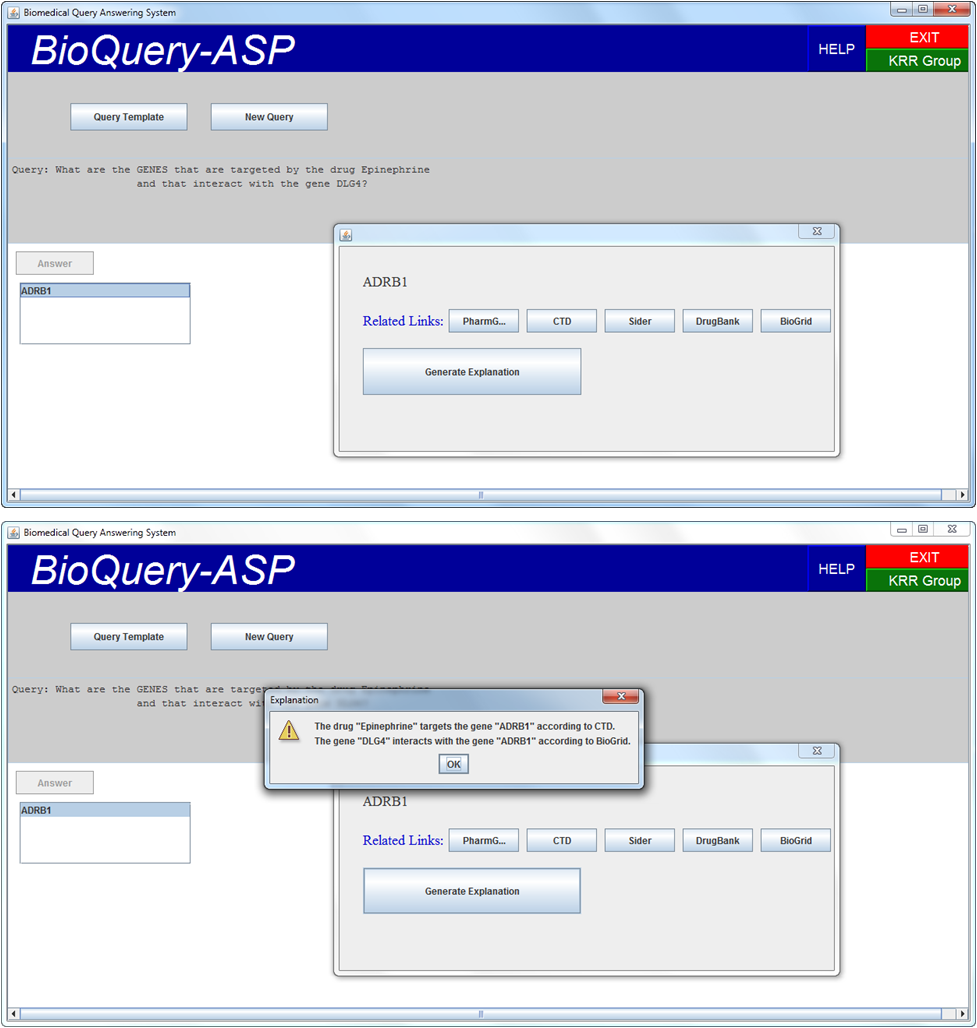}
\end{center}
\caption{A snapshot of \bqasp\ showing its explanation generation
facility.} \label{fig:expgenasp}
\end{figure}

To provide experts with further informative explanations about the
answers of biomedical queries, we have embedded \expgenasp\ into
\bqasp\ by utilizing Table~\ref{tab:lookup} as the predicate look-up
table of the system. Figure~\ref{fig:expgenasp} shows a snapshot of
the explanation generation mechanism of \bqasp.


\section{Relating Explanations to Justifications}
\label{sec:related}

\begin{figure}[h]
\begin{center}
\begin{tikzpicture}[sibling distance = 20mm]
\node{$a^+$}[->]
  child
  {
    node (b+) {$b^+$}
    edge from parent
    node[pos=0.4,left] {$+$}
  }
  child
  {
    node (c+) {$c^+$}
    child
    {
      node{$\top$}
      edge from parent
      node[left] {$+$}
    }
    edge from parent
    node[pos=0.4,right] {$+$}
  };
  \draw[->] (b+) -- (c+) node[pos=.5, above] {$+$};
\end{tikzpicture}
\end{center}
\caption{An offline justification for
Example~\ref{ex:pontelli}.} \label{fig:jstfPontelli}
\end{figure}
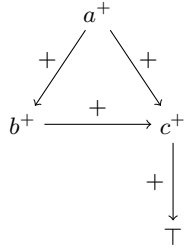

The most similar work to ours is~\cite{PontelliSE09} that study the
question ``why is an atom~$p$ in an answer set~$X$ for an ASP
program~$\Pi$''. As an answer to this question, the authors
of~\cite{PontelliSE09} finds a ``justification'', which is a labeled
graph that provides an explanation for the truth values of atoms
with respect to an answer set.
\begin{example}
\label{ex:pontelli} Let $\Pi$ be the program presented in Example~\ref{ex:andor}:
$$
\begin{array}{l}
a \lar b,c \\
a \lar d \\
d \lar \\
b \lar c \\
c \lar
\end{array}
$$
and $X =~\{a,b,c,d\}$. Figure~\ref{fig:jstfPontelli} is an offline
justification of $a^+$ with respect to $X$ and $\emptyset$.
Intuitively, $a$ is in $X$ since $b$ and $c$ are also in $X$ and
there is a rule in $\Pi$ that supports $a$ using the atoms $b$ and
$c$.  Furthermore, $b$ is in $X$ since $c$ is in $X$ and there is a
rule in $\Pi$ that supports $b$ using the atom $c$. Finally, $c$ is
in $X$ as it is a fact in $\Pi$.
\end{example}

To relate offline justifications and explanations, we need to
introduce the following definitions and notations about
justifications defined in~\cite{PontelliSE09}.

\subsection{Offline Justifications}

First, let us introduce notations related to ASP programs used
in~\cite{PontelliSE09}. The class of ASP programs studied is normal
programs, i.e., programs that consist of the rules of the form
$$
A \lar A_1, \ldots, A_k, not\ A_{k+1}, \ldots, not\ A_m
$$
where $m \geq k \geq 0$ and $A$ and each $A_i$ is an atom.
Therefore, the programs we consider are more general. Let $\Pi$ be a
normal ASP program. Then, $\mathcal{A}_{\Pi}$ is the Herbrand base
of $\Pi$. An \emph{interpretation} $I$ for a program $\Pi$ is
defined as a pair $\seq{I^+, I^-}$, where $I^+ \cup I^- \subseteq
\mathcal{A}_{\Pi}$ and $I^+ \cap I^- = \emptyset$. Intuitively,
$I^+$ denotes the set of atoms that are true, while $I^-$ denotes
the set of atoms that are false. $I$ is a \emph{complete
interpretation} if $I^+ \cup I^- = \mathcal{A}_{\Pi}$. The
\emph{reduct} $\Pi^I$ of $\Pi$ with respect to $I$ is defined as
$$
\Pi^I = \{H(r) \lar B^+(r) \sep r \in \Pi, \, B^-(r) \cap I^+ =
\emptyset\}
$$
A complete interpretation $M$ for a program $\Pi$ is an answer set
for $\Pi$ if $M^+$ is an answer set for $\Pi^M$. Also, a
\emph{literal} is either an atom or a formula of the form $not\ a$
where $a$ is an atom. The set of atoms which appear as negated
literals in $\Pi$ is denoted by $NANT(\Pi)$. For an atom $a$, $a^+$
denotes that the atom $a$ is true and $a^-$ denotes that $a^-$ is
false. Then, $a^+$ and $a^-$ are called the \emph{annotated}
versions of $a$. Moreover, it is defined that $atom(a^+) = a$ and
$atom(a^-) = a$. For a set $S$ of atoms, the following sets of
annotated atoms are defined.
\begin{itemize}
\item $S^p = \{a^+ \sep a \in S\}$
\item $S^n = \{a^- \sep a \in S\}$
\end{itemize}
Finally, the set $not\ S$ is defined as $not\ S = \{not\ a \sep a
\in S\}$.

Apart from the answer set semantics, there is another important
semantics of logic programs, called the well-founded
semantics~\cite{GelderRS91}. Since this semantics is important to
build the notion of a justification, we now briefly describe the
well-founded semantics. We consider the definition proposed
in~\cite{AptB94}, instead of the original definition proposed
in~\cite{GelderRS91}, as the authors of~\cite{PontelliSE09}
considered.
\begin{definition}[Immediate consequence] Let $\Pi$ be a normal ASP program, and $S$ and $V$ be
two sets of atoms from $\mathcal{A}_{\Pi}$. Then, the
\emph{immediate consequence} of $S$ with respect to $\Pi$ and $V$,
denoted by $T_{\Pi,V}(S)$ is the set defined as follows:
$$
T_{\Pi,V}(S) = \{a \sep \exists\ r \in \Pi,\, H(r) = a,\,
B^+(r)\subseteq S,\, B^-(r) \cap V = \emptyset\}.
$$
\end{definition}
We denote by $lfp(T_{\Pi,V})$ the least fixpoint of $T_{\Pi,V}$ when
$V$ is fixed.
\begin{definition}[The well-founded model] Let $\Pi$ be a normal ASP program,
$\Pi^+ = \{r \sep r \in \Pi,\, B^-(r) = \emptyset\}$. The sequence
$\seq{K_i,U_i}_{i \geq 0}$ is defined as follows:
$$
\begin{array}{l l l}
K_0 = lfp(T_{\Pi^+}), & \qquad &  U_0 = lfp(T_{\Pi, K_0}), \\
K_i = lfp(T_{\Pi,U_{i-1}}), & \qquad &  U_i = lfp(T_{\Pi, K_i}). \\
\end{array}
$$
Let $j$ be the first index of the computation such that
$\seq{K_j,U_j} = \seq{K_{j+1},U_{j+1}}$. Then, \emph{the
well-founded model} of $\Pi$ is $WF_{\Pi} = \seq{W^+,W^-}$ where
$W^+ = K_j$ and $W^- = \mathcal{A}_{\Pi} \backslash U_j$.
\end{definition}

\begin{example}
\label{ex:wellFounded} Let $\Pi$ be the program
$$
\begin{array}{l}
a \lar b, not\ d \\
d \lar b, not\ a \\
b \lar c \\
c \lar
\end{array}
$$
Then, the well-founded model of $\Pi$ is computed as follows:
\begin{eqnarray*}
K_0 & = & \{b,c\}, \\
U_0 & = & \{a,b,c,d\}, \\
K_1 & = & \{b,c\}, \\
U_0 & = & \{a,b,c,d\}.
\end{eqnarray*}
Thus, $WF_{\Pi} = \seq{\{b,c\},\emptyset}$.
\end{example}

\begin{figure}[tbhp]
\begin{center}
\begin{tikzpicture}[sibling distance = 20mm]
\node{$a^+$}[->]
  child
  {
    node{$b^+$}
    child
    {
      node{$assume$}
      edge from parent
      node[left] {$+$}
    }
    edge from parent
    node[left] {$+$}
  }
  child
  {
    node{$c^-$}
    child
    {
      node{$assume$}
      edge from parent
      node[left] {$-$}
    }
    edge from parent
    node[right] {$-$}
   };
\end{tikzpicture}
\end{center}
\caption{An e-graph for
Example~\ref{ex:egraph}.} \label{fig:egraph}
\end{figure}
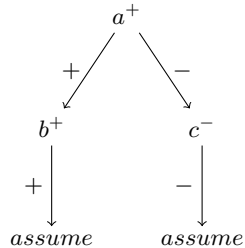

We now provide definitions regarding the notion of an offline
justification. First, we introduce the basic building of an offline
justification, a labeled graph called as e-graph.
\begin{definition}[e-graph]
\label{def:egraph} Let $\Pi$ be a normal ASP program. An
\emph{e-graph} for $\Pi$ is a labeled, directed graph $(N,E)$, where
$N \subseteq \mathcal{A}_{\Pi}^p \cup \mathcal{A}_{\Pi}^n \cup
\{assume, \top, \bot \}$ and $E \subseteq N \times N \times
\{+,-\}$, which satisfies following properties:
\begin{itemize}
\item[{\it (i)}] the only sinks (i.e., nodes without out-going edges) in the graph are $assume, \top, \bot$;
\item[{\it (ii)}] for every $b \in N \cap \mathcal{A}_{\Pi}^p$, $(b, assume, -) \notin E$ and $(b, \bot, -) \notin E$;
\item[{\it (iii)}] for every $b \in N \cap \mathcal{A}_{\Pi}^n$, $(b, assume, +) \notin E$ and $(b, \top, +) \notin E$;
\item[{\it (iv)}] for every $b \in N$, if for some $l \in \{assume, \top, \bot\}$ and $s \in \{+,-\}$, $(b,l,s) \in E$,
then $(b,l,s)$ is the only out-going edge originating from $b$.
\end{itemize}
\end{definition}
\noindent According to this definition, an edge of an e-graph
connects two annotated atoms or an annotated atom with one of the
nodes in $\{assume, \top, \bot\}$ and is marked by a label from
$\{+,-\}$. An edge is called as \emph{positive} (resp.,
\emph{negative}) if it is labeled by $+$ (resp., $-$). Also, a path
in an e-graph is called as \emph{positive} if it has only positive
edges, whereas it is called as \emph{negative} if it has at least
one negative edge. The existence of a positive path between two
nodes $v_1$ and $v_2$ is denoted by $(v_1,v_2) \in E^{*,+}$. In the
offline justification, $\top$ is used to explain facts, $\bot$ to
explain atoms which do not have defining rules, and $assume$ is for
atoms for which explanations are not needed, i.e., they are assumed
to be true or false.
\begin{example}
\label{ex:egraph} Let $\Pi$ be the program presented in Example~\ref{ex:andor},
and $X =~\{a,b,c,d\}$. Then, Figure \ref{fig:egraph} is an e-graph
for $\Pi$. Intuitively, the true state of $a$ depends on the true
state of $b$ and the false state of $c$, where $b$ is assumed to be
true and $c$ is assumed to be false.
\end{example}

In an e-graph, a set of elements that directly contributes to the
truth value of an atom can be obtained through the out-going edges
of a corresponding node. This set is defined as follows.
\begin{definition}[$support(b,G)$]
\label{def:support} Let $\Pi$ be a normal ASP program, $G=(N,E)$ be
an e-graph for $\Pi$ and $b \in N \cap (\mathcal{A}_{\Pi}^p \cup
\mathcal{A}_{\Pi}^n)$ be a node in $G$. Then, \emph{$support(b,G)$}
is defined as follows.
\begin{itemize}
\item $support(b,G) = \{l\}$, if for some $l \in \{assume, \top, \bot\}$ and $s \in \{+,-\}$, $(b,l,s)$ is in $E$;
\item $support(b,G) = \{atom(c) \sep (b,c,+) \in E\} \cup \{not\, atom(c) \sep (b,c,-) \in E\}$, otherwise.
\end{itemize}
\end{definition}
\begin{example}
\label{ex:support} Let $G$ be the e-graph in
Figure~\ref{fig:egraph}. Then, $support(a,G) = \{b, not\ c\}$,
$support(b,G) = \{assume\}$ and $support(c,G) = \{assume\}$.
\end{example}

To define the notion of a justification, an e-graph should be
enriched with explanations of truth values of atoms that are derived
from the rules of the program. For that, the concept of one step
justification of a literal is defined as follows.
\begin{definition}[Local Consistent Explanation (LCE)]
\label{def:lce} Let $\Pi$ be a normal ASP program, $b$ be an atom,
$J$ be a possible interpretation for $\Pi$, $U \subseteq
\mathcal{A}_{\Pi}$ be a set of atoms, and  $S \subseteq
\mathcal{A}_{\Pi} \cup \mbox{not} \, \mathcal{A}_{\Pi} \cup
\{assume, \top, \bot\}$ be a set of literals. We say that
\begin{itemize}
\item $S$ is an \emph{LCE} of $b^+$ with respect to $(J,U)$, if $b \in J^+$ and
\begin{itemize}
\item[-] $S = \{assume\}$ or
\item[-] $S \cap \mathcal{A}_{\Pi} \subseteq J^+$, $\{c \, | \, \mbox{not} \, c \in S \} \subseteq J^- \cup U$,
and there is a rule $r \in \Pi$ such that $H(r) = b$ and $B(r) = S$.
In case, $B(r) = \emptyset$, $S$ is denoted by the set $\{\top\}$.
\end{itemize}
\item  $S$ is an \emph{LCE} of $b^-$ with respect to $(J,U)$, if $b \in J^- \cup U$ and
\begin{itemize}
\item[-] $S = \{assume\}$ or
\item[-] $S \cap \mathcal{A} \subseteq J^- \cup U$, $\{c \, | \, not \, c \in S \} \subseteq J^+$, and
$S$ is a minimal set of literals such that for every rule $r \in
\Pi$ if $H(r) = b$, then $B^+(r) \cap S \neq \emptyset$ or $B^-(r)
\cap \{c \, | \, \mbox{not} \, c \in S\} \neq \emptyset$. In case,
$S = \emptyset$, $S$ is denoted by the set $\{\bot\}$.

\end{itemize}
\end{itemize}
The set of all the LCEs of $b^+$ with respect to $(J,U)$ is denoted
by $\mbox{LCE}_{\Pi}^{p}(b,J,U)$ and the set of all the LCEs of
$b^-$ with respect to $(J,U)$ is denoted by
$\mbox{LCE}_{\Pi}^{n}(b,J,U)$.
\end{definition}
\noindent Here, a possible interpretation $J$ denotes an answer set.
The set $U$ consists of atoms that are assumed to be false (which
will be called as Assumptions in the notion of justification later
on). The need for $U$ comes from the fact that the truth value of
some atoms is first guessed while computing answer sets.
Intuitively, if an atom $a$ is true, an LCE consists of the body of
a rule which is satisfied by $J$ and has $a$ in its head; if $a$ is
false, an LCE consists of a set of literals that are false in $J$
and falsify all rules whose head are $a$.
\begin{example}
\label{ex:lce} Let $\Pi$ and $X$ be defined as in
Example~\ref{ex:egraph}. Then, the LCEs of the atoms with respect to
$(X,\emptyset)$ is as follows.
\begin{eqnarray*}
\mbox{LCE}_{\Pi}^p(a, X, \emptyset) & = & \{\{b,c\},\{d\},\{assume\}\}\\
\mbox{LCE}_{\Pi}^p(b, X, \emptyset) & = & \{\{c\},\{assume\}\}\\
\mbox{LCE}_{\Pi}^p(c, X, \emptyset) & = & \{\{\top\},\{assume\}\}\\
\mbox{LCE}_{\Pi}^p(d, X, \emptyset) & = & \{\{\top\},\{assume\}\}\\
\end{eqnarray*}
\end{example}

Accordingly, a class of e-graphs where edges represent LCEs of the
corresponding nodes are defined as follows.
\begin{definition}[$(J,U)$-based e-graph]
\label{def:ju} Let $\Pi$ be a normal ASP program, $J$ be a possible
interpretation for $\Pi$, $U \subseteq \mathcal{A}_{\Pi}$ be a set
of atoms and $b$ be an element in $\mathcal{A}_{\Pi}^p \cup
\mathcal{A}_{\Pi}^n$. A \emph{$(J,U)$-based e-graph} $G = (N,E)$ of
$b$ is an e-graph such that
\begin{itemize}
\item[{\it (i)}] every node $c \in N$ is reachable from $b$,
\item[{\it (ii)}] for every $c \in N \backslash \{assume, \top, \bot\}$, $support(c,G)$ is an LCE of $c$ with respect to $(J,U)$;
\end{itemize}
A $(J,U)$-based e-graph $(N,E)$ is \emph{safe} if for all $b^+ \in
N, (b^+, b^+) \notin E^{*,+}$, i.e., there is no positive cycle in
the graph.
\end{definition}

We now introduce a special class of $(J,U)$-based e-graphs where
only false elements can be assumed.
\begin{definition}[Offline e-graph]
\label{def:offlinegraph} Let $\Pi$ be a normal ASP program, $J$ be a
partial interpretation for $\Pi$, $U \subseteq \mathcal{A}_{\Pi}$ be
a set of atoms and $b$ be an element in $\mathcal{A}^p \cup
\mathcal{A}^n$. An \emph{offline e-graph} $G = (N,E)$ of $b$ with
respect to $J$ and $U$ is a $(J,U)$-based e-graph of $b$ that
satisfies following properties:
\begin{itemize}
\item[{\it (i)}] there exists no $p^+ \in N$ such that $(p^+, assume, +) \in E$;
\item[{\it (ii)}] $(p^-, assume, -) \in E$ if and only if $p \in U$.
\end{itemize}
$\mathcal{E}(b,J,U)$ is the set of all offline e-graphs of $b$ with
respect to $J$ and $U$.
\end{definition}
\noindent Here, the roles of $J$ and $U$ are the same as their roles
in Definition~\ref{def:lce}. Observe that true atoms cannot be
assumed due to the first condition and only elements in the set $U$
are assumed due to the second condition.

We said earlier that in a $(J,U)$-based e-graph $J$ represents an
answer set and $U$ consists of atoms that are assumed to be false.
Here, $U$ is chosen based on some characteristics of $J$. In
particular, we want $U$ to be a set of atoms such that when its
elements are assumed to be false, the truth value of other atoms in
the program can be uniquely determined and leads to $J$.  We now
introduce regarding definitions formally.
\begin{definition}[Tentative Assumptions]
\label{def:tentative} Let $\Pi$ be a normal ASP program, $M$ be an
answer set for $\Pi$ and $WF_\Pi = <WF^+_{\Pi},WF^-_{\Pi}>$ be the
well-founded model of $\Pi$. The \emph{tentative assumptions}
$\mathcal{TA}_{\Pi}(M)$ of $\Pi$ with respect to $M$ are defined as
\begin{equation}
\mathcal{TA}_{\Pi}(M) = \{a \, | \, a \in NANT(\Pi) \wedge a \in
M^-, a \notin (WF^+_{\Pi} \cup WF^-_{\Pi})\}
\end{equation}
\end{definition}
\begin{example}
\label{ex:tentative} Let $\Pi$ be the program:
$$
\begin{array}{l}
c \lar a, not\ d \\
d \lar a, not\ c \\
a \lar b \\
b \lar
\end{array}
$$
Then, $X = \{a,b,c\}$ is an answer set for $\Pi$ and
$\seq{\{a,b\},\emptyset}$ is the well-founded model of $\Pi$. Given
that, $\mathcal{TA}_{\Pi}(X) = \{d\}$ as $d \in NANT(\Pi)$, $d
\notin X$ and $d \notin (WF^+_{\Pi} \cup WF^-_{\Pi})$.
\end{example}
\noindent In fact, tentative assumptions is a set of atoms whose
subsets might ``potentially'' form $U$.

We provide a definition that would allow one to obtain a program
from a given program $\Pi$ and a set $V$ of atoms by assuming all
the atoms in $V$ as false.
\begin{definition}[Negative Reduct]
\label{def:negReduct} Let $\Pi$ be a normal ASP program, $M$ be an
answer set for $\Pi$, and $U \subseteq \mathcal{TA}_{\Pi}(M)$ be a
set of tentative assumption atoms. The \emph{negative reduct}
$NR(\Pi,U)$ of $\Pi$ with respect to $U$ is the set of rules defined
as
\begin{equation}
NR(\Pi,U) = \Pi \backslash \{r \, | \, H(r) \in U\}
\end{equation}
\end{definition}

Finally, the concept of assumptions can be introduced formally.
\begin{definition}[Assumption]
\label{def:assumption} Let $\Pi$ be a normal ASP program, $M$ be an
answer set for $\Pi$. An \emph{assumption} of $\Pi$ with respect to
$M$ is a set $U$ of atoms that satisfies the following properties:
\begin{itemize}
\item[{\it (i)}] $U \subseteq \mathcal{TA}_{\Pi}(M)$;
\item[{\it (ii)}] the well-founded model of $NR(\Pi,U)$ is equal to $M$.
\end{itemize}
$Assumptions(\Pi,M)$ is the set of all assumptions of $\Pi$ with
respect to $M$.
\end{definition}
\begin{example}
Let $\Pi$ and $X$ be defined as in Example~\ref{ex:tentative}. Let
$U = \{d\}$. Then, $NR(\Pi,U)$ is:
$$
\begin{array}{l}
c \lar a, not\ d \\
a \lar b \\
b \lar
\end{array}
$$ and $\seq{\{a,b,c\},\emptyset}$ is
the well-founded model of $NR(\Pi,U)$. Thus, $U$ is an assumption of
$\Pi$ with respect to $X$.
\end{example}
\noindent Note that assumptions are nothing but subsets of tentative
assumptions that would allow to obtain the answer set $J$.

At last, we are ready to define the notion of offline justification.
\begin{definition}[Offline Justification]
\label{def:justification} Let $\Pi$ be a normal ASP program, $M$ be
an answer set for $\Pi$, $U$ be an assumption in
$Assumptions(\Pi,M)$ and $b$ be an annotated atom in $\mathcal{A}^p
\cup \mathcal{A}^n$. An \emph{offline justification} of $b$ with
respect to $M$ and $U$ is an element $(N,E)$ of $\mathcal{E}(b,M,U)$
which is safe.
\end{definition}
\noindent According to the definition, a justification is a
$(J,U)$-based e-graph where $J$ is an answer set and $U$ is an
assumption. Also, justifications do not allow the creation of
positive cycles in the justification of true atoms.
For instance,
for $\Pi$ and $X$ defined as in
Example~\ref{ex:andor},
Figure~\ref{fig:jstfPontelli}
illustrates an offline justification of $a^+$ with respect to $X$ and $\emptyset$.



In~\cite{PontelliSE09}, the authors prove the following proposition
which shows that for every atom in the program, there exists an
offline justification.
\begin{proposition}
\label{prop:pontelli} Let $\Pi$ be a ground normal ASP program, $X$
be an answer for $\Pi$. Then, for each atom $a$ in $\Pi$, there is
an offline justification of with respect to $X$ and $X^-\backslash
WF_{\Pi}^-$ which does not contain negative cycles.
\end{proposition}

\subsection{From Justifications to Explanations}

We relate a justification to an explanation. In particular, given an
offline justification, we show that one can obtain an explanation
tree whose atom vertices are formed by utilizing the ``annotated
atoms'' of the justification and rule vertices are formed by
utilizing the ``support'' of  annotated atoms. To compute such
explanation trees, we develop Algorithm~\ref{alg:jstf2exp}.
\begin{algorithm}[t]
\caption{Justification to
Explanation} \label{alg:jstf2exp} \KwIn{$\Pi:$ ground normal ASP
program, $X:$ answer set for $\Pi$, $p:$ atom in $X$,
      $(V,E):$ justification of $p^+$ w.r.t $X$ and some $U \in Assumptions(\Pi,X)$.}
\KwOut{A vertex-labeled tree $\seq{V',E',l,\Pi,X}$.} $V' :=
\emptyset, \, E' := \emptyset$\; $v \lar $ Create a vertex $v$ s.t.
$l(v) = p$\; $Q \lar v$\; \While{$Q \neq \emptyset$} {
  $v' \lar $ Dequeue an element from $Q$\;
  $V' := V' \cup \{v'\}$\;
  \If(\tcp*[h]{\footnotesize{$v'$ is a rule vertex}}){$l(v') \in \Pi$}
  {
    \ForEach{$a \in B^+(l(v'))$}
    {
      $v'' \lar$ Create a vertex $v''$ s.t. $l(v'') = a$\;
      $E' := E' \cup \{(v',v'')\}$ \tcp*[h]{\footnotesize{edge from rule vertex to atom vertex}}\;
      Enqueue $v''$ to $Q$\;
    }
  }

  \ElseIf(\tcp*[h]{\footnotesize{$v'$ is an atom vertex}}){$l(v') \in X$}
  {
    $r \lar $ Create a rule $r$ s.t. $H(r) = l(v')$ and $B(r) = support(l(v')^+,G)$\;
    $v'' \lar $ Create a vertex $v''$ s.t. $l(v'') = r$\;
    $E' := E' \cup \{(v',v'')\}$ \tcp*[h]{\footnotesize{edge from atom vertex to rule vertex}}\;
    Enqueue $v''$ to $Q$\;
  }
} \KwRet{$\seq{V',E',l,\Pi,X}$}\;
\end{algorithm}

Let us now explain the algorithm in detail.
Algorithm~\ref{alg:jstf2exp} takes as input a ground normal ASP
program $\Pi$ , an answer set $X$ for $\Pi$, an atom $p$ in $X$, and
a justification $(V,E)$ of $p^+$ with respect to $X$ and some $U \in
Assumptions(\Pi,X)$. Our goal is to obtain an explanation tree in
the and-or explanation tree for $p$ with respect to $\Pi$ and $X$
from the justification $(V,E)$. The algorithm starts by creating two
sets $V'$ and $E'$ (Line~$1$). Here, $V'$ and $E'$ corresponds to
the set of vertices and the set of edges of the explanation tree,
respectively. By Condition~$(ii)$ in Definition~\ref{def:expTree}
and Condition~$(i)$ in Definition~\ref{def:andor}, we know that the
label of the root of an explanation tree for $p$ with respect to
$\Pi$ and $X$ is $p$. Thus, a vertex $v$ with label $p$ is defined
(Line~$2$), and added into the queue~$Q$ (Line~$3$). Then, the
algorithm enters into a ``while'' loop which executes until $Q$
becomes empty. At every iteration of the loop, an element $v$ from
$Q$ is first extracted (Line~$5$) and added into $V'$ (Line~$6$).
This implies that every element added into $Q$ is also added into
$V'$. For instance, the vertex defined at Line~$2$ is the first
vertex extracted from $Q$ and also added into $V'$, which makes
sense since we know that the root of an explanation tree is an atom
vertex with label $v$. Then, according to the type of the extracted
vertex, its out-going edges are defined. Let $v'$ be a vertex
extracted from $Q$ at Line~$5$ in some iteration of the loop.
Consider the following two cases.

\begin{itemize}
\item[Case 1] Assume that $v'$ is an atom vertex. Then, the algorithm directly goes to Line~$13$.
By Condition~$(i)$ in Definition~\ref{def:expTree}, we know that an
explanation tree is a subtree of the and-or explanation tree. Hence,
we need to define out-going edges of $v'$ by taking into account
Condition~$(ii)$ in Definition~\ref{def:andor}, which implies that a
child of $v'$ must be a rule vertex $v''$ such that the rule that
labels $v''$ ``supports'' the atom that labels $v'$. Thus, a rule
$r$ that supports the atom that labels $v'$ is created (Line~$13$).
We ensure ``supportedness'' property by utilizing the annotated
atoms in the given offline justification which supports the
annotated version of the atom that labels $v'$. Then, a vertex $v''$
with label $r$ is created (Line~$14$), and a corresponding child of
$v'$ is added into $E'$ (Line~$15$). By Condition~$(iii)$ in
Definition~\ref{def:expTree}, we know that every atom vertex of an
explanation tree has a single child. Therefore, another child of
$v'$ is not created. Then, before finishing the iteration of the
loop, the child $v''$ of $v'$ is added into $Q$ so that its children
can be formed in the next iterations of the loop.

\item[Case 2] Assume that $v'$ is a rule vertex. Then, the condition at Line~$7$ is satisfied and the algorithm goes to Line~$8$.
In this case, while forming the children of $v'$, we should consider
Condition~$(iii)$ in Definition~\ref{def:andor}, which implies that
a child $v''$ of $v'$ must be an atom vertex such that the atom that
labels $v''$ is in the positive body of the rule that labels $v'$.
Also, by Condition $(iv)$ in Definition~\ref{def:expTree}, we should
ensure that for every atom $a$ in the positive body of the rule that
labels $v'$, there exists a child $v_a$ of $v'$ such that the atom
that labels $v_a$ is equal to $a$. Thus, the loop between
Lines~$8$--$11$ iterates for every atom $a$ in the positive body of
the label of $v'$ and a vertex $v''$ with label $a$ is created
(Line~$9$).  Then, $v''$ becomes a child of $v'$ (Line~$10$). To
form the children of $v''$ in the next iterations of the ``while''
loop, the child $v''$ of $v'$ is added into $Q$ (Line~$11$).

\end{itemize}

When the algorithm finishes processing the elements of $Q$, i.e., $Q$ becomes empty,
the ``while'' loop terminates. Then, the algorithm returns a
vertex-labeled tree (Line~$17$). We now provide the proposition
about the soundness of Algorithm~\ref{alg:jstf2exp}.

\begin{proposition}
\label{prop:jstf2exp} Given a ground normal ASP program $\Pi$, an
answer set $X$ for $\Pi$, an atom $p$ in $X$, an assumption $U$ in
$Assumption(\Pi,X)$, and an offline justification $G = (V,E)$ of
$p^+$ with respect to $X$ and $U$, Algorithm \ref{alg:jstf2exp}
returns an explanation tree $\seq{V',E',l,\Pi,X}$ in the and-or
explanation tree for $p$ with respect to $\Pi$ and $X$.
\end{proposition}

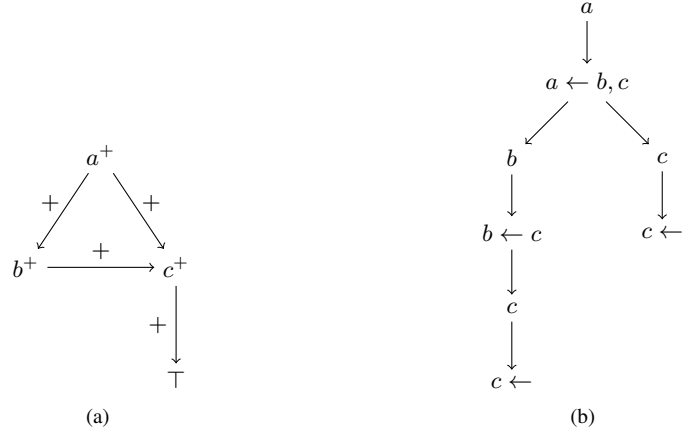
\begin{figure}[t]
\begin{center}
\subfloat[]{
\begin{minipage}[b]{0.5\linewidth}
\centering
\begin{tikzpicture}[sibling distance = 20mm]
\node (root) {$a^+$}[->]
  child
  {
    node (b+) {$b^+$}
    edge from parent
    node[pos=0.4,left] {$+$}
  }
  child
  {
    node (c+) {$c^+$}
    child
    {
      node{$\top$}
      edge from parent
      node[left] {$+$}
    }
    edge from parent
    node[pos=0.4,right] {$+$}
  };
  \draw[->] (b+) -- (c+) node[pos=.5, above] {$+$};
\end{tikzpicture}
\end{minipage}%
}
\subfloat[]{
\begin{minipage}[b]{0.5\linewidth}
\centering
\begin{tikzpicture}[level distance = 10mm, sibling distance = 20mm]
\node{$a$}[->]
  child
  {
    node {$a \lar b,c$}
    child
    {
      node {$b$}
      child
      {
        node {$b \lar c$}
    child
    {
      node {$c$}
      child
      {
        node {$c \lar$}
      }
        }
      }
    }
    child
    {
      node {$c$}
      child
      {
        node{$c \lar$}
      }
    }
  };
\end{tikzpicture}
\end{minipage}
}
\end{center}
\caption{(a) An offline justification and
(b) its corresponding explanation tree obtained by using
Algorithm~\ref{alg:jstf2exp}.} \label{fig:jstf2exp}
\end{figure}

\begin{example}
Let $\Pi$ and $X$ be defined as in Example~\ref{ex:egraph}.
Figure~\ref{fig:jstf2exp}(a) is an offline justification of $a^+$
with respect to $X$ and $\emptyset$. Figure~\ref{fig:jstf2exp}(b)
shows a corresponding explanation tree in the and-or explanation
tree for $a$ with respect to $\Pi$ and $X$ that is obtained by using
Algorithm~\ref{alg:jstf2exp}.
\end{example}

\subsection{From Explanations to Justifications}

We relate an explanation to a justification. In particular, given an
explanation tree whose labels of vertices are unique, we show that
one can obtain an offline justification by utilizing the labels of
atom vertices of the explanation tree. For that, we design
Algorithm~\ref{alg:exp2jstf}.
\begin{algorithm}[t!b!]
\caption{Explanation to
Justification} \label{alg:exp2jstf} \KwIn{$\Pi:$ ground normal ASP
program, $X:$ answer set for $\Pi$, $p:$ atom in $X$,
      $\seq{V',E',l,\Pi,X}:$ an explanation tree in the and-or explanation tree for $p$ w.r.t $\Pi$ and $X$.}
      \KwOut{$(V,E):$ justification of $p^+$ w.r.t $X$ and $\emptyset$.}
$V := \emptyset, \, E := \emptyset$\; $Q \lar $ root of
$\seq{V',E'}$\; \While{$Q \neq \emptyset$} {
  $v \lar $ Dequeue an element from $Q$\;
  $V := V \cup \{l(v)^+\}$\;
  $v' \lar $ child of $v$ in $\seq{V',E'}$\;
  \If{$l(v')$ is a fact in $\Pi^{X}$}
  {
    $E := E \cup \{(l(v)^+,\top, +)\}$\;
  }
  \ForEach{$v'' \in \ii{child}_{E'}(v')$}
  {
     $E := E \cup \{(l(v)^+,l(v'')^+, +)\}$\;
     Enqueue $v''$ to $Q$\;
  }
} $V := V \cup \{\top\}$\; \KwRet{$(V,E)$}\;
\end{algorithm}

Let us now describe the algorithm in detail.
Algorithm~\ref{alg:exp2jstf} takes as input a ground normal ASP
program $\Pi$, an answer set $X$ for $\Pi$, an atom $p$ in $X$, and
an explanation tree $T'=\seq{V',E',l,\Pi,X}$ in the and-or
explanation tree for $p$ with respect to $\Pi$ and $X$. Our goal is
to obtain an offline justification $(V,E)$ of $p^+$ in $\Pi^X$ with
respect to $X$ and~$\emptyset$. The reason to obtain the offline
justification in the reduct of $\Pi$ with respect to $X$ is that our
definition of explanation is not defined for the atoms that are not
in the answer set. Algorithm~\ref{alg:exp2jstf} starts by creating
two sets $V$ and $E$ which will correspond to the set of nodes and
the set of edges of the offline justification, respectively
(Line~$1$). Then, the root of $\seq{V',E'}$ is added into the queue
$Q$ (Line~$2$) and we enter into a ``while'' loop that iterates
until $Q$ becomes empty. At every iteration of the loop, first an
element $v$ is extracted from $Q$ (Line~$4$) and $l(v)^+$ is added
into $V$ (Line~$5$). Then, we form the out-going edges of $l(v)^+$.
Due to Condition $(iii)$ in Definition~\ref{def:expTree}, every atom
vertex in an explanation tree has a single child, which is a rule
vertex due to Condition $(i)$ in Definition~\ref{def:expTree} and
Condition $(ii)$ in Definition~\ref{def:andor}.  Then, we extract
the child $v'$ of $v$ at Line~$6$ and consider two cases. Note that
$v'$ is a rule vertex.

\begin{itemize}
\item[Case 1] Assume that $l(v')$ is a fact in $\Pi^X$. Then, $l(v')$ satisfies the condition
at Line~$7$ and we add $(l(v)^+, \top, +)$ into $E$ at Line~$8$. The
key insight behind that is as follows. Due to Condition $(ii)$ in
Definition~\ref{def:ju}, $support(l(v)^+, (V,E))$ must be an LCE of
$l(v)^+$. Due to Condition $(i)$ in Definition~\ref{def:expTree} and
Condition $(ii)$ in Definition~\ref{def:andor}, the head of $l(v')$
is $l(v)$.  As $l(v')$ is a fact in $\Pi^X$, i.e., its body is empty
in $\Pi^X$, $\{\top\}$ becomes an LCE of $l(v)^+$ with respect to
$(X,\emptyset)$, due to Definition~\ref{def:lce}. Thus, by adding
$(l(v)^+, \top, +)$ to $E$, $support(l(v)^+, (V,E))$ becomes
$\{\top\}$.

\item[Case 2] Assume that $l(v')$ is not a fact in $\Pi^X$. Then,
for every child $v''$ of $v'$, we add $(l(v)^+, l(v'')^+, +)$ into
$E$ at Line~$10$. The intuition behind this is to make sure that
$support(l(v)^+, (V,E))$ is an LCE of $l(v)^+$. Due to Condition
$(i)$ in Definition~\ref{def:expTree}, an explanation tree is a
subtree of the corresponding and-or explanation tree. Then, due to
Condition $(ii)$ in Definition~\ref{def:andor}, for every atom
vertex $v$ in an explanation tree, the atoms in the positive body of
the rule that labels the child $v'$ of $v$ are in the given answer
set~$X$. Thus, due to Definition~\ref{def:lce}, adding
$(l(v)^+,l(v'')^+,+)$ into $E$ for every child $v''$ of $v'$ ensures
that $support(l(v)^+, (V,E))$ is an LCE of $l(v)^+$ with respect to
$(X,\emptyset)$. Also, we add $v''$ into $V$ so that its children in
$V$ are formed in the next iterations of the ``while'' loop.

\end{itemize}

Due to Line~$8$, there are incoming edges of $\top$. But, $\top$ is not added into $V$
inside the ``while'' loop.  Thus, when the ``while'' loop
terminates, before returning $(V,E)$ at Line~$13$, we add $\top$
into $V$.

Algorithm~\ref{alg:exp2jstf} creates an offline justification of the
given atom in the reduct of the given ASP program with respect to
the given answer set, provided that labels of the vertices of the
given explanation tree are unique.

\begin{proposition}
\label{prop:exp2jstf} Given a ground normal ASP program $\Pi$ , an
answer set $X$ for $\Pi$, an atom $p$ in $X$, and an explanation
tree $\seq{V',E',l,\Pi,X}$ in the and-or explanation tree for~$p$
with respect to $\Pi$ and $X$ such that for every $v,v' \in V'$,
$l(v) = l(v')$ if and only if $v = v'$, Algorithm~\ref{alg:exp2jstf}
returns an offline justification of $p^+$ in $\Pi^X$ with respect to
$X$ and $\emptyset$.
\end{proposition}

\begin{figure}[h]
\begin{center}
\subfloat[]{
\begin{minipage}[b]{0.5\linewidth}
\centering
\begin{tikzpicture}[level distance = 13mm, sibling distance = 20mm]
\node{$a$}[->]
  child
  {
    node{$a \lar b,c, not\ d$}
    child
    {
      node{$b$}
      child
      {
        node{$b \lar not\ e$}
      }
    }
    child
    {
      node{$c$}
      child
      {
        node{$c \lar$}
      }
    }
  };
\end{tikzpicture}
\end{minipage}%
}
\subfloat[]{
\begin{minipage}[b]{0.5\linewidth}
\centering
\begin{tikzpicture}[level distance = 13mm, sibling distance = 20mm]
\node{$a^+$}[->]
  child
  {
    node {$b^+$}
    child
    {
      node {$\top$}
      edge from parent
      node[pos=0.4,left] {$+$}
    }
    edge from parent
    node[pos=0.4,left] {$+$}
  }
  child
  {
    node {$c^+$}
    child
    {
      node{$\top$}
      edge from parent
      node[pos=0.4,right] {$+$}
    }
    edge from parent
    node[pos=0.4,right] {$+$}
  };
\end{tikzpicture}
\end{minipage}
}
\end{center}
\caption{(a) An explanation tree and (b) its corresponding offline
justification obtained by using Algorithm~\ref{alg:exp2jstf}.}
\label{fig:exp2jstf}
\end{figure}
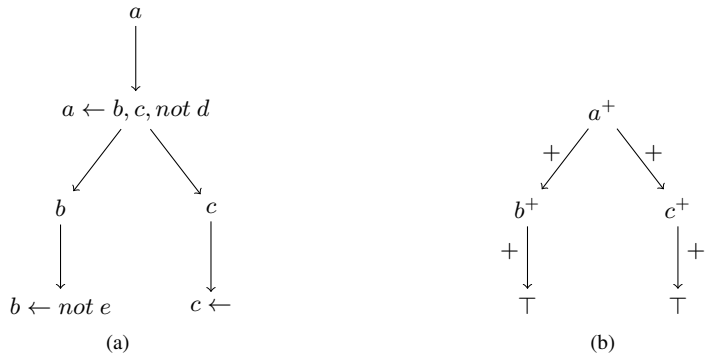

\begin{example}
Let $\Pi$ be the program:
$$
\begin{array}{l}
a \lar b, c, not\ d \\
b \lar not\ e \\
c \lar
\end{array}
$$
and $X = \{a,b,c\}$. Figure~\ref{fig:exp2jstf}(a) is an explanation
tree $T$ in the and-or explanation tree for $a$ with respect to
$\Pi$ and $X$. Then, given $\Pi,X, a$ and  $T$,
Algorithm~\ref{alg:exp2jstf} creates an offline justification of
$a^+$ in $\Pi^X$ with respect to $\Pi$ and $\emptyset$ as in
Figure~\ref{fig:exp2jstf}(b).
\end{example}

\section{Other Related Work}
\label{sec:relatedWork}

The most recent work related to explanation generation in ASP
are~\cite{BrainD05,Syrjanen06,GebserPST08,PontelliSE09,OetschPT10,OetschPT11},
in the context of debugging ASP programs.
Among those, \cite{Syrjanen06} studies why a program does not have
an answer set, and \cite{GebserPST08,OetschPT10} studies why a set
of atoms is not an answer set. As we study the problem of explaining
the reasons why atoms are in the answer set, our work differs from
those two work.

In~\cite{BrainD05}, similar to our work, the question ``why is an
atom~$p$ in an answer set~$X$ for an ASP program~$\Pi$'' is studied.
As an answer to this question, the authors of~\cite{BrainD05}
provide the rule in $\Pi$ that supports $X$ with respect to $\Pi$;
whereas we compute shortest or $k$ different explanations (as a tree
whose vertices are labeled by rules).

Pontelli et al.~\citeyear{PontelliSE09} also introduce the notion of an online
justification that aims to justify the truth values of atoms during
the computation of an answer set. In~\cite{OetschPT11}, a framework
where the users can construct interpretations through an interactive
stepping process is introduced. As a result, \cite{PontelliSE09} and
\cite{OetschPT11} can be used together to provide the users with
justifications of the truth values of atoms during the construction
of interpretations interactively through stepping.


\section{Conclusion}
\label{sec:conclusion}

We have formally defined explanations in the context of ASP. We have also introduced variations of
explanations, such as ``shortest explanations'' and ``$k$ different
explanations''.

We have proposed generic algorithms to generate explanations for biomedical queries.
In particular, we have presented algorithms to compute shortest or
$k$ different explanations. We have analyzed termination, soundness
and complexity of these algorithms. In particular, the complexity of
generating a shortest explanation for an answer (in an answer set
$X$) is $O(|\Pi|^{|X|} \times |\mathcal{B}_{\Pi}|)$ where $|\Pi|$ is
the number of ASP rules representing the query, the knowledge
extracted from biomedical resources and the rule layer, and
$|\mathcal{B}_{\Pi}|$ is the number of atoms in $\Pi$. The
complexity of generating $k$ different explanations is $O(k \times
|\Pi|^{|X| + 1} \times |\mathcal{B}_{\Pi}|)$. For $k$ different
explanations, we have defined a distance measure based on the number
of different ASP rules between explanations.

We have developed a computational tool \expgenasp\ which implements these algorithms.
We have embedded \expgenasp\ into \bqasp\ to generate explanations regarding the answers of complex biomedical queries.
We have proposed a method to present explanations in a natural
language.
No existing biomedical query answering system is capable of generating explanations; our methods have
fulfilled this need in biomedical query answering.

We have illustrated the applicability of our methods to answer complex biomedical queries over large
biomedical knowledge resources about drugs, genes, and diseases,
such as \pharmgkb, \drugbank, \biogrid, \ctd, \sider, \donto\ and
\orphadata. The total number of the facts extracted from these
resources to answer queries is approximately 10.3 million.

It is important to emphasize here that our definitions and methods
for explanation generation are general, so they can be applied to
other applications (e.g., debugging, query answering in other
domains).

One line of future work is to
generalize the notion of an explanation to queries (like Q7) that contain
choice expressions.


\section{Acknowledgments}

We would like to thank Yelda Erdem (Sanovel Pharmaceutical Inc.)
for her help in identifying biomedical queries relevant to drug discovery,
and Halit Erdogan (Sabanci University) for his help with an earlier
version of \bqasp\ which he implemented as part of his MS thesis studies.
We would like to thank Hans Tompits (Vienna University of Technology)
for his useful comments about the work presented in the paper, and
pointing out relevant references in the context of debugging ASP programs.
We also would like to thank anonymous reviewers for their useful
comments and suggestions on an earlier draft.
This work was partially supported by TUBITAK Grant 108E229.

%

\bibliographystyle{acmtrans}



\section*{Supplementary material}

Online appendix for the paper

# *Generating Explanations for Biomedical Queries*

published in Theory and Practice of Logic Programming

ESRA ERDEM, UMUT OZTOK

*Sabancı University, Orhanlı, Tuzla, İstanbul 34956, Turkey*
*(e-mail:* {esraerdem,uoztok}@sabanciuniv.edu*)*

## Appendix A  Proofs

We provide proofs of the theoretical results presented in the paper: the algorithmic analysis for generating shortest explanations (Propositions 2, 3, and 4), the algorithmic analysis for generating $k$ different explanations (Propositions 5, 6, and 8), and the analysis for the relationship between an explanation and a justification. (Propositions 10 and 11).

### *A.1  Generating Shortest Explanations*

In Section 5 of the paper, we have analyzed three properties of Algorithm 1, namely termination, soundness and complexity, resulting in Propositions 2, 3, and 4. In the following, we show the proofs of these results.

#### *A.1.1  Proof of Proposition 2 – Termination of Algorithm 1*

To prove that Algorithm 1 terminates we need the following lemma.

*Lemma 1*
Given a ground ASP program $\Pi$, an answer set $X$ for $\Pi$, an atom $p$ in $X$, and a set $L$ of atoms in $X$, Algorithm 2 terminates.

*Proof of Lemma 1*
It is sufficient to show that the recursion tree generated by Algorithm 2 is finite. That is, the branching factor of the recursion tree and the height of the recursion tree are finite. Note that each node of the recursion tree denotes a call to createTree($\Pi, X, d, L$) for some atom or rule $d$ and a set $L$ of atoms.

*Part (1)*  We show that the branching factor of the tree is finite. Branches in the tree are created in loops at Lines 5 and 13. The loop at Line 5 iterates at most the number of rules in $\Pi$ and the loop at Line 13 iterates at most the number of atoms in $X$. As $\Pi$ and $X$ are finite, the branching factor of the tree is finite.

*Part (2)* We show that the height of the recursion tree is finite. Let us first make the following observation. Consider a path $\langle v_1, \ldots \rangle$ in the recursion tree. For every node $v_i$ that denotes a call to createTree$(\Pi, X, d_i, L_i)$ where $d_i$ is an atom in $X$, the following holds for $v_{i+2}$ (if exists): $L_i \subset L_{i+2}$. This follows from the two consecutive calls to create-Tree: $L_i$ increases at every other call, due to Line 4. Now, assume that the height of the recursion tree is infinite. Then, there exists an infinite path $\langle v_1, \ldots \rangle$ in the recursion tree. Thus, $L_1 \subset L_3 \subset \ldots \subset L_{(2 \times |X|)+1} \subset L_{(2 \times |X|)+3} \subset \ldots$. Recall that we add elements into $L_i$s only from $X$ (Line 4). Thus, $L_{(2 \times |X|)+1} = X$. Then, it is not possible to have $L_{(2 \times |X|)+1} \subset L_{(2 \times |X|)+3}$. As we reach a contradiction, the height of the recursion tree is finite.

$\square$

Now, we show that Algorithm 1 terminates.

*Proposition 2*

Given a ground ASP program $\Pi$, an answer set $X$ for $\Pi$, and an atom $p$ in $X$, Algorithm 1 terminates.

*Proof of Proposition 2*

Algorithm 1 terminates only if Algorithms 2, 3, and 4 terminate. By Lemma 1, we know that Algorithm 2 terminates and that the vertex-labeled tree $T$ returned by Algorithm 2 is finite. Since Algorithm 3 and Algorithm 4 simply traverse $T$ (cf. Lines 2 and 6 in Algorithm 3, and Lines 5 and 10 in Algorithm 4), they also terminate. Thus, Algorithm 1 terminates.

$\square$

### A.1.2 Proof of Proposition 3 – Soundness of Algorithm 1

To show the proof of Proposition 3, we need the following necessary lemmas.

*Lemma 2*

Let $\Pi$ be a ground ASP program, $X$ be an answer set for $\Pi$, $d$ be an atom in $X$ or a rule in $\Pi$ and $L$ be a subset of $X$. If the vertex-labeled tree $\langle V, E, l, \Pi, X \rangle$ returned by createTree$(\Pi, X, d, L)$ is not empty, then the following hold:

*(i)* the root of $\langle V, E \rangle$ is created in createTree$(\Pi, X, d, L)$ and is mapped to $d$ by $l$;

*(ii)* for every rule vertex $v \in V$,

$$out_E(v) = \{(v, v') \mid (v, v') \in E, \, l(v') \in B^+(l(v))\};$$

*(iii)* each leaf vertex is a rule vertex.

*Proof of Lemma 2*

Let $\langle V, E, l, \Pi, X \rangle$ be the non-empty vertex-labeled tree returned by createTree$(\Pi, X, d, L)$. We show one by one that each condition in the lemma holds.

($i$) Assume that $d$ is an atom in $X$. Note that $d \notin L$ because otherwise $\langle V, E \rangle = \langle \emptyset, \emptyset \rangle$. Due to the call createTree($\Pi, X, d, L$), the algorithm, at Line 3, creates a vertex $v$ such that $l(v) = d$. We know that $E \neq \emptyset$. Then, there exists some out-going edges of $v$. At Line 9, the out-going edges of $v$ are formed. Due to Line 8, each vertex $v'$ in $(v, v')$ is the root of a vertex-labeled tree. Moreover, there is no part of the algorithm that adds a "parent" to a vertex. Therefore, $v$ is the root of $\langle V, E \rangle$.

Similar reasoning can be applied for the case where $d$ is a rule in $\Pi$.

($ii$) Let $v \in V$ be a rule vertex. Let $S_v = \{(v, v') \,|\, (v, v') \in E, \, l(v') \in B^+(l(v))\}$ denoting the set of out-going edges of $v$ to atom vertices whose labels are in the positive body of the rule that labels $v$. We show that $out_E(v) = S_v$.

Let $(v, v') \in out_E(v)$. Then, $(v, v') \in E$. Edges are created at Lines 9 and 17. As $v$ is a rule vertex, $(v, v')$ must be created at Line 17. Then, by Line 16 and the condition at Line 13, $l(v') \in B^+(l(v))$. So, $(v, v') \in S_v$ (i.e., $out_E(v) \subseteq S_v$).

Let $(v, v') \in S_v$. Then, by the definition of $S_v$, $(v, v') \in E$. Thus, $(v, v') \in out_E(v)$ (i.e., $S_v \subseteq out_E(v)$).

($iii$) Assume otherwise. Then, there exists an atom vertex $v \in V$ such that it is a leaf vertex. Vertices are created at Lines 3 and 12. As $v$ is an atom vertex, it must be created at Line 3. Since it is a leaf vertex, condition at Line 7 never holds. But, then condition at Line 10 holds. Then, the call where $v$ is created returns an empty set of vertices. That is, $v$ cannot be in $V$.

$\square$

*Lemma 3*

Let $\Pi$ be a ground ASP program and $X$ be an answer set for $\Pi$. For the two subsequent calls createTree($\Pi, X, d, L$) and createTree($\Pi, X, d', L'$) (i.e., createTree($\Pi, X, d', L'$) being called right after createTree($\Pi, X, d, L$)) on a path in the recursion tree for some execution of Algorithm 2, the following hold

  ($i$) If $d$ is an atom, then $L' = L \cup \{d\}$;
  ($ii$) If $d$ is a rule, then $L' = L$.

*Proof of Lemma 3*

Let createTree($\Pi, X, d, L$) and createTree($\Pi, X, d', L'$) be two subsequent calls on a path in the recursion tree for some execution of Algorithm 2. We show one by one that the conditions in the lemma hold.

($i$) Assume that $d$ is an atom. Then, createTree($\Pi, X, d', L'$) is called at Line 6. Due to Line 4, $L' = L \cup \{d\}$.

($ii$) Assume that $d$ is a rule. Then, createTree($\Pi, X, d', L'$) is called at Line 13. As $L$ is not modified prior to this call, $L' = L$.

$\square$

*Lemma 4*

Let $\Pi$ be a ground ASP program, $X$ be an answer set for $\Pi$ and $\langle V, E, l, \Pi, X \rangle$ be the vertex-labeled tree returned by execution $Exc$ of Algorithm 2. Let createTree$(\Pi, X, d, L)$ and createTree$(\Pi, X, d', L')$ be the two subsequent calls (i.e., createTree$(\Pi, X, d', L')$ being called right after createTree$(\Pi, X, d, L)$) on a path in the recursion tree for $Exc$, where each call on the path returns a nonempty vertex-labeled tree. Let $v$ and $v'$ be two vertices created by createTree$(\Pi, X, d, L)$ and createTree$(\Pi, X, d', L')$, respectively, such that $l(v) = d$ and $l(v') = d'$. Then, $(v, v') \in E$.

*Proof of Lemma 4*

Notice that either $d$ is an atom and $d'$ is a rule or vice versa. We show that the lemma holds for the former case. The latter case can be shown similarly. Assume that $d$ is an atom and $d'$ is a rule. Then, createTree$(\Pi, X, d', L')$ must be called at Line 6 within createTree$(\Pi, X, d, L)$. As none of the calls on the path returns empty vertex-labeled tree, the condition at Line 7 holds in createTree$(\Pi, X, d, L)$. Then, an edge $(v, v')$ is added to $E$ at Line 9. Note that $v$ is created in createTree$(\Pi, X, d, L)$ at Line 3 and $l(v) = d$. Also, due to Line 8, $v'$ is the root of the vertex-labeled tree returned by createTree$(\Pi, X, d', L')$. Then, by Lemma 2, $v'$ is a vertex created in createTree$(\Pi, X, d', L')$ and $l(v') = d'$.

$\square$

*Lemma 5*

Let $\Pi$ be a ground ASP program, $d$ be an atom in $X$ and $T = \langle V, E, l, \Pi, X \rangle$ be the vertex-labeled tree returned by createTree$(\Pi, X, d, \emptyset)$. If $T$ is not empty, then for every node createTree$(\Pi, X, d', L')$ in the recursion tree for createTree$(\Pi, X, d, \emptyset)$, where createTree$(\Pi, X, d', L')$ and its ancestors return nonempty vertex-labeled trees, $L' = anc_T(v)$ where $l(v) = d'$.

*Proof of Lemma 5*

Assume that $T$ is not empty. Then, we prove the lemma by induction on the depth of a node in the recursion tree for createTree$(\Pi, X, d, \emptyset)$.

*Base case:* Note that the node at depth 0 in the recursion tree for createTree$(\Pi, X, d, \emptyset)$ returns a nonempty vertex-labeled tree and it has no ancestors. By Lemma 2, the root $v$ of $\langle V, E \rangle$ is mapped to $d$ by $l$, i.e., $l(v) = d = d'$. As the root of a tree does not have any ancestors, $anc_T(v) = \emptyset = L'$.

*Induction step:* As an induction hypothesis, assume that for every node createTree$(\Pi, X, d', L')$ at depth less than $n$ in the recursion tree for createTree$(\Pi, X, d, \emptyset)$, where createTree$(\Pi, X, d', L')$ and its ancestors return nonempty vertex-labeled trees, $L' = anc_T(v)$ where $l(v) = d'$. Let createTree$(\Pi, X, d', L')$ be a node at depth $n$ in the recursion tree for createTree$(\Pi, X, d, \emptyset)$, where createTree$(\Pi, X, d', L')$ and its ancestors return nonempty vertex-labeled trees. We show that $L' = anc_T(v)$ where $l(v) = d'$. For that, we need to consider two cases: $d'$ is an atom and $d'$ is a rule. Let createTree$(\Pi, X, p, L_p)$ be the parent of createTree$(\Pi, X, d', L')$, as illustrated in Figure A 1.

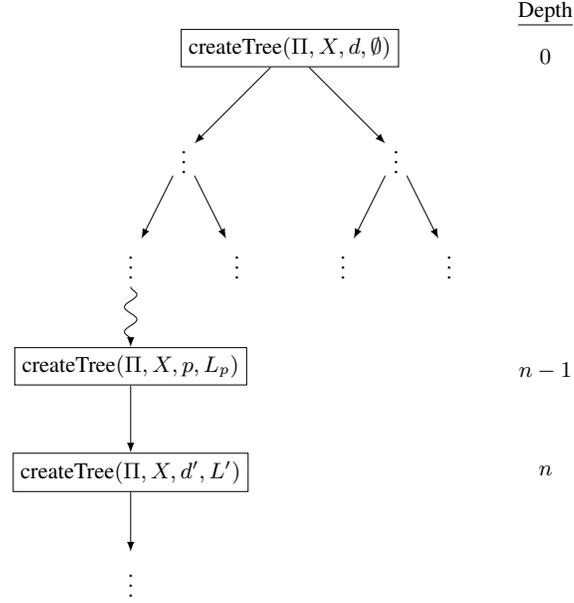

Fig. A 1: Part of the recursion tree for createTree$(\Pi, X, d, \emptyset)$.

**Case 1.** Assume that $d$ is an atom. Then, $p$ must be a rule. By Lemma 3, $L' = L_p$. Notice that the depth of the parent node is $n - 1$. Then, by the induction hypothesis, $L_p = anc_T(v)$ where $l(v) = p$. Also, by Lemma 4, $(v, v') \in E$ where $l(v') = d'$. Since $v$ is a rule vertex, $anc_T(v) = anc_T(v')$. Thus, $L' = anc_T(v')$ where $l(v') = d'$.

**Case 2.** Assume that $d$ is a rule. Then, $p$ must be an atom. By Lemma 3, $L' = L_p \cup \{p\}$. Notice that the depth of the parent node is $n - 1$. Then, by the induction hypothesis, $L_p = anc_T(v)$ where $l(v) = p$. Also, by Lemma 4, $(v, v') \in E$ where $l(v') = d'$. Since $v$ is an atom vertex, $anc_T(v) \cup \{l(v)\} = anc_T(v')$. Thus, $L' = anc_T(v')$ where $l(v') = d'$.
$\square$

*Proposition* 12 *(Soundness of Algorithm 2)*
Let $\Pi$ be a ground ASP program, $X$ be an answer set for $\Pi$, $d$ be an atom in $X$ and $T = \langle V, E, l, \Pi, X \rangle$ be the vertex-labeled tree returned by createTree$(\Pi, X, d, \emptyset)$. If $T$ is not empty, then $T$ is the and-or explanation tree for $d$ with respect to $\Pi$ and $X$.

*Proof of Proposition 12*

Suppose that $T$ is not empty. We want to show that $T$ is the and-or explanation tree for $d$ with respect to $\Pi$ and $X$. For that, $T$ must satisfy conditions $(i) - (iv)$ in Definition 2. As $T$ is not empty, conditions $(i)$, $(iii)$ and $(iv)$ hold due to Lemma 2. To complete the proof, we show condition $(ii)$ also holds in the sequel.

Let $S_v = \{(v, v') \mid (v, v') \in E, \, l(v') \in \Pi_{X, anc_T(v')}(l(v))\}$. Our goal is to show that for every atom vertex $v \in V$, $out_E(v) = S_v$. To do so, we show that $out_E(v) \subseteq S_v$ and that $S_v \subseteq out_E(v)$.

Let $v$ be an atom vertex in $V$. Now, we show that $out_E(v) \subseteq S$. Take an arbitrary element $(v, v') \in out_E(v)$. Then, $(v, v') \in E$. Throughout the algorithm edges are created at Lines 9 and 17. Since $v$ is an atom vertex, $(v, v')$ must be created at Line 9. Then, due to the condition at Line 5 and Lemma 5, $l(v') \in \Pi_{X, anc_E(v')}(l(v))$. Thus, $(v, v') \in S_v$. As the last step, we show that $S_v \subseteq out_E(v)$. Let $(v, v') \in S_v$ be an arbitrary element. Then, by the definition of $S_v$, $(v, v') \in E$. Thus, trivially, $(v, v') \in out_E(v)$.

$\square$

### Lemma 6

Let $\Pi$ be a ground ASP program, $X$ be an answer set for $\Pi$, $d$ be an atom in $X$, $T = \langle V_T, E_T, l, \Pi, X \rangle$ be the and-or explanation tree for $d$ with respect to $\Pi$ and $X$, $v$ be the root of $T$ and $\langle V, E, l, \Pi, X \rangle$ be the vertex-labeled tree returned by

$$\text{extractExp}(\Pi, X, V_T, l, v, E_T, W_T, \emptyset, min).$$

Then, for each $v' \in V$, we have

$$W_T(v') \leq min\{W_T(s) \mid s \in sibling_{E_T}(v')\}.$$

### Proof of Lemma 6

Let $v' \in V$. Vertices are added to $V$ at Line 8. Then, due to Line 7, $v'$ is a rule vertex. Also, each vertex added to $V$ corresponds to the $5^{th}$ parameter of the algorithm. Note that recursive calls are made at Lines 5 and 10. Since the $5^{th}$ parameter is a rule vertex only in the call at Line 5, the call where $v'$ is added to $V$ must be initiated at Line 5. Then, due to Line 3, $W_T(v') = min\{W_T(s) \mid s \in sibling_{E_T}(v')\}$.

$\square$

### Lemma 7

Let $\Pi$ be a ground ASP program, $X$ be an answer set for $\Pi$, $d$ be an atom in $X$, $T = \langle V_T, E_T, l, \Pi, X \rangle$ be the and-or explanation tree for $d$ with respect to $\Pi$ and $X$, $v$ be the root of $T$, and $\langle V, E, l, \Pi, X \rangle$ be the vertex-labeled tree returned by

$$\text{extractExp}(\Pi, X, V_T, l, v, E_T, W_T, \emptyset, min).$$

Let $v_1, v_2 \in V$. If $(v_1, v), (v, v_2) \in E_T$ for some $v \in V_T$, then $(v_1, v_2) \in E$.

### Proof of Lemma 7

Since $v_1 \in V$, it is added to $V$ at Line 8. Then, for each child of $v_1$, the algorithm is recursively called at Line 10. As $(v_1, v) \in E_T$, we make a call

$$\text{extractExp}(\Pi, X, V_T, l, v, E_T, W_T, v_1, min).$$

Inside that call, we add $(v_1, c)$ to $E$ (at Line 4) where $c$ is a minimum weighted child of $v$ (due to Line 3). As $(v, v_2) \in E_T$ and $v_2$ is a minimum weighted child of $v$ due to Lemma 6, $c$ is equal to $v_2$. Thus, $(v_1, v_2) \in E$.

$\square$

*Lemma 8*

Let $\Pi$ be a ground ASP program, $X$ be an answer set for $\Pi$, $d$ be an atom in $X$, $T = \langle V_T, E_T, l, \Pi, X \rangle$ be the and-or explanation tree for $d$ with respect to $\Pi$ and $X$, and *op* be a string *min*. Then, Algorithm 4 returns an explanation $\langle V, E, l, \Pi, X \rangle$ for $d$ with respect to $\Pi$ and $X$.

*Proof of Lemma 8*

Let $v$ be the root of $T$ and $S = \langle V, E, l, \Pi, X \rangle$ be the output of

$$\text{extractExp}(\Pi, X, V_T, l, v, E_T, W_T, \emptyset, min).$$

To prove that $S$ is an explanation for $d$ with respect to $\Pi$ and $X$, we need to show that there exists an explanation tree $\langle V', E', l, \Pi, X \rangle$ in $T$ which satisfies Conditions $(i)$ and $(ii)$ in Definition 4 of the paper. That is, the following hold.

  *(i)* $V = \{v \mid v$ is a rule vertex in $V'\}$;
  *(ii)* $E = \{(v_1, v_2) \mid (v_1, v), (v, v_2) \in E'$ for some atom vertex $v \in V'\}$.

To do so, we construct a vertex-labeled tree and show that it is an explanation tree in $T$ which satisfies above conditions. Thus, let us define a vertex-labeled tree $T' = \langle V', E', l, \Pi, X \rangle$ where

$$V' = V \cup \{v \mid v \in V_T \text{ s.t. } (v, v') \in E_T \text{ for some } v' \in V\} \tag{A1}$$

$$E' = \{(v, v') \mid v, v' \in V' \text{ s.t. } (v, v') \in E_T\} \tag{A2}$$

We now show that $T'$ is an explanation tree in $T$, i.e., $T'$ satisfies Conditions $(i)$–$(iv)$ in Definition 3.

$(i)$ Due to Line 8 and that $v \in V_T$, $V \subseteq V_T$. So, by (A1), $V' \subseteq V_T$. Also, by (A2), $E' \subseteq E_T$.

$(ii)$ In the first call of Algorithm 4, $v$ is the root of $T$. Then, at Line 5, the algorithm is called with a child $c$ of $v$. Note that $c$ is a rule vertex. Due to Line 8, for some $v' \in V$, $(v, v') \in E_T$.

$(iii)$ Let $v'$ be an atom vertex in $V'$. Then, by (A1) and (A2), $(v', v'') \in E'$ for some $v'' \in V$. This ensures that $deg'_E(v') \geq 1$. Assume that $deg'_E(v') > 1$. Then, for some $v''' \in V'$ ($v'' \neq v'''$), $(v', v''') \in E'$. Then, $v''' \in V$. This is not possible due to Line 3. Thus, $deg_E(v') = 1$.

$(iv)$ Let $v'$ be a rule vertex in $V'$. Then, by (A1) $v' \in V$ and $v$ is added into $V$ at Line 8. Due to Line 9 and (A1), every child $c$ of $v$ is in $V'$. Then, by (A2), $(v, c) \in E'$. That is, $out_{E_T}(v) \subseteq E'$.

As a last step, we show that $T'$ satisfies Conditions $(i)$ and $(ii)$ in Definition 4.

$(i)$ Note that every element in the set $\{v \mid v \in V_T \text{ s.t. } (v, v') \in E_T \text{ for some } v' \in V\}$ is an atom vertex. Then, by (A1), $V = \{v \mid v$ is a rule vertex in $V'\}$;

$(ii)$ Let $S = \{(v_1, v_2) \mid (v_1, v), (v, v_2) \in E'$ for some atom vertex $v \in V'\}$. We show that $E = S$.

Let $(v_1, v_2) \in E$. Then, $(v_1, v_2)$ is added into $E$ at Line 4. So, due to Lines 2 and 3, we know that there exists an atom vertex $v \in V_T$ such that $(v, v_2) \in E_T$. Then, by (A1), $v \in V'$ and, by (A2), $(v, v_2) \in E'$. Also, by Line 9, we know that $(v_1, v) \in E_T$. Then, by (A2), $(v_1, v) \in E'$. Thus, $(v_1, v_2) \in S$. That is, $E \subseteq S$.

Let $(v_1, v_2) \in S$. Then, for some atom vertex $v \in V'$, $(v_1, v), (v, v_2) \in E'$. By (A2), $v_1, v_2 \in V$ and $(v_1, v), (v, v_2) \in E_T$. Then, due to Lemma 7, $(v_1, v_2) \in E$. That is, $S \subseteq E$. $\qquad\square$

*Lemma 9*

Let $\Pi$ be a ground ASP program, $X$ be an answer set for $\Pi$, and $p$ be an atom in $X$. Let $T = \langle V, E, l, \Pi, X \rangle$ be the and-or explanation tree for $p$ with respect to $\Pi$ and $X$, $T' = \langle V', E', l, \Pi, X \rangle$ be an explanation tree in $T$ and $v$ be a rule vertex in $V'$. Then, the following inequality holds for the weight of $v$

$$W_T(v) \leq 1 + |\{u' \mid u' \in \mathit{des}_{T'}(v)\}|. \tag{A3}$$

*Proof of Lemma 9*

We prove the lemma by induction on the height of a rule vertex in the explanation tree.

*Base case:* Let $u$ be a rule vertex in $V'$ at height $0$. Then, $u$ is a leaf vertex. By the definition of the weight function, $W_T(u) = 1$. Then, (A3) holds.

*Induction step:* As an induction hypothesis, assume that for every rule vertex $i \in V'$ at height less than $n$, (A3) holds. We show that (A3) holds for every rule vertex $w \in V'$ at height $n + 1$. Let $w$ be a rule vertex at height $n + 1$. Then, by the definition of the weight function, $W_T(w) = 1 + \sum_{w' \in \mathit{child}_E(w)} W_T(w')$. Let $w'$ be a child of $w$. Note that $w'$ is an atom vertex. By the definition of an explanation tree, $w' \in V'$ and $w'$ has exactly one child $w'' \in V'$ which is a rule vertex. Then, by the definition of the weight function, $W_T(w') = \mathit{min}\{W_T(c) \mid c \in \mathit{child}_E(w')\}$ and thus $W_T(w') \leq W_T(w'')$. On the other hand, as the height of $w''$ is $n - 1$, by the induction hypothesis, $W_T(w'') \leq 1 + |\{d \mid d \in \mathit{des}_{T'}(w'')\}|$. Since $w'$ has exactly one child $w'' \in V'$, we have

$$1 + |\{d \mid d \in \mathit{des}_{T'}(w'')\}| = |\{u \mid u \in \mathit{des}_{T'}(w')\}|.$$

That is, $W_T(w') \leq |\{u \mid u \in \mathit{des}_{T'}(w')\}|$. Then, we have

$$\begin{aligned}
W_T(w) &= 1 + \textstyle\sum_{w' \in \mathit{child}_E(w)} W_T(w') \\
&\leq 1 + \textstyle\sum_{w' \in \mathit{child}_E(w)} |\{u \mid u \in \mathit{des}_{T'}(w')\}| \\
&= 1 + |\{u' \mid u' \in \mathit{des}_{T'}(w)\}|.
\end{aligned}$$

$\qquad\square$

*Lemma 10*
Let $\Pi$ be a ground ASP program, $X$ be an answer set for $\Pi$, $p$ be an atom in $X$, $T$ be the and-or explanation tree (with edges $E$) for $p$ with respect to $\Pi$ and $X$, $v$ be the root of $T$, and $T'$ be an explanation tree (with vertices $V'$) in $T$. Then,

$$W_T(v) \leq |\{u \mid u \text{ is a rule vertex in } V'\}|.$$

*Proof of Lemma 10*
We want to show that the weight of $v$, $W_T(v)$, is at most the number of rule vertices in $V'$. Note that $v$ is the root of $T'$ and there exists exactly one vertex $v' \in V'$ such that $v' \in child_E(v)$ (due to Definition 3). Then, we have

$$
\begin{aligned}
W_T(v) = \quad & min\{W_T(c) \mid c \in child_E(v)\} \quad && \text{(by Definition 6)} \\
\leq \quad & W_T(v') \quad && \text{(as } v' \in child_E(v)) \\
\leq \quad & 1 + |\{v'' \mid v'' \in des_{T'}(v')\}| \quad && \text{(by Lemma 9)} \\
= \quad & |\{u \mid u \text{ is a rule vertex in } V'\}|. \quad && \text{(as } v' \text{ is the only child of } v \text{ in } T')
\end{aligned}
$$

$\square$

*Lemma 11*
Let $\Pi$ be a ground ASP program, $X$ be an answer set for $\Pi$, $p$ be an atom in $X$, $T$ be the and-or explanation tree for $p$ with respect to $\Pi$ and $X$, $v$ be the root of $T$, and $\langle V, E, l, \Pi, X \rangle$ be an explanation for $p$ with respect to $\Pi$ and $X$. Then, $W_T(v) \leq |V|$.

*Proof of Lemma 11*
We show that the weight of $v$, $W_T(v)$, is at most $|V|$. By the definition of an explanation, there exists an explanation tree $T'$ (with vertices $V'$) of $T$ such that $|V| = |\{v' \mid v' \text{ is a rule vertex in } V'\}|$. By Lemma 10, $W_T(v) \leq |\{v' \mid v' \text{ is a rule vertex in } V'\}|$. Thus, $W_T(v) \leq |V|$.

$\square$

*Lemma 12*
Let $\Pi$ be a ground ASP program, $X$ be an answer set for $\Pi$, $p$ be an atom in $X$, $T = \langle V, E, l, \Pi, X \rangle$ be the and-or explanation tree for $p$ with respect to $\Pi$ and $X$, and $T' = \langle V', E', l, \Pi, X \rangle$ be an explanation tree in $T$. If we have

$$W_T(v) \leq min\{W_T(s) \mid s \in sibling_E(v)\}. \tag{A4}$$

for every rule vertex $v \in V'$, then the following holds for every rule vertex $v \in V'$.

$$W_T(v) = 1 + |\{u' \mid u' \in des_{T'}(v)\}|. \tag{A5}$$

*Proof of Lemma 12*
Assume that (A4) holds for every rule vertex $v \in V'$. Then, we prove the lemma by induction on the height of a rule vertex in the explanation tree.

*Base case:* Let $u$ be a rule vertex in $V'$ at height 0. Then, $u$ is a leaf vertex. By the definition of the weight function, $W_T(u) = 1$. As $u$ has no descendants, (A5) holds.

*Induction step:* As an induction hypothesis, assume that for every rule vertex $i \in V'$ at height less than $n$, (A5) holds. We show that (A5) holds for every rule vertex $w \in V'$ at height $n + 1$. Let $w \in V'$ be a rule vertex at height $n + 1$. Then, by the definition of the weight function, $W_T(w) = 1 + \sum_{w' \in child_E(w)} W_T(w')$. Let $w'$ be a child of $w$. Note that $w'$ is an atom vertex. By the definition of an explanation tree, $w' \in V'$ and $w'$ has exactly one child $w'' \in V'$, which is a rule vertex. Then, by the definition of the weight function, $W_T(w') = min\{W_T(c) \mid c \in child_E(w')\}$. Also, by (A4), $W_T(w'') \leq min\{W_T(s) \mid s \in sibling_E(w'')\}$. Then, $W_T(w'') = min\{W_T(c) \mid c \in child_E(w')\}$. Thus, $W_T(w') = W_T(w'')$. As the height of $w''$ is $n-1$, by the induction hypothesis, $W_T(w'') = 1 + |\{u \mid u \in des_{T'}(w'')\}|$. As $w''$ is the only child of $w'$, $W_T(w') = |\{u \mid u \in des_{T'}(w')\}|$. Then, we have

$$
\begin{aligned}
W_T(w) &= 1 + \sum_{w' \in child_E(w)} W_T(w') \\
&= 1 + \sum_{w' \in child_E(w)} |\{u \mid u \in des_{T'}(w')\}| \\
&= 1 + |\{u' \mid u' \in des_{T'}(w)\}|.
\end{aligned}
$$

□

*Lemma 13*

Let $\Pi$ be a ground ASP program, $X$ be an answer set for $\Pi$, $p$ be an atom in $X$, $T$ be the and-or explanation tree (with edges $E$) for $p$ with respect to $\Pi$ and $X$, $v$ be the root of $T$, and $T'$ be an explanation tree (with vertices $V'$) in $T$. If we have

$$W_T(v') \leq min\{W_T(s) \mid s \in sibling_E(v')\} \tag{A6}$$

for every rule vertex $v' \in V'$, then, the following holds

$$W_T(v) = |\{u \mid u \text{ is a rule vertex in } V'\}|.$$

*Proof of Lemma 13*

Assume that (A4) holds for every rule vertex $v' \in V'$. Then, we want to show that the weight of $v$, $W_T(v)$, is equal to the number of rule vertices in $V'$. Note that $v$ is the root of $T'$ and there exists exactly one vertex $v' \in V'$ such that $v' \in child_E(v)$ (due to Definition 3). Then, we have

$$
\begin{aligned}
W_T(v) &= min\{W_T(c) \mid c \in child_E(v)\} &&\text{(by Definition 6)} \\
&= W_T(v') &&\text{(by (A6))} \\
&= 1 + |\{u' \mid u' \in des_{T'}(v')\}| &&\text{(by Lemma 12)} \\
&= |\{u \mid u \text{ is a rule vertex in } V'\}|. &&\text{(as } v' \text{ is the only child of } v \text{ in } T')
\end{aligned}
$$

□

We are now ready to prove Proposition 3.

*Proposition* 3

Given a ground ASP program $\Pi$, an answer set $X$ for $\Pi$, and an atom $p$ in $X$, Algorithm 1 either finds a shortest explanation for $p$ with respect to $\Pi$ and $X$ or returns an empty vertex-labeled tree.

*Proof of Proposition 3*

Algorithm 1 has two return statements; Lines 6 and 8. At Line 8, it returns an empty vertex-labeled tree. We show that what Algorithm 1 returns at Line 6, $S = \langle V', E', l, \Pi, X \rangle$, is an explanation for $p$ with respect to $\Pi$ and $X$ and that there is no other explanation for $p$ with respect to $\Pi$ and $X$, with vertices $V''$, such that $|V''| < |V'|$.

Due to the condition at Line 2, the vertex-labeled tree found at Line 1 is not empty. Then, by Proposition 12, we know that $T$ is the and-or explanation tree for $p$ with respect to $\Pi$ and $X$. Then, by Lemma 8, we know that $\langle V', E', l, \Pi, X \rangle$ found at Line 5 is an explanation for $p$ with respect to $\Pi$ and $X$.

Now, suppose that there exists another explanation for $p$ with respect to $\Pi$ and $X$, with vertices $V''$, such that $|V''| < |V'|$. Let $v$ be the root of $T$. Then, by Lemma 11, $W_T(v) \leq |V''|$. Also, since $S$ is an explanation for $p$ with respect to $\Pi$ and $X$, there exists an explanation tree with vertices $V'''$ in $T$ such that $V' = \{u \mid u$ is a rule vertex in $V'''\}$. Due to Lemma 6, $W_T(v') \leq min\{W_T(s) \mid s \in sibling_E(v')\}$ for each $v' \in V'$. Then, by Lemma 13, $W_T(v) = |\{u \mid u$ is a rule vertex in $V'''\}|$. This implies that $|V'| \leq |V''|$. Since this is a contradiction, $S$ is a shortest explanation for $p$ with respect to $\Pi$ and $X$.

□

### A.1.3  Proof of Proposition 4 – Complexity of Algorithm 1

We prove Proposition 4 which shows that the time complexity of Algorithm 1 is exponential in the size of the given answer set.

*Proposition 4*

Given a ground ASP program $\Pi$, an answer set $X$ for $\Pi$, and an atom $p$ in $X$, the time complexity of Algorithm 1 is $O(|\Pi|^{|X|} \times |\mathcal{B}_\Pi|)$.

*Proof of Proposition 4*

In Algorithm 1, all the lines, expect 1, 4 and 5, take constant amount of time. At Lines 1, 4 and 5, three different algorithms are called. At Line 1, Algorithm 2 is called. This algorithm creates the and-or explanation tree recursively. As shown in Proof of Lemma 1, the branching factor of a vertex in the recursion tree is $O(max\{|X|, |\Pi|\})$ and the height of the tree is $O(|X|)$. Also, at Line 5, for an atom $d \in X$, we check whether a rule in $\Pi$ supports $d$ in $O(|\mathcal{B}_\Pi|)$. Thus, the time complexity of Algorithm 2, in the worst case, is $O(max\{|X|, |\Pi|\}^{|X|} \times |\mathcal{B}_\Pi|)$. As $|X| \leq |\Pi|$, it is $O(|\Pi|^{|X|} \times |\mathcal{B}_\Pi|)$. At Lines 4 and 5, Algorithm 3 and Algorithm 4 are called, respectively. Algorithm 3 and Algorithm 4 simply traverse the tree $T$ created by Algorithm 2 (cf. Lines 2 and 6 in Algorithm 3, and Lines 5 and 10 in Algorithm 4. By Proposition 12, we know that $T$ is the and-or explanation tree for $p$ with respect to $\Pi$ and $X$. Since the height of $T$ is $O(|X|)$ and the branching factor of a vertex in $T$ is $max\{|X|, |\Pi|\}$, the time complexity of Algorithm 2 dominates the time complexities of others. Thus, the time complexity of Algorithm 1, in the worst case, is $O(|\Pi|^{|X|} \times |\mathcal{B}_\Pi|)$.

□

### *A.2 Generating $k$ Different Explanations*

In Section 6 of the paper, we have first analyzed three properties of Algorithm 5, namely termination, soundness and complexity, resulting in Propositions 5, 6, and 8. Then, we show some characteristics of its output, resulting in Proposition 7, Corollary 1 and Corollary 2. In the following, we show the proofs of these results.

#### *A.2.1 Proof of Proposition 5 – Termination of Algorithm 5*

We now prove Proposition 5 which shows that Algorithm 5 terminates.

*Proposition 5*

Given a ground ASP program $\Pi$, an answer set $X$ for $\Pi$, an atom $p$ in $X$, and a positive integer $k$, Algorithm 5 terminates.

*Proof of Proposition 5*

Algorithm 5 calls Algorithm 2 at Line 2. By Lemma 1, we know that Algorithm 2 terminates. Then, to show that Algorithm 5 terminates, we need to show that the loop between Lines 4–10 terminates. Due to Line 4, the loop iterates at most $k$ times. If it iterates less than $k$ times, it means that it is terminated at Line 6. Thus, assume that it iterates $k$ times. Then, it is enough to show that every iteration of the loop terminates. Consider the $i^{th}$ ($1 \leq i \leq k$) iteration of the loop. First, at Line 5, Algorithm 6 is called. Observe that this algorithm simply traverses the and-or explanation tree $T$ created at Line 2 (cf. Lines 2, 3, 8 and 9 in Algorithm 6). Since $T$ is finite, Algorithm 6 terminates. Next, Algorithm 4 is called at Line 7. Similar to Algorithm 6, this algorithm also simply traverses a portion of $T$ (cf. Lines 3, 5, 9 and 10 in Algorithm 4). Hence, Algorithm 4 also terminates. As the rest of the loop consists of some assignment statements, the $i^{th}$ ($1 \leq i \leq k$) iteration of the loop terminates. Since every iteration of the loop terminates, Algorithm 5 terminates.

$\square$

#### *A.2.2 Proof of Proposition 6 – Soundness of Algorithm 5*

Before showing the soundness property of Algorithm 5, we provide some necessary lemmas.

*Proposition* 13 *(Completeness of Algorithm 2)*

Let $\Pi$ be a ground ASP program, $X$ be an answer set for $\Pi$, $p$ be an atom in $X$. Let $T$ be the and-or explanation tree $p$ with respect to $\Pi$ and $X$. Then, createTree$(\Pi, X, p, \{\})$ returns $T$.

*Proof of Proposition 13*

Let $\langle V, E, l, \Pi, X \rangle$ be the output of Algorithm 2. By Condition $(i)$ in Definition 2, the root of $T$ is an atom vertex with label $p$. Since $p$ is in $X$ and $L = \emptyset$ at the beginning of Algorithm 2, a vertex $v$ with label $p$ is created at Line 3 and added into $V$ at Line 4.

Take an atom vertex $v \in V$. Then, there should be an out-going edge $(v, v')$ of $v$ such that $l(v') \in \Pi_{X, anc_T(v')}(l(v))$ (due to Condition $(ii)$ in Definition 2). Note that, the set $L$ in Algorithm 2 is essentially $anc_T(v')$, and $l(v') \in \Pi_{X, L}(l(v))$ is checked at Line 5.

Moreover, we need to ensure that $v'$ is a rule vertex. This condition is checked at Line 6 by recursive calls.

Take a rule vertex $v \in V$. Then, for every atom $a$ in $B^+(l(v))$, there should be an outgoing edge $(v, v')$ of $v$ such that $l(v') = a$ (due to Condition $(iii)$ in Definition 2). This is satisfied by the condition at Line 13. Moreover, we need to make sure that $v'$ is atom vertex. For that, there is a recursive call at Line 14.

Take a leaf vertex $v \in V$. Then, $v$ must be a rule vertex (due to Condition $(iv)$ in Definition 2). Assume that $v$ is an atom vertex. Thus, no out-going edge is defined for $v$ at Line 9. Then, when the loop at Line 5 terminates, the condition at Line 10 is satisfied. This implies that $v$ is not in $V$. As it is a contradiction, $v$ must be a rule vertex.
□

By this proposition and Proposition 1, we know that Algorithm 2 returns the and-or explanation tree. Thus, at Line 1 of Algorithm 5, the and-or explanation tree for $p$ with respect to $\Pi$ and $X$ is created. We prove our statements by keeping this in mind.

*Lemma 14*
Let $\Pi$ be a ground ASP program, $X$ be an answer set for $\Pi$, $p$ be an atom in $X$, and $k$ be a positive integer. Let $n$ be the number of different explanations for $p$ with respect to $\Pi$ and $X$. Then, for the root $v_r$ of $T$, at each iteration $i$ $(1 \leq i \leq min\{n, k\})$ of the loop in Algorithm 5, $W_{T, R_{i-1}}(v_r) = |RVertices(K_i) \backslash R_{i-1}|$.

*Proof of Lemma 14*
Consider the $i^{th}$ $(1 \leq i \leq min\{n, k\})$ iteration of the loop. At Line 5, we call Algorithm 6 and calculate $W_{T, R_{i-1}}$ for every vertex $v$ in $V$. According to Algorithm 6, for an atom vertex $v \in V$, due to Line 4, $W_{T, R_{i-1}}(v) = max\{W_{T, R_{i-1}}(v') \mid v' \in child_E(v)\}$ and for a rule vertex $u$, due to Lines 6–9, $W_{T, R_{i-1}}(u) = x_u + \sum_{u' \in child_E(u)} W_{T, R_{i-1}}(u')$ where $x_u = 1$ if $u \notin R_{i-1}$, $x_u = 0$ otherwise. Let $v$ be an atom vertex in $V$. Let $\{v'_1, \ldots, v'_{v_z}\}$ be a set of rule vertices that "contribute" to $W_{T, R_{i-1}}(v)$, i.e., rule vertices that appear in the expanded formula of $W_{T, R_{i-1}}(v)$. This implies that $W_{T, R_{i-1}} = x_{v'_1} + \ldots + x_{v'_{v_z}}$ where $x_{v'_j} = 1$ if $v'_j \notin R_{i-1}$, $x_{v'_j} = 0$ otherwise, for $1 \leq j \leq v_z$. Also, at Line 7 in Algorithm 5, we extract an explanation using Algorithm 4. Then, at Line 8, we assign the output $\langle V', E', l, \Pi, X \rangle$ of Algorithm 4 to $K_i$. Let $RVertices(K_i) = \{v_1, \ldots, v_z\}$. Observe in Algorithm 4 that for every atom vertex, we process its maximum weighted child recursively (Line 3), and for every rule vertex we choose every child of it recursively (Line 9). Then, due to the observation and the calculation of $W_{T, R_{i-1}}$, $RVertices(K_i)$ is a set of rule vertices that contribute to $W_{T, R_{i-1}}(v_r)$. Then, $W_{T, R_{i-1}}(v_r) = x_1 + \ldots + x_z$ where $x_j = 1$ if $v_j \notin R_{i-1}$, $x_j = 0$ otherwise, for $1 \leq j \leq z$. That is, $W_{T, R_{i-1}}(v_r) = |RVertices(K_i) \backslash R_{i-1}|$.
□

Now, we prove Proposition 6.

*Proposition 6*
Let $\Pi$ be a ground ASP program, $X$ be an answer set for $\Pi$, $p$ be an atom in $X$, and $k$ be a positive integer. Let $n$ be the number of different explanations for $p$ with respect to $\Pi$ and $X$. Then, Algorithm 5 returns $min\{n, k\}$ different explanations for $p$ with respect to $\Pi$ and $X$.

*Proof of Proposition 6*

First, assume that $n \geq k$. Then, we show that Algorithm 5 returns at Line 11 a set $K$ of $k$ different explanations for $p$ with respect to $\Pi$ and $X$. Note that an element is added into $K$ at each iteration of the loop between Lines 4–10. Since that loop iterates $k$ times, we need to show that two properties hold: *(i)* for every iteration $i$ ($1 \leq i \leq k$), $K_i$ (the element added into $K$ at the $i^{th}$ iteration) is an explanation for $p$ with respect to $\Pi$ and $X$. *(ii)* for all iterations $i, j$ ($1 \leq i < j \leq k$), $K_i$ and $K_j$ are different. In the following, we consider those two properties.

*(i)* For every iteration $i$ ($1 \leq i \leq k$), $K_i$ is formed at Line 8. According to Line 7, $K_i$ is an output of Algorithm 4. Then, due to Lemma 8, we conclude that $K_i$ is an explanation for $p$ with respect to $\Pi$ and $X$.

*(ii)* Assume otherwise. Then, there exists $i, j$ ($1 \leq i < j \leq k$) such that $K_i = K_j$. Consider the $j^{th}$ iteration of the loop. At Line 5, we call Algorithm 6 and calculate $W_{T, R_{j-1}}$ for every vertex $v$ in $V$. Then, at Line 7, we extract an explanation using Algorithm 4. Let $K_j = \langle V', E', l, \Pi, X \rangle$, $v_r$ be the root of $\langle V', E' \rangle$ and $RVertices(K_j) = \{v_1, \ldots, v_l\}$. Then, by Lemma 14, $W_{T, R_{j-1}}(v_r) = x_1 + \ldots + x_l$ where $x_z = 1$ if $v_z \notin R_{j-1}$, $x_z = 0$ otherwise, for $1 \leq z \leq l$. Since $K_i = K_j$, $\{v_1, \ldots, v_l\} = RVertices(K_i)$. Due to Line 10, we know that $RVertices(K_i) \subseteq R_{j-1}$. Then, for $1 \leq z \leq l$, $v_z \in R_{j-1}$. Thus, $W_{T, R_{j-1}}(v_r) = 0$. This implies that for every vertex $v \in V$, $W_{T, R_{j-1}}(v) = 0$. That is, every rule vertex in $V$ is also in $R_{j-1}$. However, as we are at the $j^{th}$ iteration and at each iteration one explanation is computed, $R_{j-1}$ might contain rule vertices for at most $j - 1$ explanations. We know that $n \geq k$. Thus, for some $v' \in V$, the following should hold: $v' \notin R_{j-1}$. As we reach a contradiction, $K_i \neq K_j$.

Now, assume that $n < k$. Then, we show that Algorithm 5 returns at Line 6 a set $K$ of $n$ different explanations for $p$ with respect to $\Pi$ and $X$. Note that at the end of $n^{th}$ iteration of the loop, $K$ contains $n$ different explanations (due to the same reasoning above). Then, in the next iteration, $W_{T, R_{n+1}}(v) = 0$. This is because $R_n$ contains the rule vertices of $n$ different explanations and the total number of explanations for $p$ with respect to $\Pi$ and $X$ is $n$. Thus, the condition at Line 6 is satisfied and $n$ different explanations are returned.

$\square$

### A.2.3 *Some Properties of Algorithm 5*

Before presenting some characteristics of the output of Algorithm 5, we provide some necessary definitions and lemmas.

*Definition 18* ($W_{T, R}$)

Let $\Pi$ be a ground ASP program, $X$ be an answer set for $\Pi$, $p$ be an atom in $X$, $T = \langle V, E, l, \Pi, X \rangle$ be the and-or explanation tree for $p$ with respect to $\Pi$ and $X$, and $R$ be a set of rule vertices in $V$. Then, $W_{T, R}$ is a function that maps vertices in $V$ to nonnegative

integers as follows.

$$W_{T,R}(v) = \begin{cases} max\{W_{T,R}(v') \mid v' \in child_E(v)\} & \text{if } v \text{ is an atom vertex;} \\ \sum_{v' \in child_E(v)} W_{T,R}(v') & \text{if } v \text{ is a rule vertex and } v \in R; \\ 1 + \sum_{v' \in child_E(v)} W_{T,R}(v') & \text{otherwise.} \end{cases}$$

*Lemma 15*

Let $\Pi$ be a ground ASP program, $X$ be an answer set for $\Pi$, and $p$ be an atom in $X$. Let $T = \langle V, E, l, \Pi, X \rangle$ be the and-or explanation tree for $p$ with respect to $\Pi$ and $X$, $T' = \langle V', E', l, \Pi, X \rangle$ be an explanation tree in $T$ and $R$ be a set of rule vertices in $V$. Then, for every rule vertex $v$ in $V'$, the following holds

$$W_{T,R}(v) \geq \begin{cases} 1 + |\{u' \mid u' \in (des_{T'}(v)\backslash R)\}| & \text{if } v \notin R; \\ |\{u' \mid u' \in (des_{T'}(v)\backslash R)\}| & \text{otherwise.} \end{cases} \tag{A7}$$

*Proof of Lemma 15*

We prove the lemma by induction on the height of a rule vertex in the explanation tree.

*Base case:* Let $v$ be a rule vertex in $V'$ at height $0$. Then, $v$ is a leaf vertex. By Definition 18, $W_{T,R}(v) = x_v$ where $x_v = 1$ if $v \notin R$, $x_v = 0$ otherwise. Then, (A7) holds.

*Induction step:* As an induction hypothesis, assume that for every rule vertex $i \in V'$ at height less than $n+1$, (A7) holds. We show that (A7) holds for every rule vertex $v \in V'$ at height $n+1$. Let $v$ be a rule vertex in $V'$ at height $n+1$. Then, by Definition 18, $W_{T,R}(v) = x_v + \sum_{v' \in child_E(v)} W_{T,R}(v')$ where $x_v = 1$ if $v \notin R$, $x_v = 0$ otherwise. Let $v'$ be a child of $v$. Note that $v'$ is an atom vertex. By Conditions $(iii)$ and $(iv)$ in Definition 3, $v' \in V'$ and $v'$ has exactly one child $v'' \in V'$ which is a rule vertex. Then, by Definition 18, $W_{T,R}(v') = max\{W_{T,R}(c) \mid c \in child_E(v')\}$ and thus $W_{T,R}(v') \geq W_{T,R}(v'')$. On the other hand, as the height of $v''$ is $n-1$, by the induction hypothesis, (A7) holds for $v''$. Thus, we obtain the following.

$$W_{T,R}(v') \geq \begin{cases} 1 + |\{u' \mid u' \in (des_{T'}(v'')\backslash R)\}| & \text{if } v'' \notin R; \\ |\{u' \mid u' \in (des_{T'}(v'')\backslash R)\}| & \text{otherwise.} \end{cases} \tag{A8}$$

Since $v'$ has exactly one child $v'' \in V'$, we derive

$$\{v''\} \cup \{d \mid d \in des_{T'}(v'')\} = |\{u \mid u \in des_{T'}(v')\}|. \tag{A9}$$

Then, due to (A8) and (A9),

$$W_{T,R}(v') \geq |\{u \mid u \in (des_{T'}(v')\backslash R)\}|. \tag{A10}$$

Then, to conclude the proof, we consider two cases.

**Case 1.** Suppose that $v \notin R$. Then, we can derive the following.

$$\begin{aligned} W_{T,R}(v) = \quad & 1 + \sum_{v' \in child_E(v)} W_{T,R}(v') && \text{(by Definition 18)} \\ \geq \quad & 1 + \sum_{v' \in child_E(v)} |\{u \mid u \in (des_{T'}(v')\backslash R)\}| && \text{(by (A10))} \\ = \quad & 1 + |\{u' \mid u' \in (des_{T'}(v)\backslash R)\}|. \end{aligned}$$

(as every child of $v$ is an atom vertex)

**Case 2.** Suppose that $v \in R$. Then, we can derive the following.

$$
\begin{aligned}
W_{T,R}(v) &= \textstyle\sum_{v' \in child_E(v)} W_{T,R}(v') && \text{(by Definition 18)} \\
&\geq \textstyle\sum_{v' \in child_E(v)} |\{u \mid u \in (des_{T'}(v') \backslash R)\}| && \text{(by (A10))} \\
&= |\{u' \mid u' \in (des_{T'}(v) \backslash R)\}|. \\
&&& \text{(as every child of $v$ is an atom vertex)}
\end{aligned}
$$

□

*Lemma 16*

Let $\Pi$ be a ground ASP program, $X$ be an answer set for $\Pi$, $p$ be an atom in $X$, $T = \langle V, E, l, \Pi, X \rangle$ be the and-or explanation tree for $p$ with respect to $\Pi$ and $X$, $v$ be the root of $T$, $T'$ be an explanation tree (with vertices $V'$) in $T$, and $R$ be a set of rule vertices in $V$. Then,

$$W_{T,R}(v) \geq |\{u \mid u \in (RVertices(T') \backslash R)\}|.$$

*Proof of Lemma 16*

We want to show that $W_{T,R}(v)$ is equal to at least the number of rule vertices in $V'$ but $R$. Recall that $v$ is the root of $T'$ and there exists exactly one vertex $v' \in V'$ such that $v' \in child_E(v)$ (due to Condition $(iii)$ in Definition 3). Then, we consider two cases.

*Case 1.* Assume that $v' \notin R$. Then, we can derive the following.

$$
\begin{aligned}
W_{T,R}(v) &= max\{W_{T,R}(c) \mid c \in child_E(v)\} && \text{(by Definition 18)} \\
&\geq W_{T,R}(v') && \text{(as $v' \in child_E(v)$)} \\
&\geq 1 + |\{v'' \mid v'' \in (des_{T'}(v') \backslash R)\}| && \text{(by Lemma 15)} \\
&= |\{u \mid u \in (RVertices(T') \backslash R)\}|. && \text{(as $v'$ is the only child of $v$ in $T'$)}
\end{aligned}
$$

*Case 2.* Assume that $v' \in R$. Then, we can derive the following.

$$
\begin{aligned}
W_{T,R}(v) &= max\{W_{T,R}(c) \mid c \in child_E(v)\} && \text{(by Definition 18)} \\
&\geq W_{T,R}(v') && \text{(as $v' \in child_E(v)$)} \\
&\geq |\{v'' \mid v'' \in (des_{T'}(v') \backslash R)\}| && \text{(by Lemma 15)} \\
&= |\{u \mid u \in (RVertices(T') \backslash R)\}|. && \text{(as $v'$ is the only child of $v$ in $T'$)}
\end{aligned}
$$

□

*Lemma 17*

Let $\Pi$ be a ground ASP program, $X$ be an answer set for $\Pi$, $p$ be an atom in $X$, $T$ be the and-or explanation tree (with vertices $V$) for $p$ with respect to $\Pi$ and $X$, $v$ be the root of $T$, $T' = \langle V', E', l, \Pi, X \rangle$ be an explanation for $p$ with respect to $\Pi$ and $X$, and $R$ be a set of rule vertices in $V$. Then, $W_{T,R}(v) \geq |RVertices(T') \backslash R|$.

*Proof of Lemma 17*
We show that $W_{T,R}(v)$ is equal to at least $|RVertices(T')\backslash R|$. By Definition 4, there exists an explanation tree $T''$ (with vertices $V''$) in $T$ such that $V' = \{v' \mid v'$ is a rule vertex in $V''\}$. That is, $RVertices(T') = RVertices(T'')$. By Lemma 16, we know that $W_{T,R}(v) \geq |\{u \mid u \in (RVertices(T'')\backslash R)\}|$. Thus, we conclude that $W_{T,R}(v) \geq |RVertices(T')\backslash R|$.
□

We can now prove our main result which simply indicates that at each iteration $i$ of the loop in Algorithm 5 the distance $\Delta_D(R_{i-1}, K_i)$ is maximized.

*Proposition* 7
Let $\Pi$ be a ground ASP program, $X$ be an answer set for $\Pi$, $p$ be an atom in $X$, and $k$ be a positive integer. Let $n$ be the number of explanations for $p$ with respect to $\Pi$ and $X$. Then, at the end of each iteration $i$ ($1 \leq i \leq min\{n, k\}$) of the loop in Algorithm 5, $\Delta_D(R_{i-1}, RVertices(K_i))$ is maximized, i.e., there is no other explanation $K'$ such that $\Delta_D(R_{i-1}, RVertices(K_i)) < \Delta_D(R_{i-1}, RVertices(K'))$.

*Proof of Proposition 7*
The proof is by contradiction. Assume that there exists an explanation $K'$ such that $\Delta_D(R_{i-1}, RVertices(K_i)) < \Delta_D(R_{i-1}, RVertices(K'))$. That is, $|RVertices(K')\backslash R_{i-1}| > |RVertices(K_i)\backslash R_{i-1}|$. Let $v_r$ be the root of $T$. Then, by Lemma 17, $W_{T,R_{i-1}}(v_r) \geq |RVertices(K')\backslash R_{i-1}|$. Also, by Lemma 14, we know that $W_{T,R_{i-1}}(v_r) = |RVertices(K_i)\backslash R_{i-1}|$. Therefore, we obtain that $|RVertices(K_i)\backslash R_{i-1}| \geq |RVertices(K')\backslash R_{i-1}|$. But, that contradicts the assumption $|RVertices(K')\backslash R_{i-1}| > |RVertices(K_i)\backslash R_{i-1}|$. Thus, there is no $K'$ such that $\Delta_D(R_{i-1}, RVertices(K_i)) < \Delta_D(R_{i-1}, RVertices(K'))$, i.e., $\Delta_D(R_{i-1}, RVertices(K_i))$ is maximized.
□

Now, we provide the proof of the corollary that shows how to compute longest explanations.

*Corollary* 1
Let $\Pi$ be a ground ASP program, $X$ be an answer set for $\Pi$, $p$ be an atom in $X$, and $k = 1$. Then, Algorithm 5 computes a longest explanation for $p$ with respect to $\Pi$ and $X$.

*Proof of Corollary 1*
Since $k = 1$, the loop in Algorithm 5 iterates once. At that iteration, by Proposition 6, we know that an explanation $K_1$ for $p$ with respect to $\Pi$ and $X$ is computed. By Proposition 7, we also know that $\Delta_D(R_0, RVertices(K_1))$ is maximized. That is, there exists no explanation $K'$ for $p$ with respect to $\Pi$ and $X$ such that $|RVertices(K')\backslash R_0| > |RVertices(K_1)\backslash R_0|$. Note that $R_0$ is an empty set. Therefore, there exists no explanation $K'$ for $p$ with respect to $\Pi$ and $X$ such that $|RVertices(K')| > |RVertices(K_1)|$. As every vertex of an explanation is a rule vertex, we conclude that $K_1$ is a longest explanation for $p$ with respect to $\Pi$ and $X$.
□

Next, we indicate the proof of the corollary that shows Algorithm 5 computes $min\{n,k\}$ different explanations such that for every $i$ $(1 \leq i \leq min\{n,k\})$ the $i^{th}$ explanation is the most distant explanation from the previously computed $i-1$ explanations.

*Corollary* 2

Let $\Pi$ be a ground ASP program, $X$ be an answer set for $\Pi$, $p$ be an atom in $X$, and $k$ be a positive integer. Let $n$ be the number of explanations for $p$ with respect to $\Pi$ and $X$. Then, Algorithm 5 computes $min\{n,k\}$ different explanations $K_1, \ldots, K_{min\{n,k\}}$ for $p$ with respect to $\Pi$ and $X$ such that for every $j$ $(2 \leq j \leq min\{n,k\})$ $\Delta_D(\bigcup_{z=1}^{j-1} RVertices(K_z), K_j)$ is maximized.

*Proof of Corollary 2*

By Proposition 6, we know that $K_1, \ldots, K_{min\{n,k\}}$ are $min\{n,k\}$ different explanations for $p$ with respect to $\Pi$ and $X$. Also, by Proposition 7, for every $i$ $(2 \leq i \leq min\{n,k\})$, we obtain that $\Delta_D(R_{i-1}, K_i)$ is maximized. Due to Lines 1 and 10 of Algorithm 5, $R_{i-1} = \bigcup_{j=0}^{i-1} RVertices(K_j)$. Since $R_0$ is an empty set, we conclude that $\Delta_D(\bigcup_{j=1}^{i-1} RVertices(K_j), K_i)$ is maximized.

□

### *A.2.4 Proof of Proposition 8 – Complexity of Algorithm 5*

We prove Proposition 8 which shows that the time complexity of Algorithm 5 is exponential in the size of the given answer set.

*Proposition* 8

Given a ground ASP program $\Pi$, an answer set $X$ for $\Pi$, an atom $p$ in $X$ and a positive integer $k$, the time complexity of Algorithm 5 is $O(k \times |\Pi|^{|X|+1} \times |\mathcal{B}_\Pi|)$.

*Proof of Proposition 8*

Algorithm 5 calls Algorithm 2 at Line 2. In the proof of Proposition 4, it is shown that the worst case time complexity of Algorithm 2 is $O(|\Pi|^{|X|} \times \mathcal{B}_\Pi)$. Moreover, Algorithm 5 has a loop between Lines 4–10, which iterates at most $k$ times. In every iteration of the loop, Algorithm 6 and Algorithm 4 are called at Lines 5 and 7. Algorithm 6 simply traverses the and-or explanation tree $T$ recursively (cf. Lines 2, 3, 8 and 9 in Algorithm 6). The height of $T$ is $O(|X|)$ and the branching factor of a vertex in $T$ is $O(|\Pi|)$. Also, at Line 8 in Algorithm 6, we check whether a rule vertex in $V$ is in $R$. As $R$ is a subset of the rule vertices in $V$, this check can be done in $O(|\Pi| \times |\mathcal{B}_\Pi|)$ time. Thus, the time complexity of Algorithm 6, in the worst case, is $O(|\Pi|^{|X|} \times |\Pi| \times |\mathcal{B}_\Pi|)$. Similar to Algorithm 6, Algorithm 4 just traverses a portion of $T$ (cf. Lines 3, 5, 9 and 10 in Algorithm 4). Thus, the time complexity of every iteration of the loop in Algorithm 5 is $O(|\Pi|^{|X|+1} \times |\mathcal{B}_\Pi|)$. As the loop iterates at most $k$ times, the time complexity of Algorithm 5, in the worst case, is $O(k \times |\Pi|^{|X|+1} \times |\mathcal{B}_\Pi|)$.

□

### *A.3 Relations between Explanations and Justifications*

In Section 10 of the paper, we have related explanations to justifications, resulting in Propositions 10 and 11. In the following, we show the proofs of these results.

*A.3.1 Proof of Proposition 10 – Soundness of Algorithm 7*

In the proof of Proposition 10, the idea is to show that Algorithm 7 returns at Line 17 an explanation tree $T'$ in the and-or explanation tree for $p$ with respect to $\Pi$ and $X$. That is, $T'$ satisfies Conditions $(i)$–$(iv)$ in Definition 3. Before providing the proof of Proposition 10, we consider some useful lemmas and corollaries.

*Lemma 18*

Let $\Pi$ be a ground normal ASP program, $X$ be an answer set for $\Pi$, $p$ be an atom in $X$, $U$ be an assumption in $Assumptions(\Pi, X)$, $G = (V, E)$ be an offline justification of $p^+$ with respect to $X$ and $U$, $T = \langle V', E', l, \Pi, X \rangle$ be an output of Algorithm 7 and $P = \langle v_1, \ldots, v_n \rangle$ be a path in $T$. Take any three consecutive elements $v_i, v_{i+1}$ and $v_{i+2}$ in $P$ such that $v_i$ and $v_{i+2}$ are atom vertices and $v_{i+1}$ is a rule vertex. Then, $(l(v_i)^+, l(v_{i+2})^+, +) \in E$ holds.

*Proof of Lemma 18*

As $v_i$ is an atom vertex, its out-going edges are formed at Line 15 of Algorithm 7. Then, due to Lines 13 and 14, $v_{i+1}$ is a rule vertex such that $B(l(v_{i+1})) = support(l(v_i)^+, G)$. As $v_{i+1}$ is a rule vertex, its out-going edges are formed at Line 10. Then, due to Lines 8 and 9, $v_{i+2}$ is an atom vertex where $l(v_{i+2}) \in B^+(l(v_{i+1}))$. Since $B(l(v_{i+1})) = support(l(v_i)^+, G)$ and $l(v_{i+2}) \in B^+(l(v_{i+1}))$, by Definition 10, $(l(v_i)^+, l(v_{i+2})^+, +) \in E$ holds. □

*Corollary* 3

Let $\Pi$ be a ground normal ASP program, $X$ be an answer set for $\Pi$, $p$ be an atom in $X$, $U$ be an assumption in $Assumptions(\Pi, X)$, $G = (N, E)$ be an offline justification of $p^+$ with respect to $X$ and $U$, $T = \langle V', E', l, \Pi, X \rangle$ be an output of Algorithm 7 and $P = \langle v_1, \ldots, v_n \rangle$ be a path in $T$ where $v_1$ and $v_n$ are atom vertices. Then, $l(v_n)^+$ is reachable from $l(v_1)^+$ by a positive path in $G$.

*Proof of Corollary 3*

Note that an edge in $E'$ is either defined from an atom vertex to a rule vertex or vice versa due to Lines 10 and 15 of Algorithm 7. By this observation, in $P$, as $v_1$ is an atom vertex, $v_i$ is an atom vertex (resp. rule vertex) if $i$ is an odd number (resp. a even number). Then, by Lemma 18, for $1 \leq i \leq n - 2$ and $i \mod 2 \neq 0$ (i.e., $i$ is an odd number), $(l(v_i)^+, l(v_{i+2})^+, +) \in E$. Thus, there exists a positive path $\langle l(v_1)^+, l(v_3)^+, l(v_5)^+, \ldots, l(v_{n-2}^+), l(v_n)^+ \rangle$ in $G$. That is, $l(v_n)^+$ is reachable from $l(v_1)^+$ by a positive path in $G$. □

*Lemma 19*

Let $\Pi$ be a ground normal ASP program, $X$ be an answer set for $\Pi$, $p$ be an atom in $X$, $U$ be an assumption in $Assumptions(\Pi, X)$, $G = (N, E)$ be an offline justification of $p^+$ with respect to $X$ and $U$, $T = \langle V', E', l, \Pi, X \rangle$ be an output of Algorithm 7 and $v$ be an atom vertex in $V'$ such that $v$ is the first element added into the queue $Q$ in Algorithm 7. Consider a sequence $S = \langle v_1 = v, v_2, \ldots, v_n \rangle$ of $n$ elements where $v_i \in V'$ for $1 \leq i \leq n$

such that $v_{j+1}$ is added into $Q$ right after $v_j$ for $1 \le j < n$. Then, every vertex $v_i \in V'$ ($1 < i \le n$) is reachable from $v$ by a path in $T$.

### Proof of Lemma 19

The proof is by induction on the length of $S$.

*Base case:* Assume that $S$ has two elements, i.e., $S = \langle v_1 = v, v_2 \rangle$. Since $v$ is the first vertex added into $Q$ by definition, it is the first vertex dequeued from $Q$ at Line 5. Then, since $v$ is an atom vertex by definition, the second vertex is added into $Q$ at Line 16. By definition of $S$, $v_2$ is the second vertex added into $Q$. Thus, by Line 15, $(v, v_2) \in E'$, i.e., $v_2$ is reachable from $v$ by a path in $T$.

*Induction step:* As an induction hypothesis, assume that for a sequence $S' = \langle v_1 = v, v_2, \ldots, v_k \rangle$ of $k$ elements, where $v_i \in V'$ for $1 \le i \le k$ such that $v_{j+1}$ is added into $Q$ right after $v_j$ for $1 \le j < k$, every vertex $v_l \in V$ ($1 < l \le k$) is reachable from $v$ by a path in $T$. Let $S'' = \langle v_1 = v, v_2, \ldots, v_{k+1} \rangle$ be a sequence of $k + 1$ elements, where $v_i \in V'$ for $1 \le i \le k + 1$ such that $v_{j+1}$ is added into $Q$ right after $v_j$ for $1 \le j < k + 1$. We show that $v$ reaches every vertex in $S''$ by using edges in $E'$. Consider the subsequence $S''_{sub}$ of $S''$ that consists of the first $k$ elements of $S''$, i.e., a prefix of $S''$ with length $k$. By the induction hypothesis, every vertex $v_i \in S''_{sub}$ is reachable from $v$ by a path in $T$. Now, we show that $v_{k+1}$ is reachable from $v$ by a path in $T$. We consider two cases.

*Case 1.* Assume that $v_{k+1}$ is an atom vertex. Then, $v_{k+1}$ is added into $V'$ at Line 11. So, by Line 10, there exists a vertex $v'$ such that $(v', v_{k+1}) \in E'$. But, due to Line 5, $v'$ must be added into $Q$ prior to $v_{k+1}$. That is, $v' \in S''_{sub}$. So, by the induction hypothesis, $v$ reaches $v'$ by a path $P$ in $T$. Thus, by $P$ and $(v', v_{k+1})$, $v_{k+1}$ is reachable from $v$ by a path in $T$.

*Case 2.* Assume that $v_{k+1}$ is a rule vertex. Then, $v_{k+1}$ is added into $V'$ at Line 16. So, by Line 15, there exists a vertex $v'$ such that $(v', v_{k+1}) \in E'$. But, due to Line 5, $v'$ must be added into $Q$ prior to $v_{k+1}$. That is, $v' \in S''_{sub}$. So, by the induction hypothesis, $v$ reaches $v'$ by a path $P$ in $T$. Thus, by $P$ and $(v', v_{k+1})$, $v_{k+1}$ is reachable from $v$ by a path in $T$.
$\square$

We now prove Proposition 10 that shows the soundness of Algorithm 7.

### Proposition 10

Given a ground normal ASP program $\Pi$, an answer set $X$ for $\Pi$, an atom $p$ in $X$, an assumption $U$ in $Assumption(\Pi, X)$, and an offline justification $G = (V, E)$ of $p^+$ with respect to $X$ and $U$, Algorithm 7 returns an explanation tree $\langle V', E', l, \Pi, X \rangle$ in the and-or explanation tree for $p$ with respect to $\Pi$ and $X$.

### Proof of Proposition 10

We show what Algorithm 7 returns at Line 17, $T' = \langle V', E', l, \Pi, X \rangle$, is an explanation tree in the and-or explanation tree $T = \langle V_T, E_T, l, \Pi, X \rangle$ for $p$ with respect to $\Pi$ and $X$. That is, $T'$ satisfies Conditions $(i)$–$(iv)$ in Definition 3. In the following, we study each condition separately.

$(ii)$ To show that the root of $\langle V', E'\rangle$ is a vertex whose label is $p$, we need to show three cases hold; (1) there exists a vertex $v$ in $V'$ with label $p$, (2) every vertex $v' \in V'$ is reachable from $v$ by a path in $T'$. (3) $v$ has no in-going edge. In the following, we show that each case holds.

*Case 1.* Observe that a vertex is in $V'$ if and only if it is added into the queue $Q$. Then, due to Lines 2 and 3, there exists a vertex $v \in V'$ such that $l(v) = p$.

*Case 2.* The first element added into $Q$ is an atom vertex $v$ with $l(v) = p$, due to Lines 2 and 3. Note that a vertex is in $V'$ if and only if it is added into $Q$. Then, as $v$ is added into $Q$, by the observation, $v \in V'$. Now, pick a vertex $v' \in V'$. By the same observation, $v'$ is also added into $Q$. Since $v$ is the first element queued in $Q$, $v'$ is queued in $V'$ later on. Thus, by Lemma 19, $v'$ is reachable from $v$ by a path in $T'$.

*Case 3.* Assume otherwise. That is, $(v', v) \in E'$ for some vertex $v' \in V'$. The edges in $E'$ are constructed at Lines 10 and 15. Then, as $v$ is an atom vertex, $v'$ is a rule vertex. Observe that a vertex is in $V'$ if and only if it is added into $Q$. Then, since $v'$ is a rule vertex in $V'$, it is added into $Q$ at Line 16. So, due to Line 15, there exists an atom vertex $v'' \in V'$ such that $(v'', v') \in E'$. Thus, as $(v'', v'), (v', v) \in E'$, there exists a path $\langle v'', v', v\rangle$ in $T'$. Then, by Corollary 3, $l(v)^+$ is reachable from $l(v'')^+$ by a positive path $P$ in $G$. Moreover, as shown in Case 2 above, $v''$ is reachable from $v$ by a path $\langle v_1 = v, v_2, \ldots, v_n = v''\rangle$ in $T'$. So, by Corollary 3, $l(v'')^+$ is reachable from $l(v)^+$ by a positive path $P'$ in $G$. Then, by $P$ and $P'$, a positive cycle exists in $G$. As every offline justification is a safe offline e-graph due to Definition 17, we reach a contradiction.

$(i)$ By Condition $(ii)$ in Definition 3, we know that the root of $\langle V', E'\rangle$ is a vertex $v$ with $l(v) = p$. Then, if we show that for every vertex $v' \in V'$, $out_{E'}(v')$ is a subset of $E_T$, then we can conclude that $\langle V', E'\rangle$ is a subtree of $\langle V_T, E_T\rangle$. Now, pick a vertex $v' \in V'$. We consider two cases:

*Case 1.* Assume that $v'$ is an atom vertex. Take an out-going edge $(v', v'')$ of $v'$ in $E'$. To show that $(v', v'') \in E_T$, we need to show that $l(v'') \in \Pi_{X, anc_{T'}(v'')}(l(v'))$, due to Condition $(ii)$ in Definition 2. For that, we should show that $H(l(v'')) = l(v')$, $B^+(l(v'')) \subseteq X \backslash anc_{T'}(v'')$, $B^-(l(v'')) \cap X = \emptyset$ and $X \models B_{card}(l(v''))$, due to (8) in Section 4. As $\Pi$ is a ground normal ASP program, it does not contain cardinality expressions in its body. Thus, $X \models B_{card}(l(v''))$ trivially. In Algorithm 7, out-going edges of atom vertices are formed at Line 15. Then, due to Lines 13 and 14, $H(l(v'')) = l(v')$ and $B(l(v'')) = support(l(v')^+, G)$. Since $G$ is an offline justification, $G$ is an offline e-graph by Definition 17. Then, by Definition 13, $G$ is an $(X, U)$-based e-graph. So, by Definition 12, $support(l(v')^+, G)$ is an LCE of $l(v')^+$ with respect to $(X, U)$, so does $B(l(v''))$. According to Definition 11, as $B(l(v'')) \neq \{assume\}$, $B^+(l(v'')) \subseteq X$ and $B^-(l(v'')) \cap X = \emptyset$. To complete this part, it remains to show that $B^+(l(v'')) \subseteq X \backslash anc_{T'}(v'')$. Since $B^+(l(v'')) \subseteq X$, it is enough to show that $B^+(l(v'')) \cap anc_{T'}(v'') = \emptyset$. For that, assume $B^+(l(v'')) \cap anc_{T'}(v'') \neq \emptyset$. Let $a \in B^+(l(v'')) \cap anc_{T'}(v'')$. Since $a \in B^+(l(v''))$ and $B(l(v'')) = support(l(v')^+, G)$, $(l(v')^+, a^+, +) \in E$ holds. As $a \in anc_{T'}(v'')$, there exists a vertex $u \in V'$ where $l(u) = a$ such that a path $P = \langle v_1 = u, \ldots, v_n = v'\rangle$ in $T'$ exists. Then, due to Corollary 3, $l(v')^+$ is reachable from $a^+$ by a positive path in $G$. But, as $(l(v')^+, a^+, +) \in E$, $(a^+, a^+) \in E^{*,+}$

holds, i.e., there is a positive cycle in the offline justification, which is a contradiction, due to Definition 17.

*Case 2.* Assume that $v'$ is a rule vertex. Take an out-going edge $(v', v'')$ of $v'$ in $E'$. To show that $(v', v'') \in E_T$, we need to show that $l(v'') \in B^+(l(v'))$, due to Condition $(iii)$ in Definition 2. Out-going edges of $v'$ are formed at Line 10 which is reached only by satisfying the condition at Line 8. Then, due to Line 9, $l(v'') \in B^+(l(v'))$.

$(iii)$ Observe that an element is added into $Q$ once. Thus, we dequeue each atom vertex only once. Then, due to Lines 12–16, every atom vertex $v' \in V'$ has exactly one out-going edge, i.e., $deg_{E'}(v') = 1$.

$(iv)$ Take a rule vertex $v' \in V'$. Its edges are formed at Line 10. Then, due to the condition at Line 8 and the statement at Line 9, for every $a \in B^+(l(v'))$, there exists a vertex $v''$ such that $l(v'') = a$. Then, the condition holds.
□

Also, we can show that the and-or explanation tree exists for every atom in an answer set.

*Proposition* 1

Let $\Pi$ be a ground ASP program and $X$ be an answer set for $\Pi$. For every $p$ be in $X$, the and-or explanation tree for $p$ with respect to $\Pi$ and $X$ is not empty.

*Proof of Proposition 1*

By Proposition 9, we know that there exists an offline justification of $p^+$ with respect to $X$ and $X^- \backslash WF_{\Pi}^-$. Then, by Proposition 10, there is an explanation tree in the and-or explanation tree for $p$ with respect to $\Pi$ and $X$. That is, the and-or explanation tree for $p$ with respect to $\Pi$ and $X$ is not empty.
□

### A.3.2 Proof of Proposition 11 – Soundness of Algorithm 8

Before proving Proposition 11, we provide the necessary lemmas.

*Lemma 20*

Let $\Pi$ be a ground normal ASP program, $X$ be an answer set for $\Pi$, $p$ be an atom in $X$, $T = \langle V', E', l, \Pi, X \rangle$ be an explanation tree in the and-or explanation tree for $p$ with respect to $\Pi$ and $X$ such that for every $v, v' \in V'$, $l(v) = l(v')$ if and only if $v = v'$, $(V, E)$ be the output of Algorithm 8 called with inputs $\Pi, X, p$ and $T$, and $\langle v_1, v_2, v_3 \rangle$ be a path in $T$ such that $l(v_1)^+ \in V$. Then, $(l(v_1)^+, l(v_3)^+, +) \in E$ and $l(v_3)^+ \in V$.

*Proof of Lemma 20*

All of the nodes, except the one with label $\top$, are added to $V$ at Line 5 of Algorithm 8. As $l(v_1)^+ \in V$, $v_1$ is extracted from the queue $Q$ at Line 4. Then, since $\langle v_1, v_2, v_3 \rangle$ is a path in $T$ and every atom vertex in $V'$ has exactly one child due to Condition $(iii)$ in Definition 3, $v_2$ is obtained at Line 6. By the condition at Line 9, every child of $v_2$ is considered in the loop. Accordingly, at Line 10, the edge $(l(v_1)^+, l(v_3)^+, +)$ is added into $E$. Also, $v_3$ is

added into $Q$ at Line 11. Then, due to the condition at Line 3 and the statements at Lines 4 and 5, $l(v_3)^+ \in V$.

$\square$

*Corollary* 4

Let $\Pi$ be a ground normal ASP program, $X$ be an answer set for $\Pi$, $p$ be an atom in $X$, $T = \langle V', E', l, \Pi, X \rangle$ be an explanation tree in the and-or explanation tree for $p$ with respect to $\Pi$ and $X$ such that for every $v, v' \in V'$, $l(v) = l(v')$ if and only if $v = v'$, $(V, E)$ be the output of Algorithm 8 called with inputs $\Pi, X, p$ and $T$, and $\langle v_1, v_2, \ldots, v_n \rangle$ be a path in $T$ such that $l(v_1)^+ \in V$. Then, $\langle l(v_1)^+, l(v_3)^+, l(v_5)^+, \ldots, l(v_n)^+ \rangle$ is a path in $(V, E)$.

*Proof of Corollary 4*

Since $\langle v_1, v_2, v_3 \rangle$ is a path in $T$ and $l(v_1)^+ \in V$, by Lemma 20, $(l(v_1)^+, l(v_3)^+, +) \in E$ and $l(v_3)^+ \in V$. Similarly, $(l(v_3)^+, l(v_5)^+, +) \in E$ and $l(v_5)^+ \in V$. Then, this incremental application of Lemma 20 leads that $\langle l(v_1)^+, l(v_3)^+, l(v_5)^+, \ldots, l(v_n)^+ \rangle$ is a path in $(V, E)$.

$\square$

We now prove Proposition 11 which shows that Algorithm 8 creates an offline justification of the given atom in the reduct of the given ASP program with respect to the given answer set, provided that labels of the vertices of the given explanation tree are unique labels, i.e., no two different vertices labels the same entity.

*Proposition* 11

Given a ground normal ASP program $\Pi$ , an answer set $X$ for $\Pi$, an atom $p$ in $X$, and an explanation tree $\langle V', E', l, \Pi, X \rangle$ in the and-or explanation tree for $p$ with respect to $\Pi$ and $X$ such that for every $v, v' \in V'$, $l(v) = l(v')$ if and only if $v = v'$, Algorithm 8 returns an offline justification of $p^+$ in $\Pi^X$ with respect to $X$ and $\emptyset$.

*Proof of Proposition 11*

To show that the output $(V, E)$ of Algorithm 8 at Line 13 is an offline justification of $p^+$ in $\Pi^X$ with respect to $X$ and $\emptyset$, one needs to show that the following conditions hold.

(1) $\emptyset \in Assumptions(\Pi^X, X)$;
(2) $(V, E)$ is an e-graph for $\Pi^X$;
(3) $(V, E)$ is a $(X, \emptyset)$-based e-graph of $p^+$;
(4) $(V, E)$ is an offline e-graph of $p^+$ with respect to $X$ and $\emptyset$.

*Condition (1)* We show that $\emptyset \in Assumptions(\Pi^X, X)$. As $\Pi^X$ is a positive program, i.e., it does not contain any negative atoms, by Definition 14, $\mathcal{TA}_{\Pi^X}(X) = \emptyset$. Also, due to Definition 15, $NR(\Pi^X, X) = \Pi^X$. Then, by Definition 16, $\emptyset \in Assumptions(\Pi^X, X)$.

*Condition (2)* We show that $(V, E)$ is an e-graph for $\Pi^X$. For that, $(V, E)$ should satisfy Conditions $(i) - (iv)$ in Definition 9.

*(i)* Consider two cases.

*Case 1.* Take a node $b^+ \in V \setminus \{\top\}$. It is added to $V$ at Line 5. Then, there exists an atom vertex $v \in V'$ such that $l(v) = b$. By Condition *(iii)* in Definition 3, $v$ has exactly one child $v'$ in $\langle V', E' \rangle$, which is a rule vertex. Then, depending on whether $l(v')$ is a fact or not, an out-going edge of $b^+$ in $E$ is formed at Line 8 or 10. So, $b^+$ is not a sink.

*Case 2.* Let $l$ be $\top$. Then, it is added to $V$ at Line 12. The edges in $E$ are formed at Lines 8 and 10. Accordingly, $l$ has no out-going edge, i.e., it is a sink.

Therefore, $(V, E)$ is an e-graph for $\Pi^X$.

*(ii)* Due to Lines 8 and 10, clearly, for every $b \in V$ there is no edges in the form of $(b, assume, -)$ and $(b, \bot, -)$ in $E$.

*(iii)* Similar to *(ii)*, for every $b \in V$ there is no edges in the form of $(b, assume, +)$ and $(b, \top, +)$ in $E$.

*(iv)* Let $b^+ \in V$ such that $(b^+, \top, +) \in E$. As out-going edges are created at Lines 8 and 10, $(b, \top, +)$ must be formed at Line 8. Then, there exists an atom vertex $v \in V'$ such that $l(v) = b$ and the label of the child $v'$ of $v$ is a fact (Line 7). Then, since the condition at Line 9 is not satisfied, it is not possible that $b^+$ has another out-going edge.

*Condition (3)* We show that $(V, E)$ is an $(X, \emptyset)$-based e-graph of $p^+$. By Condition (1) above, we know that $(V, E)$ is an e-graph. To show that $(V, E)$ is an $(X, \emptyset)$-based e-graph of $p^+$, we need to show that Conditions *(i)* and *(ii)* in Definition 12 hold.

*(i)* Take a node $c^+ \in V$. It is added to $V$ at Line 5. Then, there exists an atom vertex $v' \in V'$ such that $l(v') = c$. By Condition *(ii)* in Definition 3, we know that the root of $\langle V', E' \rangle$ is a vertex $v$ where $l(v) = p$. So, $v'$ is reachable from $v$ by a path $\langle v_1 = v, v_2, v_3, \dots, v_n = v' \rangle$ in $\langle V', E' \rangle$. Also, as $v$ is added into $Q$ at Line 2, we know that $l(v)^+ \in V$ by Line 5. Then, by Corollary 4, $\langle l(v)^+, l(v_3)^+, \dots, l(v')^+ \rangle$ is a path in $(V, E)$. That is, $c^+$ is reachable from $p^+$.

*(ii)* Let $G = (V, E)$. Take a node $c^+ \in V \setminus \{\top\}$. Then, $support(c^+, G) = \{a \mid (c^+, a^+, +) \in E\}$ or $support(c^+, G) = \{\top\}$.

Assume that $support(c^+, G) = \{a \mid (c^+, a^+, +) \in E\}$. As $c^+ \in V \setminus \{\top\}$, there exists an atom vertex $v \in V'$ such that $l(v) = c$. By Condition *(iii)* in Definition 3, there exists exactly one child $v'$ of $v$ in $\langle V', E' \rangle$. Then, due to the condition at Line 9, for each $a \in support(c^+, G)$, $(v', v'') \in E'$ where $v'' \in V'$ with $l(v'') = a$. Thus, $B^+(l(v')) = support(c^+, G)$. By Condition *(ii)* in Definition 2, $H(l(v')) = c$ and $B^+(l(v')) \subseteq X$. Hence, $support(c^+ G)$ is an LCE of $(X, \emptyset)$.

*Condition (4)* We show that $(V, E)$ is an offline e-graph of $p^+$ with respect to $X$ and $\emptyset$. By Condition (3) above, we know that $(V, E)$ is an $(X, \emptyset)$-based e-graph of $p^+$. Then, to show that $(V, E)$ is an offline e-graph of $p^+$ with respect to $X$ and $\emptyset$, we need to show that $(V, E)$ satisfies Conditions *(i)* and *(ii)* in Definition 13. As edges in $E$ are formed at Lines 8 and 10, for every $b \in V$ there are no edges in the form of $(b, assume, +)$ and $(b, assume, -)$ in $E$. Then, conditions are trivially satisfied.

□



\end{document}